\documentclass{article} 
\usepackage[dvipsnames]{xcolor} 
\usepackage{iclr2025_conference,times}

\usepackage{amsmath,amsfonts,bm}

\def\eqref#1{equation~\ref{#1}}

\def\1{\bm{1}}

\DeclareMathAlphabet{\mathsfit}{\encodingdefault}{\sfdefault}{m}{sl}
\SetMathAlphabet{\mathsfit}{bold}{\encodingdefault}{\sfdefault}{bx}{n}

\DeclareMathOperator*{\argmax}{arg\,max}

\usepackage{hyperref}
\usepackage{url}

\usepackage[utf8]{inputenc} 
\usepackage[T1]{fontenc}    
\usepackage{booktabs}       
\usepackage{amsfonts}       
\usepackage{nicefrac}       
\usepackage{microtype}      

\usepackage{arydshln}
\usepackage{subcaption}
\usepackage{bm}
\usepackage{enumitem}
\usepackage{amsmath}
\usepackage{amssymb}
\usepackage{amsthm}
\usepackage{mathtools}
\usepackage[english]{babel}
\usepackage[capitalize,noabbrev]{cleveref}

\allowdisplaybreaks

\newtheorem{theorem}{Theorem}
\newtheorem{lemma}{Lemma}
\newtheorem{corollary}{Corollary}
\newtheorem{proposition}{Proposition}
\newtheorem{definition}{Definition}
\newtheorem{assumption}{Assumption}
\newtheorem{remark}{Remark}

\newcommand{\indep}{\perp\!\!\!\!\perp}
\newcommand{\equal}{=}

\usepackage{tikz}
\usetikzlibrary{calc,fit,positioning,backgrounds}

\newcommand{\tikzxmark}{%
\tikz[scale=0.23] {
    \draw[line width=0.7,line cap=round] (0,0) to [bend left=6] (1,1);
    \draw[line width=0.7,line cap=round] (0.2,0.95) to [bend right=3] (0.8,0.05);
}}
\newcommand{\tikzcmark}{%
\tikz[scale=0.23] {
    \draw[line width=0.7,line cap=round] (0.25,0) to [bend left=10] (1,1);
    \draw[line width=0.8,line cap=round] (0,0.35) to [bend right=1] (0.23,0);
}}

\newcommand{\circo}{~\raisebox{1pt}{\tikz \draw[line width=0.6pt] circle(1.2pt);}~}

\title{Counterfactual Generative Modeling with \\Variational Causal Inference}

\author{%
  Yulun Wu \\
  University of California, Berkeley \\
  \texttt{yulun\_wu@berkeley.edu} \\
  \And
  Louie McConnell \\
  Genentech \\
  \texttt{mcconnl3@gene.com} \\
  \And
  Claudia Iriondo \\
  Genentech \\
  \texttt{iriondoc@gene.com} \\
}

\iclrfinalcopy 
\begin{document}

\maketitle

\begin{abstract}
  Estimating an individual's counterfactual outcomes under interventions is a challenging task for traditional causal inference and supervised learning approaches when the outcome is high-dimensional (e.g. gene expressions, facial images) and covariates are relatively limited. In this case, to predict one's outcomes under counterfactual treatments, it is crucial to leverage individual information contained in the observed outcome in addition to the covariates. Prior works using variational inference in counterfactual generative modeling have been focusing on neural adaptations and model variants within the conditional variational autoencoder formulation, which we argue is fundamentally ill-suited to the notion of counterfactual in causal inference. In this work, we present a novel variational Bayesian causal inference framework and its theoretical backings to properly handle counterfactual generative modeling tasks, through which we are able to conduct counterfactual supervision end-to-end during training without any counterfactual samples, and encourage disentangled exogenous noise abduction that aids the correct identification of causal effect in counterfactual generations. In experiments, we demonstrate the advantage of our framework compared to state-of-the-art models in counterfactual generative modeling on multiple benchmarks.

\end{abstract}

\section{Introduction}
\label{introduction}

In traditional causal inference, heterogeneous treatment effect is typically formulated as the interventional model $p(Y | X, \mathrm{do}(T))$ (or $\mathbb E[Y | X, \mathrm{do}(T)]$) with outcome $Y$, covariates $X$, treatment $T$, and estimated by 
the observed covariate-specific efficacy $p(Y | X, T)$ under the ignorability assumption \citep{rosenbaum1983central, pearl1995causal}. However, in cases such as the single-cell perturbation datasets \citep{dixit2016perturb, norman2019exploring, schmidt2022crispr} where $Y$ has thousands of dimensions while $X$ has only two or three categorical features, such model could hardly be relied on to produce useful individualized results. To construct outcome $Y'$ under a counterfactual treatment $T'$ in such high-dimensional outcome scenario, it is important and necessary to learn the individual-specific efficacy $p(Y' | Y, X, T, T')$ conditioning on the factual outcome $Y$ itself, which leverages the rich information embedded in the factual outcome that cannot be recovered by the handful of covariates. For example, given a cell with type A549 ($X$) that received SAHA drug treatment ($T$), we may want to know what its gene expression profile ($Y'$) would have looked like if it had received Dex drug treatment ($T'$) instead. In this case, we would want to take the profiles of other cells with type A549 that indeed received Dex as reference, such that the counterfactual construction $Y'$ exhibits the treatment characteristics of Dex on A549, but would also want to extract as much individual information as possible from its own expression profile ($Y$) such that the counterfactual construction could preserve this cell's individuality such as cell state.

Yet the lack of observability of counterfactual outcome $Y'$ makes this objective intractable and hence presents a major difficulty for supervised learning when $Y$ is taken as an input. In previous works that involve self-supervised counterfactual generation \citep{kocaoglu2017causalgan, louizos2017causal, yoon2018ganite, pawlowski2020deep, yang2021causalvae, kim2021counterfactual, lotfollahi2021learning, sauer2021counterfactual, feng2022principled, shen2022weakly} based on Variational Autoencoders (VAE) \citep{kingma2013auto} and Generative Adversarial Networks (GAN) \citep{goodfellow2014generative}, there is no counterfactual supervision during training, and hence no explicit regulation on the trade-off between treatment characteristics and individuality. When such regulation does not exist, there is a disconnection between training and counterfactual inference. To see this, consider a conditional generative model $g_\theta (y, t)$ that is only supervised on the reconstruction loss with respect to outcome $y$ during training. In this case, the model is free to ignore the extra treatment $t$ and converge to a state such that $g_\theta (y, t) = g_\theta (y, 0)$ \citep{chen2016infogan}, hence preserving maximum individuality yet minimum treatment characteristics in counterfactual construction $g_\theta (y, t')$ during inference. 
This is particularly an issue for high-fidelity counterfactual generation with state-of-the-art Hierarchical Variational Autoencoders (HVAE) \citep{vahdat2020nvae, child2020very}, as the bottleneck latent representations possess much larger number of dimensions compared to the observed outcome $y$. In more recent works on counterfactual generative modeling with HVAEs and diffusion models \citep{sanchez2022diffusion, monteiro2023measuring, ribeiro2023high}, only \citet{ribeiro2023high} touched on the issue of counterfactual supervision, yet its proposed method does not have a theoretical backing in variational inference, nor could it be conducted end-to-end during training. As we will see in the proceeding sections of this work, there is a fundamental incompatibility of the conditional VAE formulation with counterfactual generative modeling, which is the root cause of such issue. In fact, this incompatibility also exhibits itself in another important aspect of counterfactual generative modeling which is latent disentanglement -- the partially abducted exogenous noise (i.e. shallow-level latent representation) in conditional HVAEs and diffusion models is able to encode individuality better than traditional VAEs, but at the same time is still heavily entangled with treatment variables and often result in lingering characteristics of the factual treatment in counterfactual construction \citep{monteiro2023measuring, sanchez2022diffusion}. More introduction regarding latent disentanglement can be found in Appendix \ref{sec:intro-disentanglement}.

Our contributions in this work are summarized as follow: \textbf{1)} we propose a formulation and stochastic optimization scheme for counterfactual generative modeling by specifically formulating counterfactual variables and using variational inference to derive the evidence lower bound (ELBO) for the individual-level likelihood $p(Y' | Y, X, T, T')$ -- a well-motivated and fitting objective for counterfactual generative modeling instead of the marginal-level (interventional) conditional likelihood $p(Y | X, T)$ or even joint likelihood $p(Y, X, T)$ used by VAE-based prior works. \textbf{To emphasize, the contribution here is not to propose another model variant under the conditional VAE formulation such as conditional HVAEs or diffusion models \citep{sanchez2022diffusion, monteiro2023measuring, ribeiro2023high}, but to fundamentally change the VAE formulation.} We call this framework Variational Causal Inference (VCI). An explanation to why the conditional VAE formulation (including diffusion models) is ill-suited to counterfactual generative modeling and a straightforward comparison between VAE's formulation and ours can be found in Appendix \ref{sec:comparison-variational}. \textbf{2)} Our proposed optimization scheme allows us to conduct \textbf{counterfactual supervision end-to-end during training} alongside self-supervision, even without any counterfactual observations, which has not been done in prior works to the best of our knowledge. \textbf{3)} This workflow of constructing and supervising counterfactual outcomes during training presents us the unique opportunity to naturally enforce \textbf{disentangled exogenous noise abduction} through distribution alignment on matching pairs \citep{shu2019weakly, locatello2020weakly, brehmer2022weakly} without having paired data, which has not been done in counterfactual generative modeling, to the best of our knowledge. \textbf{4)} We further propose a robust estimation scheme for high-dimensional marginal causal parameters leveraging this individual-level counterfactual construction framework. No prior work in counterfactual generative modeling conducts such asymptotically efficient marginal estimation upon acquiring individual predictions, to the best of our knowledge. \textbf{5)} In experiments, our proposed method is evaluated on datasets with vector outcomes -- single cell perturbation datasets, as well as datasets with image outcomes -- facial imaging and handwritten digits datasets, and compared to state-of-the-art models in the two domains. The results show that ours outperformed state-of-the-arts in both domains with notable margins. Related work and comparative analysis can be found in Appendix \ref{sec:related-work}.


\section{Proposed Method}
\label{proposed-method}


\subsection{Formulation and Intuition}
\label{semi-autoencode}

\begin{figure}
    \centering
    \begin{subfigure}[b]{0.45\textwidth}
        \centering
        \begin{tikzpicture}
            \tikzstyle{main}=[circle, minimum size = 8mm, thick, draw =black!80, node distance = 8mm]
            \tikzstyle{connect}=[-latex, thick]
            \tikzstyle{box}=[rectangle, draw=black!100]
              \node[main, fill = white!100] (X) [label=below:$X$] { };
              \node[main, fill = black!10] (Tp) [right=of X,label=right:$T'$] { };
              \node[main, fill = white!100] (T) [left=of X,label=below:$T$] { };
              \node[main, fill = black!50] (Z) [above=of X,label=above:$Z$] {};
              \node[main, fill = black!50] (Yp) [above=of Tp,label=right:$Y'$] { };
              \node[main, fill = white!100] (Y) [above=of T,label=above:$Y$] { };
              \path (X) edge [connect] (Z)
                    (X) edge [connect] (T)
                    (X) edge [connect] (Tp)
            		(Z) edge [connect] (Y)
            		(Z) edge [connect] (Yp)
            		(T) edge [connect] (Y)
            		(Tp) edge [connect] (Yp);
        \end{tikzpicture}
        \caption{Bayesian network}
        \label{causal_diagram-bayesian}
    \end{subfigure}
    \begin{subfigure}[b]{0.45\textwidth}
        \centering
        \begin{tikzpicture}
            \tikzstyle{main}=[circle, minimum size = 8mm, thick, draw =black!80, node distance = 8mm]
            \tikzstyle{connect}=[-latex, thick]
            \tikzstyle{box}=[rectangle, draw=black!100]
              \node[main, fill = white!100] (X) [label=below:$X$] { };
              \node[main, fill = black!50, minimum size = 4mm] (UX) [left=of X,label=left:{\tiny$U_X$}, xshift=6mm, yshift=4mm] { };
              \node[main, fill = white!100] (T) [right=of X,label=below:$T$, xshift=6mm] { };
              \node[main, fill = black!50, minimum size = 2mm] (UT) [left=of T,label=left:{\tiny $U_T$}, xshift=6mm, yshift=4mm] { };
              \begin{scope}[on background layer]
                \node[main, fill = black!10] (Tp) [right=of X,label=right:$T'$, xshift=8mm] { };
              \end{scope}
              \node[main, fill = black!50] (Z) [above=of X,label=above:$Z$, xshift=10mm] {};
              \node[main, fill = black!50, minimum size = 4mm] (UY) [left=of Z,label=left:{\tiny$U_Y$}, xshift=6mm, yshift=4mm] { };
              \node[main, fill = white!100] (Y) [right=of Z,label=above:$Y$, xshift=6mm] { };
              \begin{scope}[on background layer]
                \node[main, fill = black!50] (Yp) [right=of Z,label=right:$Y'$, xshift=8mm] { };
              \end{scope}
              \path (UX) edge [connect] (X)
                    (X) edge [connect] (Z)
                    (UY) edge [connect] (Z)
                    (X) edge [connect] (T)
                    (UT) edge [connect] (T)
                    (Z) edge [connect] (Y)
                    (T) edge [connect] (Y);
        \end{tikzpicture}
        \caption{Counterfactual view of the Bayesian network}
        \label{causal_diagram-counterfactual}
    \end{subfigure}
    \caption{The causal diagram where variables are generated from the following SCM: $X=f_X(U_X)$; $T=f_T(X, U_T)$, $T'=f_T(X, U'_T)$ where $U_T$, $U'_T$ are i.i.d.; $Z=f_Z(X, U_Y)$; $Y=f_Y(Z, T, \epsilon_Y)$, $Y'=f_Y(Z, T', \epsilon'_Y)$ where $\epsilon_Y$, $\epsilon'_Y$ are i.i.d., independent of all $U$s, and are constants w.p. 1 if the consistency assumption is assumed. Endogenous variables follow the Bayesian network on the left under the ignorability assumption, with permissibly one additional edge from either $Z$ or $T$ to $X$ if $U_Y$ or $U_T$ is dependent of $U_X$. We call $O=(X, T, Y)$ observed variables, $B=(X, T, T', Y)$ sample variables ($T'$ are sampled for model training and evaluation), $D=(X, T, T', Y, Y')$ full data variables and $W=(Z, X, T, T', Y, Y')$ all variables. White nodes are observed, light grey nodes are assigned during training and inference, dark grey nodes are unobserved.
    }
    \label{causal_diagram}
\end{figure}
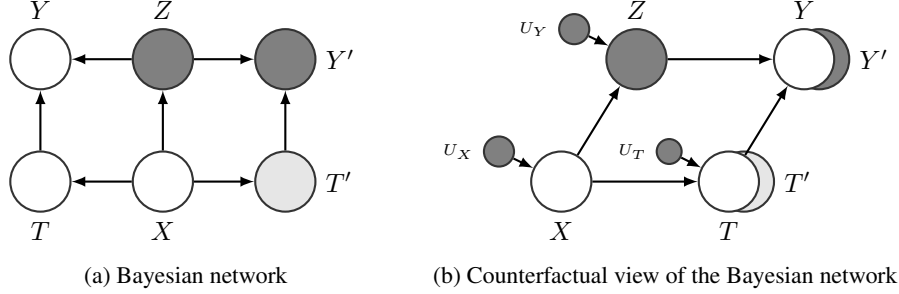

Let $Y: \Omega \rightarrow (\mathcal{Y}, \Sigma_\mathcal{Y})$ be the outcome, $X: \Omega \rightarrow (\mathcal{X}, \Sigma_\mathcal{X})$ be the covariates, $T: \Omega \rightarrow (\mathcal{T}, \Sigma_\mathcal{T})$ be the treatments and $Z: \Omega \rightarrow (\mathcal{Z}, \Sigma_\mathcal{Z})$ be a latent feature vector on probability space $(\Omega, \Sigma, P)$. Suppose the causal relations between random variables (and random vectors) follow a structural causal model (SCM) \citep{pearl1995causal} defined in Figure \ref{causal_diagram}, where we adopt twin networks \citep{balke2022probabilistic} to formulate counterfactuals $Y'$ and $T'$ as separate variables apart from $Y$ and $T$ having a conditional distribution $p(Y', T' | Z, X)$ identical to that of its factual counterpart $p(Y, T | Z, X)$. Different from \citet{vlontzos2023estimating}, we introduce latent $Z$ in high-dimensional outcome settings as a feature vector of $Y$ such that random factor $\epsilon_Y$ does not capture any more uncertainty that can be retrieved from its counterfactual counterpart, i.e. $\epsilon_Y \indep Y'$ and similarly $\epsilon_Y' \indep Y$. The consistency and ignorability assumptions corresponding to this causal formulation can be found in Appendix \ref{sec:causal-assumptions}. Under the consistency assumption (Assumption \ref{assumption-consistency} and Remark \ref{remark-consistency}), $\epsilon_Y = 0$ and there is no more uncertainty in $Y$ beyond $Z$, i.e. the shared random factor $U_Y$ fully captures the exogenous noise injected to $Y$ and $Y'$. Hence, $Z$ in this setting can be seen as an unobserved summary random vector of the covariates $X$ and exogenous noise $U_Y$. 
Note that the consistency assumption is not generally required for our subsequent variational inference, but is required for $Y'_{\mathrm{do}(T')}$ to be formally defined as the counterfactual of $Y$ under the three-layer causal hierarchy of \citet{pearl2009causality}. The ignorability assumption (Assumption \ref{assumption-ignorability} and Remark \ref{remark-ignorability}) is generally required, under which $do(T'=t')$ and $T'=t'$ result in the same outcome distribution on $Y'$ conditioned on $X$ or $Z$. See Appendix \ref{sec:comparison-causal} for the connection between our formulation and the traditional SCM formulation.

\paragraph{Semi-autoencoding} Under this formulation, $Z$ has a posterior distribution $p(Z | Y, T, X)$ given the observed variables; $Y$ and $Y'$ can be constructed by $p(Y | Z, T)$ and $p(Y' | Z, T')$ respectively with latent features $Z$. Similar to prior works in autoencoder \citep{vincent2008extracting, bengio2014deep}, we can estimate the latent recognition model (i.e. exogenous noise abduction model) and outcome construction model with deep neural network encoder $q_\phi$ and decoder $p_\theta$. 
Given a sample $b=(x, t, t', y)$, while the reconstruction $y_{\theta, \phi}$ of $y$ is self-supervised, we can only assess the counterfactual construction $y'_{\theta, \phi}$ under $t'$ by looking at its resemblance to similar individuals that indeed received treatment $t'$. Hence, naively, we may conduct counterfactual supervision during training with a semi-autoencoding (SAE) loss function as follow
\begin{align}
\label{sae_loss}
    L_{\theta, \phi}(b) = L_2(y_{\theta,\phi}, y) -\omega\cdot \ell_{\hat{p}(Y' | x, t')}(y'_{\theta,\phi})
\end{align}
where $\omega$ is a scaling coefficient and $\hat{p}$ is the traditional covariate-specific outcome model fit on the observed variables $(X, T, Y)$ (notice that $p(Y' | x, T'=t')=p(Y | x, T=t')$). The intuition is that, if $y'_{\theta,\phi}$ is indeed one's outcome under $t'$, then the likelihood of $y'_{\theta,\phi}$ coming from the outcome distribution of individuals with the same attributes $x$ that factually received treatment $t'$ should be high. In practice, 
we can fit $\hat{p}(Y | X, T)$ in an end-to-end fashion using a discriminator $\mathcal{D}(X, T, Y)$ on factual triplets $(x, t, y)$ and counterfactual triplets $(x, t', y'_{\theta,\phi})$ with the adversarial approach \citep{goodfellow2014generative}. Specifically, use discriminator loss $L_\mathcal{D}(b, y'_{\theta,\phi}) = -\log [ \mathcal{D}(x, t, y) ] - \log [ 1 - \mathcal{D}(x, t', y'_{\theta,\phi}) ]$ to train model $\mathcal{D}$, and use generator loss $\ell_{\hat{p}(Y' | x, t')}(y'_{\theta,\phi}) = \log \mathcal{D}(x, t', y'_{\theta,\phi})$ in $L_{\theta, \phi}(b)$ (notice that $p(X, T) = p(X, T')$). This end-to-end approach prevents $(p_\theta, q_\phi)$ from exploiting a pre-trained $\hat{p}$ model. In cases where covariates are limited and discrete such as the single-cell perturbation datasets, $\hat{p}(Y | X, T)$ can simply be a smoothed empirical outcome distribution under treatment $T$ stratified by covariates $X$. 
The SAE approach motivates the derivation of our main theorem presented in the next section, and serves as a baseline comparison to our main algorithm in the ablation study of the experiments section.

\subsection{Variational Causal Inference}
\label{variational-causal-inference}

\begin{figure}
    \centering
    \includegraphics[width=0.8\linewidth]{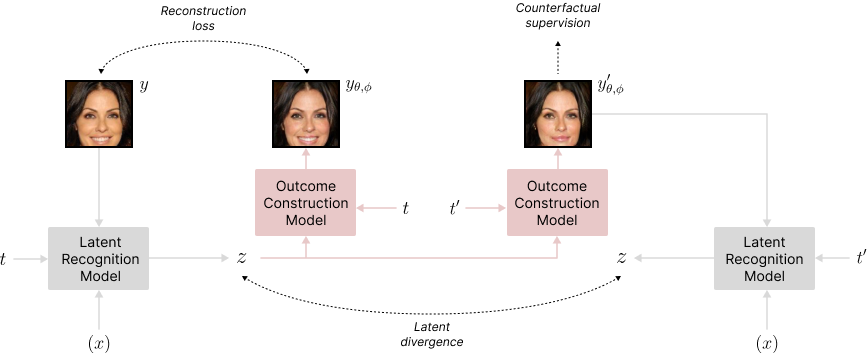}

    \caption{Model workflow of the Variational Causal Inference (VCI) framework. In a forward pass, the encoding model takes outcome $y$ as well as its treatments $t$ and covariates $x$ (if any) as inputs and attains latent feature $z$; $(z, t)$ and $(z, t')$ where $t'$ is the counterfactual treatments are separately passed into the decoding model to attain reconstruction $y_{\theta, \phi}$ and counterfactual construction $y'_{\theta, \phi}$, for which reconstruction loss and counterfactual supervision loss are evaluated; $y'_{\theta, \phi}$ is then passed back into the encoding model along with $t'$ and $x$ to attain $z$ again, which is encouraged to match the latent feature $z$ previously acquired from encoding the factual outcome $y$ and treatment $t$ through the latent divergence term. The neural architectures of the latent recognition and outcome construction models for image generation tasks can be found in Appendix \ref{sec:architecture}.}
    \label{fig:hvci-workflow}
\end{figure}

Now we present our main result by rigorously formulating the objective, providing a probabilistic theoretical backing, and specifying an optimization scheme that reflects the intuition described in the previous section. Suppose we want to optimize $p(Y' | Y, X, T, T')$ instead of the traditional outcome likelihood $p(Y | X, T)$. The following theorem induces the evidence lower bound (ELBO) and thus provides a roadmap for stochastic optimization:

\begin{theorem}
    \label{elbo}
    Suppose a collection of random variables $W$ follows a causal structure defined by the Bayesian network in Figure \ref{causal_diagram}. Then $J(D) = \log [ p (Y' | Y, X, T, T') ]$ has the following variational lower bound:
    \begin{align}
        \label{eq-elbo}
        J(D) &\geq \mathbb E_{p (Z | Y, T, X)} \log \left[ p (Y | Z, T) \right] - D \left[ p (Y | X, T) \parallel p (Y' | X, T') \right] \nonumber \\
        &\quad - D_\mathrm{KL} \left[ p (Z | Y, T, X) \parallel p (Z | Y', T', X) \right]
    \end{align}
    where $D [ p \parallel q ] = \log p - \log q$.
\end{theorem}

Proof of Theorem \ref{elbo} and proofs of all theoretical results in the following sections can be found in Appendix \ref{proofs}. Theorem \ref{elbo} suggests that, in order to maximize the counterfactual outcome likelihood $p(Y' | Y, X, T, T')$, we can maximize the following estimated ELBO parameterized with estimating models $p_\theta$ and $q_\phi$ given each sample $b=(x, t, t', y)$:
\begin{align}
    \label{VCI-objective}
    J_{\theta, \phi}(b) &= \mathbb E_{q_\phi (z | y, t, x)} \log \left[ p_\theta (y | z, t) \right] + \log \left[ \hat{p} (y'_{\theta,\phi} | x, t') \right] \nonumber \\
    &\quad - D_\mathrm{KL} \left[ q_\phi (z | y, t, x) \parallel q_\phi (z | y'_{\theta,\phi}, t', x) \right],
\end{align}
where $y'_{\theta,\phi} = \mathbb E_{p_\theta (\cdot | z_\phi, t')} Y' = \mathbb E_{p_\theta (\cdot | \mathbb E_{q_\phi (\cdot | y, t, x)} Z, t')} Y'$ and $\hat{p}$ can be estimated with the approaches described at the end of Section \ref{semi-autoencode}. Note that $\hat{p}(y | x, t)$ does not impose gradient on $(p_\theta, q_\phi)$ and is thus omitted in $J_{\theta, \phi}(b)$. The stochastic optimization of ELBO is hence conducted on the objective $J_{p_\mathrm{data}} (p_\theta, q_\phi) = \mathbb E_{p_\mathrm{data}} J_{\theta, \phi}(b)$ where $t' \sim p_\mathrm{data}(T | x)$. A figure demonstrating the workflow of this stochastic optimization scheme can be found in Figure \ref{fig:hvci-workflow}.

As can be seen from Equation \ref{VCI-objective}, the ELBO consists of the individual-specific factual outcome likelihood $E_{q_\phi (z | y, t, x)} \log [ p_\theta (y | z, t) ]$ and the covariate-specific counterfactual outcome likelihood $\log [ p (y'_{\theta,\phi} | x, t') ]$, echoing the intuition highlighted by Equation \ref{sae_loss}, with an additional divergence term $-D_\mathrm{KL} [ q_\phi (z | y, t, x) \parallel q_\phi (z | y'_{\theta,\phi}, t', x) ]$ that regularizes the similarity across latent distributions under different potential scenarios. Note that this framework is fundamentally different from VAE formulations such as CVAE \citep{sohn2015learning} or CEVAE \citep{louizos2017causal}, since the variational lower bound is derived directly from the causal objective and the KL-divergence term serves causal purposes that are entirely different from the prior latent bounding purpose of VAEs. An interpretation of this divergence term is given in the paragraph below. A comparison between VCI and conditional VAE using standard VAE notations is given in Appendix \ref{sec:comparison-variational}. A summary of different optional gradient detaching patterns for training stability is given in Appendix \ref{detaching-pattern}. An approach to conduct counterfactual supervision implicitly using traditional variational inference is given in Appendix \ref{sec:implicit-cf-supervision}.

\paragraph{Divergence Interpretation} The divergence term $-D_\mathrm{KL} [ q_\phi (z | y, t, x) \parallel q_\phi (z | y'_{\theta,\phi}, t', x) ]$ is the key of the framework and it serves two purposes. Firstly, it adds robustness to latent encoding by encouraging the encoding model to recognize individual features contained in both factual and counterfactual outcomes. In fact, these common features are well disentangled from the treatment variables, which we will discuss in-depth in the next section. Secondly, it encourages the preservation of individuality in counterfactual outcome constructions. To see this, notice that if the counterfactual outcome construction was only penalized by the covariate-specific likelihood loss $-\log [ \hat{p} (y'_{\theta,\phi} | x, t') ]$, the counterfactual decoding model $p_\theta(Y' | z_\phi, t')$ could learn to completely discard the identity of individual subjects represented in latent features $z_\phi$ once it detects a counterfactual treatment $t' \neq t$ in the inputs. Such behavior is regulated by the divergence term, since otherwise we would not be able to recover a latent distribution $q_\phi (Z | y'_{\theta,\phi}, t', x)$ close to $q_\phi (Z | y, t, x)$.

\paragraph{Robust Marginal Estimation} Similar to the augmented inverse propensity weighted (AIPW) estimator in traditional causal inference, the individual-level model predictions acquired by our framework can also aid the robust estimation of average treatment effect $\Psi(p) = \mathbb E_p [ Y'_{\mathrm{do}(T'=\alpha)} ]$ for a treatment level $\alpha$ of interest if such estimation is desired. Under the formulation in Figure \ref{causal_diagram}, we have the following robust estimator that is asymptotically efficient under some regularity conditions \citep{van2006targeted}:
\begin{align}
    \label{ATT-estimator}
    \hat{\Psi}_n (o) = \frac{1}{n} \sum_{k=1}^n \left\{ \frac{I(t_k = \alpha)}{\hat{e}(t_k | x_k)} \cdot y_k + \left(1-\frac{I(t_k = \alpha)}{\hat{e}(t_k | x_k)}\right) \cdot \mathbb E_{p_\theta} \left[ Y' | z_{k, \phi}, T'=\alpha \right] \right\},
\end{align}
where $o_k = (x_k, t_k, y_k) \overset{\mathrm{iid}}{\sim} p(x, t, y)$ is the vector of observed triplet of the $k$-th individual, $z_{k,\phi} \sim q_\phi (y_k, t_k, x_k)$, and $\hat{e}$ is an estimation of the propensity score that satisfies the positivity assumption \citep{robins1986new}. See Appendix \ref{sec:covar-spec_ATT} for the derivation of this estimator. Notice that the regression adjustment term $\mathbb E_{p_\theta} [ Y' | z_{k, \phi}, T'=\alpha ]$ in ours has the ability to vary across different individuals within the same covariate group, which is not the case for that of the AIPW estimator in traditional causal formulation. Hence, it is meaningful to conduct the robust estimation of covariate-specific marginal treatment effect $\Xi(p) = \mathbb E_p [ Y'_{\mathrm{do}(X=c, T'=\alpha)} ]$ for a given covariate $c$ of interest, see Appendix \ref{sec:covar-spec_ATT} for details.

\subsection{Latent Identifiability and Exogenous Noise Disentanglement}

In any deep probabilistic model, there is a certain degree of freedom and arbitrarity to the learnt latent distribution, even with latent restrictions. Naturally, one might question the latent identification of our variational causal framework: when identifying counterfactual $Y'$ under $T'$, does the learnt latent $Z$ inevitably carry over some characteristics of the factual treatment in $Y$ and $T$? Ideally, we would like to aid counterfactual construction by attaining a clean disentanglement between the learnt latent and the observed treatment, such that the counterfactual generation does not preserve influence from the factual treatment. Such disentangled exogenous noise abduction fits the semantic of our generating process (Figure \ref{causal_diagram}), where $Z$ and $T$ are not descendant of each other and $Z \indep T | X$, as well as the semantic of causal identification in the traditional formulation, where the identifiability of $Y_{\mathrm{do}(T)}$ relies on the ignorability assumption that exogenous noise does not contain unobserved confounders. In this section, we show that optimizing the VCI framework in fact grants such disentanglement between $Z$ and $T$ under mild assumptions, leading to a more in-depth understanding of the purpose and behavior of the latent divergence term in VCI's optimization scheme.

Similar to \citet{shu2019weakly}, we formally define and measure disentanglement in terms of oracle consistency and oracle restrictiveness, except that our definitions are generalized to the case of stochastic models and models with auxiliary attributes. For notation simplicity of the oracle operation on functions with auxiliary inputs, we use $\gamma \circo \zeta (a; b)$ to denote $\gamma(c, b)$ where $c = \zeta(a)$ if $\zeta$ is a deterministic function and $c \sim \zeta(a)$ if $\zeta$ is a probability measure. Denote the true data generating distribution as $p_0$. The definitions of oracle consistency, oracle restrictiveness, and disentanglement are then given in Appendix \ref{sec:disentanglement}.

Consider our model class $\{(p_., q_.) \mid p_.: \mathcal{Z} \times \mathcal{T} \to [0, 1]^{\Sigma_\mathcal{Y}}, \; q_.: \mathcal{Y} \times \mathcal{T} \times \mathcal{X} \to [0, 1]^{\Sigma_\mathcal{Z}}\}$ and let $S := (Z, T)$, $V := (Y, T)$. We regard $T$ as part of the model outputs such that the oracle operations above are well-defined, i.e. let $\mathcal{H} = \{ (\tilde{p}_., \tilde{q}_.) \mid \tilde{p}_.(s) = p_.(Y | z, t) \times \delta_t(T), \; \tilde{q}_.(v, x) = q_.(Z | y, t, x) \times \delta_t(T) \}$ where $\delta_t$ is the Dirac measure ($\delta_t(T)=1$ if $T=t$ else $0$). Then we have the following lemma:

\begin{lemma}
    \label{lemma:disentanglement}
    If $q$ minimizes the latent divergence in Equation \ref{eq-elbo} over $p_0(D)$, i.e.
    \begin{align}
    \label{eq-identifiability-1}
        \mathbb E_{p_0} D_\mathrm{KL} \left[ q (z | y, t, x) \parallel q (z | y', t', x) \right] = 0,
    \end{align}
    then $\tilde{q}$ disentangles $Z$ and $T$.
\end{lemma}

Lemma \ref{lemma:disentanglement} builds a direct connection between the latent divergence term and disentangled abduction of exogenous noise. Intuitively, it states that if the encoder encodes the same latent representation for any pair of potential outcomes of the same individual, then the latent must have been disentangled from the treatment. However, does optimizing the ELBO necessarily imply that the latent divergence term reaches minimum? We proceed to show that this can be implied under mild assumptions.

\begin{proposition}
    \label{thm:identifiability-1}
    With the true distribution $p_0$ satisfying Assumption \ref{assumption-consistency} in Appendix \ref{sec:causal-assumptions} and Assumption \ref{assumption-uniqueness} in Appendix \ref{proof:identifiability-1}, if $(p_*, q_*)$ maximizes the evidence lower bound over $p_0(D)$ and any $\hat{p}$, i.e.
    \begin{align}
    \label{eq-identifiability-2}
        (p_*, q_*) \in \operatorname*{arg\,max}_{p_., q_.} &\mathbb E_{p_0} \left\{\mathbb E_{q_. (z | y, t, x)} \log \left[ p_. (y | z, t) \right] + \log \left[ \hat{p} (y' | x, t') \right] \right. \nonumber \\
        &\left. - D_\mathrm{KL} \left[ q_. (z | y, t, x) \parallel q_. (z | y', t', x) \right] \right\}
    \end{align}
    then $\tilde{q}_*$ disentangles $Z$ and $T$.
\end{proposition}

Intuitively, Proposition \ref{thm:identifiability-1} states that as long as the treatments do not have any unknown side effect and do not completely wipe out different features in different individuals, the solution to maximizing the ELBO disentangles $Z$ and $T$. In practice, we can formulate model class as deterministic functions (i.e. degenerate distributions) plus noise similar to prior work, and encoder noise as standard normal:
\begin{align}
    \left\{ p_. \right. & \left. \mid g_.(Y | z, t) := \delta_{\mathbb E_{p_.(\cdot | z, t)} Y}, \; \epsilon_.(z, t) := p_.(Y - \mathbb E_{p_.(\cdot | z, t)} Y | z, t) \right\} \label{p_.-practice}\\
    \left\{q_. \right. & \left. \mid e_.(Z | y, t, x) := \delta_{\mathbb E_{q_.(\cdot | y, t, x)} Z}, \; q_.(Z - \mathbb E_{q_.(\cdot | y, t, x)} Z | y, t, x) = \mathcal{N}(0, 1) \right\} \label{q_.-practice}
\end{align}
and learn model $e_.$, $g_.$ and $\epsilon_.$ during training. The following proposition states that disentanglement under the estimating model's own oracle can be achieved under this setting by optimizing the VCI objective in real-world optimizations, where neither the true generating mechanisms nor the true factual-counterfactual pairs are available.

\begin{proposition}
    \label{thm:identifiability-2}
    Given dataset $\{(x_k, t_k, y_k)\} |_{k=1}^n$, let $p_\mathrm{data}$ be its empirical distribution and the model class in Equation \ref{p_.-practice} and Equation \ref{q_.-practice} satisfy the following uniqueness conditions:
    \begin{enumerate}
        \item[1)] $\forall \alpha \in \mathcal{T}$, we have $g_.(Y | z, \alpha) \neq g_.(Y | z', \alpha)$ if $z \neq z'$;
        \item[2)] $\forall x, t, y, t': p_\mathrm{data}(x, t, y) \cdot p_\mathrm{data}(t' | x) > 0$ and $y'_. = \mathbb E_{g_.(\cdot | z, t') e_. (z | y, t, x)} Y$, we have $p_\mathrm{data}(x, t', y'_.) = 0$ if $(x, t, y) \neq (x, t', y'_.)$.
    \end{enumerate}
    If $(p_*, q_*)$ maximizes the VCI objective in Equation \ref{VCI-objective} over $p_\mathrm{data}$ and any $\hat{p}$, i.e.
    \begin{align}
    \label{eq-identifiability-3}
        (p_*, q_*) \in &\operatorname*{arg\,max}_{p_., q_.} J_{p_\mathrm{data}}(p_., q_.)
    \end{align}
    then $(\tilde{g}_*, \tilde{q}_*)$ disentangles $Z$ and $T$ under $\int_y e_* p_\mathrm{data}$.
\end{proposition}

We note that uniqueness condition 2) in Proposition \ref{thm:identifiability-2} is reasonable in high-dimensional outcome settings: taking the facial imaging dataset as an example, it wouldn't make sense for a model's constructed image for a certain individual to be an exact match of the factual outcome of another individual in the dataset
, as long as the dataset does not contain multiple images of the same individual in the same environment
.

\begin{corollary}
    \label{thm:identifiability-2.2}
    Under the same settings and conditions of Proposition \ref{thm:identifiability-2}, $(\tilde{g}_*, \tilde{e}_*)$ disentangles $Z$ and $T$ under $\int_y e_* p_\mathrm{data}$.
\end{corollary}

Corollary \ref{thm:identifiability-2.2} states that the optimal model is guaranteed to disentangle $Z$ and $T$ under the constructed latent distribution in inference mode without noise injection. In summary, the disentanglement results in our framework rely on the simple fact that the latent divergence term is established on an individual level, and does not necessarily compete with outcome supervisions under mild assumptions. Prior works in VAEs have proposed bounding latent on conditional marginal prior \citep{sohn2015learning, khemakhem2020variational}, and VCI can be seen as a further relaxed form of latent restriction in the sense that the latent for any given subject need not be restricted to any prior distribution -- we only require the latent distributions of a subject's different potential outcomes to match each other. More insights on how the latent divergence term helps the training of VCI in practice can be found in the ablation study in the experiments section.

\section{Experiments}
\label{experiments}

We present experiment results of our framework on two datasets with vector outcomes (sci-Plex dataset from \citet{srivatsan2020massively} (Sciplex) and the CRISPRa dataset from \citet{schmidt2022crispr} (Marson)) and two datasets with image outcomes: Morpho-MNIST \citep{castro2019morpho} and CelebA-HQ \citep{karras2017progressive}. Results from the former are compared to state-of-the-art models in single-cell perturbation prediction and results from the latter are compared to state-of-the-art models in counterfactual image generation. In both cases, our model exhibited superior performance against benchmarks. Details of model and dataset settings for each experiment can be found in Appendix \ref{setting:experiment}.

\subsection{Single-cell Perturbation Datasets}
\label{experiment:single-cell-pert}

Same as \citet{lotfollahi2021learning}, we evaluate our model and benchmarks on a widely accepted and biologically meaningful metric --- the $R^2$ (coefficient of determination) of the average prediction against the true average from the out-of-distribution (OOD) set on all genes and differentially-expressed (DE) genes. Definitions of OOD set and DE genes can be found in Appendix \ref{setting:single-cell-pert}. Results over five independent runs are shown in Table \ref{result-table1}.

\begin{table}[ht!]
  \caption{$\bar{R^2}$ of OOD predictions on single-cell perturbation datasets. AE is a naive baseline adapting Autoencoder to counterfactual generation, see Appendix \ref{ae-adaptation}. GANITE is GANITE's counterfactual block adapting to high-dimensional outcome plus multi-level treatment (see Appendix \ref{ganite-adaptation}).}
  \label{result-table1}
  \centering
  \small 
  \begin{tabular}{lcccc}
    \toprule
    & \multicolumn{2}{c}{Sciplex} & \multicolumn{2}{c}{Marson} \\
    \cmidrule(r){2-3} \cmidrule(r){4-5}
    & all genes & DE genes & all genes & DE genes \\
    \midrule
    AE & 0.740 $\pm$ 0.043 & 0.421 $\pm$ 0.021 & 0.804 $\pm$ 0.020 & 0.448 $\pm$ 0.009 \\ 
    CEVAE \citep{louizos2017causal} & 0.760 $\pm$ 0.019 & 0.436 $\pm$ 0.014 & 0.795 $\pm$ 0.014 & 0.424 $\pm$ 0.015 \\ 
    GANITE \citep{yoon2018ganite} & 0.751 $\pm$ 0.013 & 0.417 $\pm$ 0.014 & 0.795 $\pm$ 0.017 & 0.443 $\pm$ 0.025 \\ 
    CPA \citep{lotfollahi2021learning} & \textbf{0.836} $\pm$ 0.002 & 0.474 $\pm$ 0.014 & 0.876 $\pm$ 0.005 & 0.549 $\pm$ 0.019 \\
    \hdashline
    VCI & 0.832 $\pm$ 0.008 & \textbf{0.496} $\pm$ 0.011 & \textbf{0.891} $\pm$ 0.007 & \textbf{0.658} $\pm$ 0.040 \\
    \bottomrule
  \end{tabular}
\end{table}

For each perturbation of each covariate level (e.g. each cell type of each donor) in the OOD set, the $R^2$ (coefficient of determination) is computed with the average outcome predictions for all genes and DE genes using samples from the validation set against the true empirical average over samples from the OOD set. The average $R^2$ over all perturbations of all covariate levels is then calculated as the evaluation metric $\bar{R^2}$. In these experiments, our variational Bayesian causal inference framework excelled state-of-the-art models in both experiments, with the largest fractional improvement on DE genes which are most causally affected by the perturbations. \textbf{Results of marginal estimation using the robust estimator can be found in Appendix \ref{experiment:marginal-est}.}

\subsection{Morpho-MNIST}

\begingroup

\setlength{\tabcolsep}{5pt} 

\begin{table}
  \caption{The errors between counterfactual predictions and counterfactual truths on Morpho-MNIST. 
  Images are scaled to $[0, 1]$ in evaluations. On top of randomly sampling counterfactual treatment, we also randomly sample the ratio of modification, which could be any value such that the thickness and intensity of the counterfactual truth is in range. \textbf{The standard error of runs and the violin plot of errors are reported in Appendix \ref{experiment:morpho-mnist-std}.}}
  \label{result-table-morpho}
  \centering
  \resizebox{\textwidth}{!}{
      \begin{tabular}{lcccccccccc}
        \toprule
        & & \multicolumn{3}{c}{Image MSE $\downarrow$ ($\cdot 10^{-2}$)} & \multicolumn{3}{c}{Thickness (th) MAE $\downarrow$} & \multicolumn{3}{c}{Intensity (in) MAE $\downarrow$ ($\cdot 10^{-1}$)} \\
        \cmidrule(r){3-5} \cmidrule(r){6-8} \cmidrule(r){9-11}
        & $\beta$ & $do($th$)$ & $do($in$)$ & mix & $do($th$)$ & $do($in$)$ & mix & $do($th$)$ & $do($in$)$ & mix \\
        \midrule
        DEAR \citep{shen2022weakly}  & & 6.93 & 3.21 & 5.08 & 0.82 & 0.74 & 0.85 & 6.23 & 3.39 & 4.61 \\
        Diff-SCM \citep{sanchez2022diffusion}  & & 7.72 & 4.90 & 6.30 & 0.70 & 0.76 & 0.73 & 3.48 & 2.17 & 2.80 \\
        CHVAE \citep{monteiro2023measuring}  & 1 & 4.92 & 4.81 & 4.81 & 0.61 & 0.66 & 0.62 & 3.23 & 1.88 & 2.81 \\
        CHVAE \citep{monteiro2023measuring}  & 3 & 6.93 & 6.19 & 6.40 & 0.71 & 1.34 & 0.97 & 2.37 & 1.31 & 1.89 \\
        MED \citep{ribeiro2023high}  & 1 & 2.76 & 0.54 & 1.65 & 0.54 & 0.35 & 0.46 & 1.17 & 0.43 & 0.78 \\
        MED \citep{ribeiro2023high}  & 3 & 2.20 & 0.61 & 1.39 & 0.31 & 0.29 & 0.33 & 0.44 & 0.34 & 0.40 \\
        \hdashline
        SAE (Section \ref{semi-autoencode})  & & 0.67 & 0.20 & 0.47 & 0.32 & 0.21 & 0.29 & 0.52 & 0.26 & 0.41 \\
        VCI (Section \ref{variational-causal-inference})  & & \textbf{0.42} & \textbf{0.13} & \textbf{0.36} & \textbf{0.20} & \textbf{0.14} & \textbf{0.22} & \textbf{0.42} & \textbf{0.22} & \textbf{0.33}  \\
        \bottomrule
      \end{tabular}
  }
\end{table}

\endgroup

In an ideal world, we would like to evaluate model performance against benchmarks simply by measuring the error between the counterfactual prediction and the counterfactual truth for each individual. When the counterfactual truth is attainable, there is no need to resort to approximation metrics such as \citet{monteiro2023measuring, melistas2024benchmarking}. For this reason, we specifically chose the Morpho-MNIST \citep{castro2019morpho} dataset to present our main evaluation results because 
the counterfactual truth can be directly computed on this dataset based on the intervention (modifying thickness or modifying intensity of hand-written digits) even with the existence of exogenous noise. Similar to \citet{ribeiro2023high}, we evaluate counterfactual constructions under single modifications as well as mix of modifications, and compare them to state-of-the-art models on high-fidelity image counterfactual generation. Contrary to \citet{ribeiro2023high}, the magnitude of modification is randomly sampled, which makes the task significantly harder. The results are shown in Table \ref{result-table-morpho} (standard error in Table \ref{result-table-morpho-std}) which demonstrate that ours beat state-of-the-arts by a wide margin. To understand why conditional diffusion model such as Diff-SCM \citep{sanchez2022diffusion} does not perform well in counterfactual generative modeling, see Appendix \ref{sec:discussion-diffusion}. Note that \citet{ribeiro2023high} only evaluates mean absolute error (MAE) on the thickness and intensity of counterfactual constructions, which took treatment characteristics into account but completely left out how much individuality of the original images has been preserved, whereas mean squared error (MSE) on the image is a comprehensive metric that takes both factors into account and should be the primary evaluation metric when the counterfactual truth is available. \textbf{A model inspection using axiomatic soundness metrics \citep{monteiro2023measuring} can be found in Appendix \ref{experiment:c-e-r}.}

To investigate the impact of counterfactual supervision and latent divergence terms in optimizations, we present an ablation study below to compare VCI to SAE and hierarchical autoencoder (HAE) with the same neural architecture.

\begin{figure}[ht!]
    \centering
    \includegraphics[width=\linewidth]{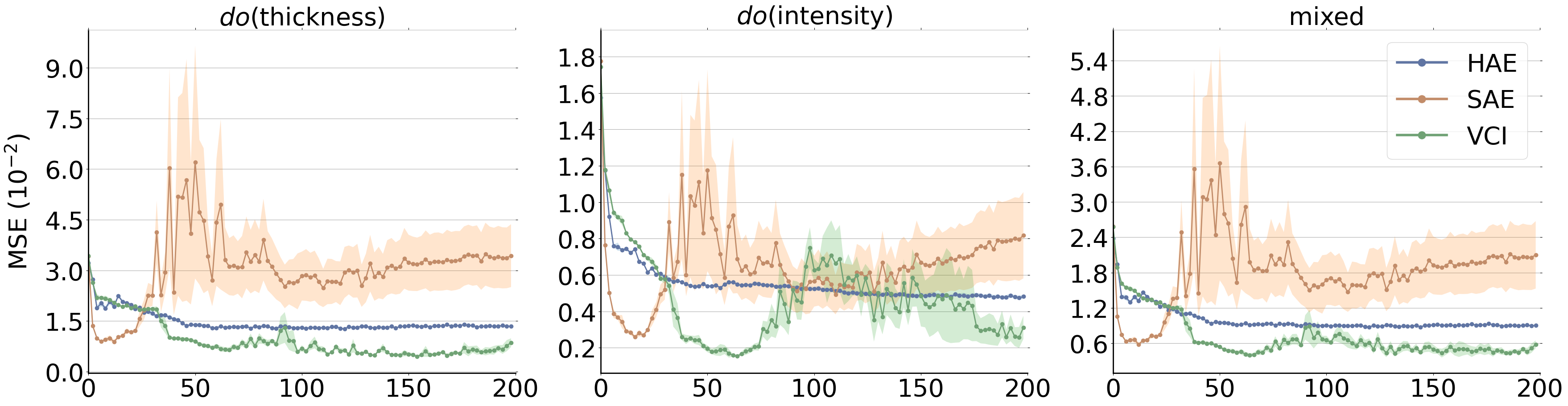}

    \caption{Ablation Study: the error of counterfactual prediction across epochs during the training of HAE (VCI without counterfactual supervision and latent divergence), SAE (VCI without latent divergence) and VCI over five independent runs. Note that VCI with latent divergence but without counterfactual supervision does not make logical sense, but for the completeness of the ablation study, we present the results for such setting in Appendix \ref{experiment:complete-ablation}.}
  \label{fig:morpho-ablation}
\end{figure}

\begin{figure}[ht!]
    \centering
    \begin{subfigure}{0.38\linewidth}
        \centering
        \begin{subfigure}{0.88\linewidth}
            \centering
            \includegraphics[width=\linewidth]{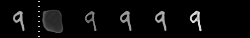}
        \end{subfigure}
    
        \begin{subfigure}{0.88\linewidth}
            \centering
            \includegraphics[width=\linewidth]{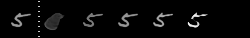}
        \end{subfigure} 
    
        \begin{subfigure}{0.88\linewidth}
            \centering
            \includegraphics[width=\linewidth]{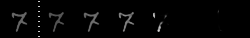}
        \end{subfigure} 
    
        \begin{subfigure}{0.88\linewidth}
            \centering
            \includegraphics[width=\linewidth]{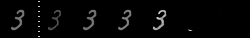}
        \end{subfigure} 
    
        \begin{subfigure}{0.88\linewidth}
            \centering
            \includegraphics[width=\linewidth]{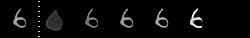}
        \end{subfigure} 
    
    \caption{SAE}
    \label{fig:morphoMNIST-ablation-sae}
    \end{subfigure}
    \begin{subfigure}{0.38\linewidth}
        \centering
        \begin{subfigure}{0.88\linewidth}
            \centering
            \includegraphics[width=\linewidth]{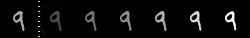}
        \end{subfigure}
    
        \begin{subfigure}{0.88\linewidth}
            \centering
            \includegraphics[width=\linewidth]{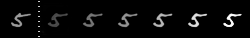}
        \end{subfigure} 
    
        \begin{subfigure}{0.88\linewidth}
            \centering
            \includegraphics[width=\linewidth]{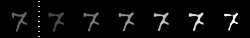}
        \end{subfigure} 
    
        \begin{subfigure}{0.88\linewidth}
            \centering
            \includegraphics[width=\linewidth]{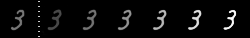}
        \end{subfigure} 
    
        \begin{subfigure}{0.88\linewidth}
            \centering
            \includegraphics[width=\linewidth]{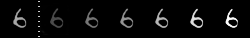}
        \end{subfigure} 
    
    \caption{VCI}
    \label{fig:morphoMNIST-ablation-vci}
    \end{subfigure}
    
    \caption{Illustration of the latent divergence term's long-term impact. Samples are drawn from the last epoch of SAE and VCI on $do$(intensity). The left most image of each set is the original image. \textbf{More sampled results from VCI on $do$(thickness) and $do$(intensity) can be found in Figure \ref{fig:morpho-mnist-sample}.} \textbf{Results from intervening digit i.e. $do$(digit) can be found in Figure \ref{fig:morpho-mnist-sample-digit}.}}
    \label{fig:morpho-ablation-sample}
\end{figure}

As can be seen in Figure \ref{fig:morpho-ablation}, incorporating counterfactual supervision alone could greatly accelerate optimization in the initial epochs of training, and achieves an ideal model state in the early stage. This is the same phenomenon we observed in the optimization on single-cell perturbation datasets with respect to $\bar{R^2}$. 
However, although both SAE and VCI beat HAE convincingly in terms of best result, SAE is largely unstable in the long-run without the proposed latent divergence restriction, and the full VCI optimization scheme consistently achieves better performance in terms of best result and final result. To further help the readers understand how and why the latent divergence term stabilizes VCI training, we include a discussion below along with some illustrations on $do$(intensity) in Figure \ref{fig:morpho-ablation-sample}.

In a GAN training scheme, the discriminator is a neural model that could often struggle with out-of-distribution evaluations. As can be seen from the generated samples by SAE in Figure \ref{fig:morpho-ablation-sample}, when the intervention labels are close to the original label, the counterfactual generations are reasonable. For intervention labels that are significantly different from the original, the images become nonsensical. This is because the combination of an intervention label that is far from the original and an image similar to the original is highly out-of-distribution and rarely present in the observed dataset. In other words, the discriminator has never encountered an image with the same digit and style as the original having such extreme intensity variations, so whatever the generator produces, the discriminator cannot tell if it is real or fake due to the lack of reference. This results in the generator exploiting the neural discriminator in these out-of-distribution scenarios in the long run. However, the latent divergence term prevents the generator from producing these nonsensical images because, otherwise, the encoder would not be able to recover a latent distribution close to that of the original image. \textbf{An evaluation of latent disentanglement under the latent divergence term can be found in Appendix \ref{experiment:disentanglement-evaluation}.}

\subsection{CelebA-HQ}
\label{sec:exp-celeba-hq}

\begin{figure}[ht]
    \centering
    \begin{subfigure}{0.96\linewidth}
        \centering
        \begin{subfigure}{.24\linewidth}
            \centering
            \includegraphics[width=\linewidth]{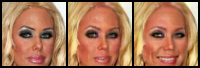}
        \end{subfigure}
        \hfill
        \begin{subfigure}{.24\linewidth}
            \centering
            \includegraphics[width=\linewidth]{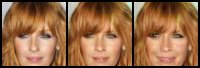}
        \end{subfigure}
        \hfill
        \begin{subfigure}{.24\linewidth}
            \centering
            \includegraphics[width=\linewidth]{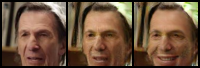}
        \end{subfigure}
        \hfill
        \begin{subfigure}{.24\linewidth}
            \centering
            \includegraphics[width=\linewidth]{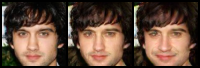}
        \end{subfigure}
    
        \begin{subfigure}{.24\linewidth}
            \centering
            \includegraphics[width=\linewidth]{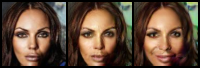}
        \end{subfigure}
        \hfill
        \begin{subfigure}{.24\linewidth}
            \centering
            \includegraphics[width=\linewidth]{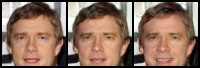}
        \end{subfigure}
        \hfill
        \begin{subfigure}{.24\linewidth}
            \centering
            \includegraphics[width=\linewidth]{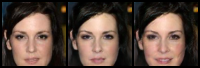}
        \end{subfigure}
        \hfill
        \begin{subfigure}{.24\linewidth}
            \centering
            \includegraphics[width=\linewidth]{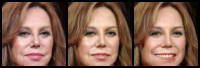}
        \end{subfigure}
    
        \caption{Add smiling. Left: original, middle: not smiling (reconstruction), right: smiling (counterfactual).}
        \label{fig:add-smiling}
    \end{subfigure}

    \begin{subfigure}{0.96\linewidth}
        \centering
        \begin{subfigure}{.24\linewidth}
            \centering
            \includegraphics[width=\linewidth]{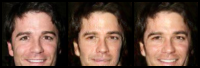}
        \end{subfigure}
        \hfill
        \begin{subfigure}[b]{.24\linewidth}
            \centering
            \includegraphics[width=\linewidth]{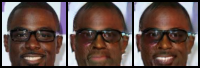}
        \end{subfigure}
        \hfill
        \begin{subfigure}[b]{.24\linewidth}
            \centering
            \includegraphics[width=\linewidth]{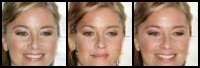}
        \end{subfigure}
        \hfill
        \begin{subfigure}{.24\linewidth}
            \centering
            \includegraphics[width=\linewidth]{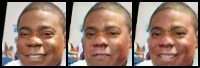}
        \end{subfigure}
    
        \begin{subfigure}{.24\linewidth}
            \centering
            \includegraphics[width=\linewidth]{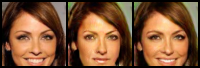}
        \end{subfigure}
        \hfill
        \begin{subfigure}[b]{.24\linewidth}
            \centering
            \includegraphics[width=\linewidth]{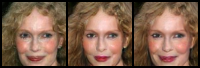}
        \end{subfigure}
        \hfill
        \begin{subfigure}[b]{.24\linewidth}
            \centering
            \includegraphics[width=\linewidth]{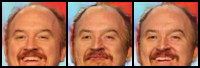}
        \end{subfigure}
        \hfill
        \begin{subfigure}[b]{.24\linewidth}
            \centering
            \includegraphics[width=\linewidth]{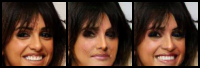}
        \end{subfigure}
    
        \caption{Remove smiling. Left: original, middle: not smiling (counterfactual), right: smiling (reconstruction).}
        \label{fig:remove-smiling}
    \end{subfigure}

    \begin{subfigure}{0.96\linewidth}
        \centering
        \begin{subfigure}{.24\linewidth}
            \centering
            \includegraphics[width=\linewidth]{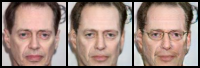}
        \end{subfigure}
        \hfill
        \begin{subfigure}{.24\linewidth}
            \centering
            \includegraphics[width=\linewidth]{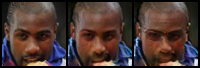}
        \end{subfigure}
        \hfill
        \begin{subfigure}{.24\linewidth}
            \centering
            \includegraphics[width=\linewidth]{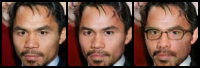}
        \end{subfigure}
        \hfill
        \begin{subfigure}{.24\linewidth}
            \centering
            \includegraphics[width=\linewidth]{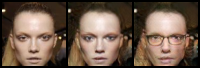}
        \end{subfigure}
    
        \begin{subfigure}{.24\linewidth}
            \centering
            \includegraphics[width=\linewidth]{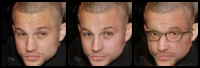}
        \end{subfigure}
        \hfill
        \begin{subfigure}{.24\linewidth}
            \centering
            \includegraphics[width=\linewidth]{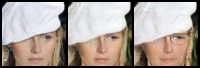}
        \end{subfigure}
        \hfill
        \begin{subfigure}{.24\linewidth}
            \centering
            \includegraphics[width=\linewidth]{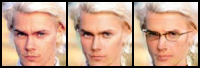}
        \end{subfigure}
        \hfill
        \begin{subfigure}{.24\linewidth}
            \centering
            \includegraphics[width=\linewidth]{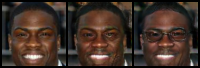}
        \end{subfigure}
    
        \caption{Add glasses. Left: original, middle: no glasses (reconstruction), right: with glasses (counterfactual).}
        \label{fig:add-glasses}
    \end{subfigure}

    \begin{subfigure}{0.96\linewidth}
        \centering
        \begin{subfigure}{.24\linewidth}
            \centering
            \includegraphics[width=\linewidth]{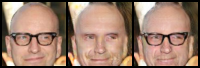}
        \end{subfigure}
        \hfill
        \begin{subfigure}{.24\linewidth}
            \centering
            \includegraphics[width=\linewidth]{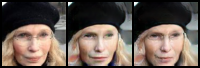}
        \end{subfigure}
        \hfill
        \begin{subfigure}{.24\linewidth}
            \centering
            \includegraphics[width=\linewidth]{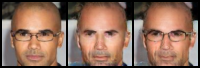}
        \end{subfigure}
        \hfill
        \begin{subfigure}{.24\linewidth}
            \centering
            \includegraphics[width=\linewidth]{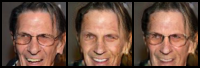}
        \end{subfigure}
        
        \begin{subfigure}{.24\linewidth}
            \centering
            \includegraphics[width=\linewidth]{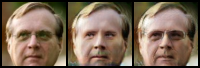}
        \end{subfigure}
        \hfill
        \begin{subfigure}{.24\linewidth}
            \centering
            \includegraphics[width=\linewidth]{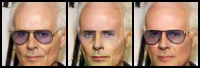}
        \end{subfigure}
        \hfill
        \begin{subfigure}{.24\linewidth}
            \centering
            \includegraphics[width=\linewidth]{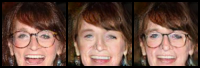}
        \end{subfigure}
        \hfill
        \begin{subfigure}{.24\linewidth}
            \centering
            \includegraphics[width=\linewidth]{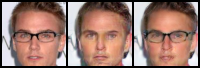}
        \end{subfigure}
    
        \caption{Remove glasses. Left: original, middle: no glasses (counterfactual), right: with glasses (reconstruction).}
        \label{fig:remove-glasses}
    \end{subfigure}
    
    \caption{Results on the test set of CelebA-HQ.}
    \label{fig:celeba-hq}
\end{figure}

For any real-world dataset where the counterfactual truth is not available, there is not a definitive metric to evaluate how good the counterfactual constructions really are. For that reason, facial imaging datasets are prevailing benchmarks for examining counterfactual goodness because even in the lack of a quantitative metric, human can judge the quality of counterfactuals 
without the need for any domain knowledge. Same as \citet{monteiro2023measuring}, we use the CelebA-HQ \citep{karras2017progressive} dataset on 64$\times$64 resolution to evaluate the model's capability of counterfactual constructions on two factors -- smiling and glasses. Some sampled results from our model are shown in Figure \ref{fig:celeba-hq}.

While MSE of the counterfactual construction cannot be measured, readers can empirically compare ours to prior work such as CHVAE \citep{monteiro2023measuring} which serves as the backbone for the state-of-the-art model in high fidelity counterfactual image generation MED \citep{ribeiro2023high}. Aside from compiling counterfactual credibility across all generation tasks, ours show a strong capability of disentangling factual treatment from observed images in counterfactual construction, which can be particularly observed in the task of ``removing glasses'' as shown in \Cref{fig:remove-glasses}. 
Note that contrary to prior works, we construct and supervise counterfactual outcomes during training, and the reconstruction and counterfactual construction share the same decoding mechanism. Hence, ``reconstruction'' is really counterfactual construction with the factual label. This can be most clearly observed in the third set of comparison in line 2 of \Cref{fig:add-smiling}, where the original image has a label of ``not smiling'' yet the person is really in-between smiling and not smiling. In this case, the ``reconstruction'' is really a counterfactual construction with treatment abiding to the original label. 
Lastly, we include a discussion here regarding counterfactual fairness. Note that adding glasses is a relatively easy engineering task, but relatively hard deep learning task on CelebA-HQ due to the small sample size of images with glasses. 
As can be seen from \Cref{fig:add-glasses}, our model learned to not only attach glasses at the right position, but also attach glasses according to a person's style and context. However, readers may naturally wonder if this could induce any fairness issue, such as attaching certain type of glasses more frequently on certain demographic. We note that our framework can reduce spurious or unintended relations between given factors, as long as practitioners measure these factors during data collection and include them as observed labels for the composed datasets, since the learnt latent representations of our framework are encouraged to be disentangled from the observed factors. In the case of CelebA, style of glasses is unmeasured, and the model cannot reasonably tell if attaching different types of glasses on different demographics is unintended or not.

\section{Conclusion}
\label{conclusion}

In this work, we introduced a variational Bayesian causal inference framework for estimating high-dimensional counterfactual outcomes, as well as consequent robust marginal estimators. With this framework, treatment characteristics and individuality in the predicted outcomes can be explicitly balanced and optimized, and learnt latent representations are disentangled from the treatments. 
As for limitations, the theoretical results of this work are established upon the common causal assumption that there exists no unobserved confounders, through which causal effects are identifiable. Besides, when multiple treatments are concerned, this work does not explicitly address the situation where a causal relation exists between treatments. Thus, a straightforward extension of this work could be to incorporate explicit modeling of the causal relation among treatment variables and update the descendants of given counterfactual treatment, similar to \citet{pawlowski2020deep}, before feeding them into our counterfactual outcome construction framework.

\subsubsection*{Author Contributions}
Yulun Wu developed the proposed framework, conducted experiments on single-cell perturbation datasets, Morpho-MNIST and CelebA-HQ, and wrote the manuscript; Louie McConnell conducted experiments on Morpho-MNIST; Claudia Iriondo reviewed related work and edited the manuscript.

\subsubsection*{Acknowledgments}
The authors thank Cian Eastwood, Toru Shirakawa, Pieter Abbeel, Jason Hartford, Dominik Janzing, Ivana Malenica, Ahmed M. Alaa, and David E. Heckerman for the insightful discussions. The authors thank Layne C. Price, Zichen Wang, Robert A. Barton, Vassilis N. Ioannidis, and George Karypis for their feedback in the early development.

\bibliography{iclr2025_conference}

\begin{thebibliography}{55}
\providecommand{\natexlab}[1]{#1}
\providecommand{\url}[1]{\texttt{#1}}
\expandafter\ifx\csname urlstyle\endcsname\relax
  \providecommand{\doi}[1]{doi: #1}\else
  \providecommand{\doi}{doi: \begingroup \urlstyle{rm}\Url}\fi

\bibitem[Balke \& Pearl(1994)Balke and Pearl]{balke2022probabilistic}
Alexander Balke and Judea Pearl.
\newblock Probabilistic evaluation of counterfactual queries.
\newblock In \emph{Proceedings of the AAAI conference on artificial intelligence}, volume~12, 1994.

\bibitem[Bengio et~al.(2014)Bengio, Laufer, Alain, and Yosinski]{bengio2014deep}
Yoshua Bengio, Eric Laufer, Guillaume Alain, and Jason Yosinski.
\newblock Deep generative stochastic networks trainable by backprop.
\newblock In \emph{International Conference on Machine Learning}, pp.\  226--234. PMLR, 2014.

\bibitem[Brehmer et~al.(2022)Brehmer, De~Haan, Lippe, and Cohen]{brehmer2022weakly}
Johann Brehmer, Pim De~Haan, Phillip Lippe, and Taco~S Cohen.
\newblock Weakly supervised causal representation learning.
\newblock \emph{Advances in Neural Information Processing Systems}, 35:\penalty0 38319--38331, 2022.

\bibitem[Castro et~al.(2019)Castro, Tan, Kainz, Konukoglu, and Glocker]{castro2019morpho}
Daniel~C Castro, Jeremy Tan, Bernhard Kainz, Ender Konukoglu, and Ben Glocker.
\newblock Morpho-mnist: quantitative assessment and diagnostics for representation learning.
\newblock \emph{Journal of Machine Learning Research}, 20\penalty0 (178):\penalty0 1--29, 2019.

\bibitem[Chen et~al.(2016)Chen, Duan, Houthooft, Schulman, Sutskever, and Abbeel]{chen2016infogan}
Xi~Chen, Yan Duan, Rein Houthooft, John Schulman, Ilya Sutskever, and Pieter Abbeel.
\newblock Infogan: Interpretable representation learning by information maximizing generative adversarial nets.
\newblock \emph{Advances in neural information processing systems}, 29, 2016.

\bibitem[Child(2020)]{child2020very}
Rewon Child.
\newblock Very deep vaes generalize autoregressive models and can outperform them on images.
\newblock \emph{arXiv preprint arXiv:2011.10650}, 2020.

\bibitem[Dash et~al.(2022)Dash, Balasubramanian, and Sharma]{dash2022evaluating}
Saloni Dash, Vineeth~N Balasubramanian, and Amit Sharma.
\newblock Evaluating and mitigating bias in image classifiers: A causal perspective using counterfactuals.
\newblock In \emph{Proceedings of the IEEE/CVF Winter Conference on Applications of Computer Vision}, pp.\  915--924, 2022.

\bibitem[Dixit et~al.(2016)Dixit, Parnas, Li, Chen, Fulco, Jerby-Arnon, Marjanovic, Dionne, Burks, Raychowdhury, et~al.]{dixit2016perturb}
Atray Dixit, Oren Parnas, Biyu Li, Jenny Chen, Charles~P Fulco, Livnat Jerby-Arnon, Nemanja~D Marjanovic, Danielle Dionne, Tyler Burks, Raktima Raychowdhury, et~al.
\newblock Perturb-seq: dissecting molecular circuits with scalable single-cell rna profiling of pooled genetic screens.
\newblock \emph{cell}, 167\penalty0 (7):\penalty0 1853--1866, 2016.

\bibitem[Feng et~al.(2022)Feng, Xiao, Zheng, Zhao, Zhou, Sun, and Zha]{feng2022principled}
Ruili Feng, Jie Xiao, Kecheng Zheng, Deli Zhao, Jingren Zhou, Qibin Sun, and Zheng-Jun Zha.
\newblock Principled knowledge extrapolation with gans.
\newblock In \emph{International Conference on Machine Learning}, pp.\  6447--6464. PMLR, 2022.

\bibitem[Goodfellow et~al.(2014)Goodfellow, Pouget-Abadie, Mirza, Xu, Warde-Farley, Ozair, Courville, and Bengio]{goodfellow2014generative}
Ian Goodfellow, Jean Pouget-Abadie, Mehdi Mirza, Bing Xu, David Warde-Farley, Sherjil Ozair, Aaron Courville, and Yoshua Bengio.
\newblock Generative adversarial nets.
\newblock \emph{Advances in neural information processing systems}, 27, 2014.

\bibitem[Gresele et~al.(2021)Gresele, Von~K{\"u}gelgen, Stimper, Sch{\"o}lkopf, and Besserve]{gresele2021independent}
Luigi Gresele, Julius Von~K{\"u}gelgen, Vincent Stimper, Bernhard Sch{\"o}lkopf, and Michel Besserve.
\newblock Independent mechanism analysis, a new concept?
\newblock \emph{Advances in neural information processing systems}, 34:\penalty0 28233--28248, 2021.

\bibitem[H{\"a}lv{\"a} et~al.(2021)H{\"a}lv{\"a}, Le~Corff, Leh{\'e}ricy, So, Zhu, Gassiat, and Hyvarinen]{halva2021disentangling}
Hermanni H{\"a}lv{\"a}, Sylvain Le~Corff, Luc Leh{\'e}ricy, Jonathan So, Yongjie Zhu, Elisabeth Gassiat, and Aapo Hyvarinen.
\newblock Disentangling identifiable features from noisy data with structured nonlinear ica.
\newblock \emph{Advances in Neural Information Processing Systems}, 34:\penalty0 1624--1633, 2021.

\bibitem[Ho et~al.(2020)Ho, Jain, and Abbeel]{ho2020denoising}
Jonathan Ho, Ajay Jain, and Pieter Abbeel.
\newblock Denoising diffusion probabilistic models.
\newblock \emph{Advances in neural information processing systems}, 33:\penalty0 6840--6851, 2020.

\bibitem[Hyvarinen et~al.(2019)Hyvarinen, Sasaki, and Turner]{hyvarinen2019nonlinear}
Aapo Hyvarinen, Hiroaki Sasaki, and Richard Turner.
\newblock Nonlinear ica using auxiliary variables and generalized contrastive learning.
\newblock In \emph{The 22nd International Conference on Artificial Intelligence and Statistics}, pp.\  859--868. PMLR, 2019.

\bibitem[Karras et~al.(2017)Karras, Aila, Laine, and Lehtinen]{karras2017progressive}
Tero Karras, Timo Aila, Samuli Laine, and Jaakko Lehtinen.
\newblock Progressive growing of gans for improved quality, stability, and variation.
\newblock \emph{arXiv preprint arXiv:1710.10196}, 2017.

\bibitem[Khemakhem et~al.(2020)Khemakhem, Kingma, Monti, and Hyvarinen]{khemakhem2020variational}
Ilyes Khemakhem, Diederik Kingma, Ricardo Monti, and Aapo Hyvarinen.
\newblock Variational autoencoders and nonlinear ica: A unifying framework.
\newblock In \emph{International Conference on Artificial Intelligence and Statistics}, pp.\  2207--2217. PMLR, 2020.

\bibitem[Kim et~al.(2021)Kim, Shin, Jang, Song, Joo, Kang, and Moon]{kim2021counterfactual}
Hyemi Kim, Seungjae Shin, JoonHo Jang, Kyungwoo Song, Weonyoung Joo, Wanmo Kang, and Il-Chul Moon.
\newblock Counterfactual fairness with disentangled causal effect variational autoencoder.
\newblock In \emph{Proceedings of the AAAI Conference on Artificial Intelligence}, volume~35, pp.\  8128--8136, 2021.

\bibitem[Kingma \& Welling(2013)Kingma and Welling]{kingma2013auto}
Diederik~P Kingma and Max Welling.
\newblock Auto-encoding variational bayes.
\newblock \emph{arXiv preprint arXiv:1312.6114}, 2013.

\bibitem[Kocaoglu et~al.(2017)Kocaoglu, Snyder, Dimakis, and Vishwanath]{kocaoglu2017causalgan}
Murat Kocaoglu, Christopher Snyder, Alexandros~G Dimakis, and Sriram Vishwanath.
\newblock Causalgan: Learning causal implicit generative models with adversarial training.
\newblock \emph{arXiv preprint arXiv:1709.02023}, 2017.

\bibitem[Komanduri et~al.(2023)Komanduri, Wu, Chen, and Wu]{komanduri2023learning}
Aneesh Komanduri, Yongkai Wu, Feng Chen, and Xintao Wu.
\newblock Learning causally disentangled representations via the principle of independent causal mechanisms.
\newblock \emph{arXiv preprint arXiv:2306.01213}, 2023.

\bibitem[Komanduri et~al.(2024)Komanduri, Zhao, Chen, and Wu]{komanduri2024causal}
Aneesh Komanduri, Chen Zhao, Feng Chen, and Xintao Wu.
\newblock Causal diffusion autoencoders: Toward counterfactual generation via diffusion probabilistic models.
\newblock \emph{arXiv preprint arXiv:2404.17735}, 2024.

\bibitem[Lachapelle et~al.(2022)Lachapelle, Rodriguez, Sharma, Everett, Le~Priol, Lacoste, and Lacoste-Julien]{lachapelle2022disentanglement}
S{\'e}bastien Lachapelle, Pau Rodriguez, Yash Sharma, Katie~E Everett, R{\'e}mi Le~Priol, Alexandre Lacoste, and Simon Lacoste-Julien.
\newblock Disentanglement via mechanism sparsity regularization: A new principle for nonlinear ica.
\newblock In \emph{Conference on Causal Learning and Reasoning}, pp.\  428--484. PMLR, 2022.

\bibitem[Levy(2019)]{levy2019tutorial}
Jonathan Levy.
\newblock Tutorial: Deriving the efficient influence curve for large models.
\newblock \emph{arXiv preprint arXiv:1903.01706}, 2019.

\bibitem[Lillicrap et~al.(2015)Lillicrap, Hunt, Pritzel, Heess, Erez, Tassa, Silver, and Wierstra]{lillicrap2015continuous}
Timothy~P Lillicrap, Jonathan~J Hunt, Alexander Pritzel, Nicolas Heess, Tom Erez, Yuval Tassa, David Silver, and Daan Wierstra.
\newblock Continuous control with deep reinforcement learning.
\newblock \emph{arXiv preprint arXiv:1509.02971}, 2015.

\bibitem[Locatello et~al.(2020)Locatello, Poole, R{\"a}tsch, Sch{\"o}lkopf, Bachem, and Tschannen]{locatello2020weakly}
Francesco Locatello, Ben Poole, Gunnar R{\"a}tsch, Bernhard Sch{\"o}lkopf, Olivier Bachem, and Michael Tschannen.
\newblock Weakly-supervised disentanglement without compromises.
\newblock In \emph{International Conference on Machine Learning}, pp.\  6348--6359. PMLR, 2020.

\bibitem[Lotfollahi et~al.(2021)Lotfollahi, Susmelj, De~Donno, Ji, Ibarra, Wolf, Yakubova, Theis, and Lopez-Paz]{lotfollahi2021learning}
Mohammad Lotfollahi, Anna~Klimovskaia Susmelj, Carlo De~Donno, Yuge Ji, Ignacio~L Ibarra, F~Alexander Wolf, Nafissa Yakubova, Fabian~J Theis, and David Lopez-Paz.
\newblock Learning interpretable cellular responses to complex perturbations in high-throughput screens.
\newblock \emph{bioRxiv}, 2021.

\bibitem[Louizos et~al.(2017)Louizos, Shalit, Mooij, Sontag, Zemel, and Welling]{louizos2017causal}
Christos Louizos, Uri Shalit, Joris~M Mooij, David Sontag, Richard Zemel, and Max Welling.
\newblock Causal effect inference with deep latent-variable models.
\newblock \emph{Advances in neural information processing systems}, 30, 2017.

\bibitem[Melistas et~al.(2024)Melistas, Spyrou, Gkouti, Sanchez, Vlontzos, Panagakis, Papanastasiou, and Tsaftaris]{melistas2024benchmarking}
Thomas Melistas, Nikos Spyrou, Nefeli Gkouti, Pedro Sanchez, Athanasios Vlontzos, Yannis Panagakis, Giorgos Papanastasiou, and Sotirios~A Tsaftaris.
\newblock Benchmarking counterfactual image generation.
\newblock \emph{arXiv preprint arXiv:2403.20287}, 2024.

\bibitem[Monteiro et~al.(2023)Monteiro, Ribeiro, Pawlowski, Castro, and Glocker]{monteiro2023measuring}
Miguel Monteiro, Fabio De~Sousa Ribeiro, Nick Pawlowski, Daniel~C Castro, and Ben Glocker.
\newblock Measuring axiomatic soundness of counterfactual image models.
\newblock \emph{arXiv preprint arXiv:2303.01274}, 2023.

\bibitem[Norman et~al.(2019)Norman, Horlbeck, Replogle, Ge, Xu, Jost, Gilbert, and Weissman]{norman2019exploring}
Thomas~M Norman, Max~A Horlbeck, Joseph~M Replogle, Alex~Y Ge, Albert Xu, Marco Jost, Luke~A Gilbert, and Jonathan~S Weissman.
\newblock Exploring genetic interaction manifolds constructed from rich single-cell phenotypes.
\newblock \emph{Science}, 365\penalty0 (6455):\penalty0 786--793, 2019.

\bibitem[Pawlowski et~al.(2020)Pawlowski, Coelho~de Castro, and Glocker]{pawlowski2020deep}
Nick Pawlowski, Daniel Coelho~de Castro, and Ben Glocker.
\newblock Deep structural causal models for tractable counterfactual inference.
\newblock \emph{Advances in Neural Information Processing Systems}, 33:\penalty0 857--869, 2020.

\bibitem[Pearl(1988)]{pearl1988probabilistic}
Judea Pearl.
\newblock \emph{Probabilistic reasoning in intelligent systems: networks of plausible inference}.
\newblock Morgan kaufmann, 1988.

\bibitem[Pearl(1995)]{pearl1995causal}
Judea Pearl.
\newblock Causal diagrams for empirical research.
\newblock \emph{Biometrika}, 82\penalty0 (4):\penalty0 669--688, 1995.

\bibitem[Pearl(2009)]{pearl2009causality}
Judea Pearl.
\newblock \emph{Causality}.
\newblock Cambridge university press, 2009.

\bibitem[Ribeiro et~al.(2023)Ribeiro, Xia, Monteiro, Pawlowski, and Glocker]{ribeiro2023high}
Fabio De~Sousa Ribeiro, Tian Xia, Miguel Monteiro, Nick Pawlowski, and Ben Glocker.
\newblock High fidelity image counterfactuals with probabilistic causal models.
\newblock \emph{arXiv preprint arXiv:2306.15764}, 2023.

\bibitem[Robins(1986)]{robins1986new}
James Robins.
\newblock A new approach to causal inference in mortality studies with a sustained exposure period—application to control of the healthy worker survivor effect.
\newblock \emph{Mathematical modelling}, 7\penalty0 (9-12):\penalty0 1393--1512, 1986.

\bibitem[Rosenbaum \& Rubin(1983)Rosenbaum and Rubin]{rosenbaum1983central}
Paul~R Rosenbaum and Donald~B Rubin.
\newblock The central role of the propensity score in observational studies for causal effects.
\newblock \emph{Biometrika}, 70\penalty0 (1):\penalty0 41--55, 1983.

\bibitem[Sanchez \& Tsaftaris(2022)Sanchez and Tsaftaris]{sanchez2022diffusion}
Pedro Sanchez and Sotirios~A Tsaftaris.
\newblock Diffusion causal models for counterfactual estimation.
\newblock \emph{arXiv preprint arXiv:2202.10166}, 2022.

\bibitem[Sauer \& Geiger(2021)Sauer and Geiger]{sauer2021counterfactual}
Axel Sauer and Andreas Geiger.
\newblock Counterfactual generative networks.
\newblock \emph{arXiv preprint arXiv:2101.06046}, 2021.

\bibitem[Schmidt et~al.(2022)Schmidt, Steinhart, Layeghi, Freimer, Bueno, Nguyen, Blaeschke, Ye, and Marson]{schmidt2022crispr}
Ralf Schmidt, Zachary Steinhart, Madeline Layeghi, Jacob~W Freimer, Raymund Bueno, Vinh~Q Nguyen, Franziska Blaeschke, Chun~Jimmie Ye, and Alexander Marson.
\newblock Crispr activation and interference screens decode stimulation responses in primary human t cells.
\newblock \emph{Science}, 375\penalty0 (6580):\penalty0 eabj4008, 2022.

\bibitem[Schulman et~al.(2017)Schulman, Wolski, Dhariwal, Radford, and Klimov]{schulman2017proximal}
John Schulman, Filip Wolski, Prafulla Dhariwal, Alec Radford, and Oleg Klimov.
\newblock Proximal policy optimization algorithms.
\newblock \emph{arXiv preprint arXiv:1707.06347}, 2017.

\bibitem[Shen et~al.(2022)Shen, Liu, Dong, Lian, Chen, and Zhang]{shen2022weakly}
Xinwei Shen, Furui Liu, Hanze Dong, Qing Lian, Zhitang Chen, and Tong Zhang.
\newblock Weakly supervised disentangled generative causal representation learning.
\newblock \emph{The Journal of Machine Learning Research}, 23\penalty0 (1):\penalty0 10994--11048, 2022.

\bibitem[Shu et~al.(2019)Shu, Chen, Kumar, Ermon, and Poole]{shu2019weakly}
Rui Shu, Yining Chen, Abhishek Kumar, Stefano Ermon, and Ben Poole.
\newblock Weakly supervised disentanglement with guarantees.
\newblock \emph{arXiv preprint arXiv:1910.09772}, 2019.

\bibitem[Sohn et~al.(2015)Sohn, Lee, and Yan]{sohn2015learning}
Kihyuk Sohn, Honglak Lee, and Xinchen Yan.
\newblock Learning structured output representation using deep conditional generative models.
\newblock \emph{Advances in neural information processing systems}, 28, 2015.

\bibitem[Spirtes et~al.(2000)Spirtes, Glymour, and Scheines]{spirtes2000causation}
Peter Spirtes, Clark~N Glymour, and Richard Scheines.
\newblock \emph{Causation, prediction, and search}.
\newblock MIT press, 2000.

\bibitem[Srivatsan et~al.(2020)Srivatsan, McFaline-Figueroa, Ramani, Saunders, Cao, Packer, Pliner, Jackson, Daza, Christiansen, et~al.]{srivatsan2020massively}
Sanjay~R Srivatsan, Jos{\'e}~L McFaline-Figueroa, Vijay Ramani, Lauren Saunders, Junyue Cao, Jonathan Packer, Hannah~A Pliner, Dana~L Jackson, Riza~M Daza, Lena Christiansen, et~al.
\newblock Massively multiplex chemical transcriptomics at single-cell resolution.
\newblock \emph{Science}, 367\penalty0 (6473):\penalty0 45--51, 2020.

\bibitem[Vahdat \& Kautz(2020)Vahdat and Kautz]{vahdat2020nvae}
Arash Vahdat and Jan Kautz.
\newblock Nvae: A deep hierarchical variational autoencoder.
\newblock \emph{Advances in neural information processing systems}, 33:\penalty0 19667--19679, 2020.

\bibitem[Van Der~Laan \& Rubin(2006)Van Der~Laan and Rubin]{van2006targeted}
Mark~J Van Der~Laan and Daniel Rubin.
\newblock Targeted maximum likelihood learning.
\newblock \emph{The international journal of biostatistics}, 2\penalty0 (1), 2006.

\bibitem[Van~der Vaart(2000)]{van2000asymptotic}
Aad~W Van~der Vaart.
\newblock \emph{Asymptotic statistics}, volume~3.
\newblock Cambridge university press, 2000.

\bibitem[Van~Hasselt et~al.(2016)Van~Hasselt, Guez, and Silver]{van2016deep}
Hado Van~Hasselt, Arthur Guez, and David Silver.
\newblock Deep reinforcement learning with double q-learning.
\newblock In \emph{Proceedings of the AAAI conference on artificial intelligence}, volume~30, 2016.

\bibitem[Vincent et~al.(2008)Vincent, Larochelle, Bengio, and Manzagol]{vincent2008extracting}
Pascal Vincent, Hugo Larochelle, Yoshua Bengio, and Pierre-Antoine Manzagol.
\newblock Extracting and composing robust features with denoising autoencoders.
\newblock In \emph{Proceedings of the 25th international conference on Machine learning}, pp.\  1096--1103, 2008.

\bibitem[Vlontzos et~al.(2023)Vlontzos, Kainz, and Gilligan-Lee]{vlontzos2023estimating}
Athanasios Vlontzos, Bernhard Kainz, and Ciar{\'a}n~M Gilligan-Lee.
\newblock Estimating categorical counterfactuals via deep twin networks.
\newblock \emph{Nature Machine Intelligence}, 5\penalty0 (2):\penalty0 159--168, 2023.

\bibitem[von K{\"u}gelgen et~al.(2023)von K{\"u}gelgen, Besserve, Liang, Gresele, Keki{\'c}, Bareinboim, Blei, and Sch{\"o}lkopf]{von2023nonparametric}
Julius von K{\"u}gelgen, Michel Besserve, Wendong Liang, Luigi Gresele, Armin Keki{\'c}, Elias Bareinboim, David~M Blei, and Bernhard Sch{\"o}lkopf.
\newblock Nonparametric identifiability of causal representations from unknown interventions.
\newblock \emph{arXiv preprint arXiv:2306.00542}, 2023.

\bibitem[Yang et~al.(2021)Yang, Liu, Chen, Shen, Hao, and Wang]{yang2021causalvae}
Mengyue Yang, Furui Liu, Zhitang Chen, Xinwei Shen, Jianye Hao, and Jun Wang.
\newblock Causalvae: Disentangled representation learning via neural structural causal models.
\newblock In \emph{Proceedings of the IEEE/CVF conference on computer vision and pattern recognition}, pp.\  9593--9602, 2021.

\bibitem[Yoon et~al.(2018)Yoon, Jordon, and Van Der~Schaar]{yoon2018ganite}
Jinsung Yoon, James Jordon, and Mihaela Van Der~Schaar.
\newblock Ganite: Estimation of individualized treatment effects using generative adversarial nets.
\newblock In \emph{International Conference on Learning Representations}, 2018.

\end{thebibliography}
\bibliographystyle{iclr2025_conference}

\clearpage

\appendix

\section{Introduction on Latent Disentanglement}
\label{sec:intro-disentanglement}

The study of latent disentanglement in deep causal modeling has mainly focused on two areas. Most prior works lie in the first area that is causal representation learning, in which researchers seek to learn latent variables corresponding to true causal factors and identify the structure of their causal relations \citep{yang2021causalvae, halva2021disentangling, gresele2021independent, lachapelle2022disentanglement, komanduri2023learning}, predominantly leveraging recent advances in non-linear independent component analysis (ICA) \citep{hyvarinen2019nonlinear, khemakhem2020variational} and identify causal structure up to Markov equivalence \citep{spirtes2000causation} or graph isomorphism under rather heavy assumptions. The second area and the area this work is concerned of is counterfactual generative modeling, in which the focus is to conduct disentangled exogenous noise abduction that aids counterfactual outcome generation \citep{lotfollahi2021learning, shen2022weakly}. In this context, the goal is to acquire latent representations disentangled from the observed treatment to aid the correct identification of the causal effect of counterfactual treatments.

\section{Related Work}
\label{sec:related-work}

Key prior works in deep causal generative modeling fall primarily into three categories: variational-based methods \textbf{CEVAE} \citep{louizos2017causal}, \textbf{DeepSCM} \citep{pawlowski2020deep}, \textbf{Diff-SCM} \citep{sanchez2022diffusion}, \textbf{CHVAE} \citep{monteiro2023measuring}, \textbf{MED} \citep{ribeiro2023high}; adversarial-based methods \textbf{CausalGAN} \citep{kocaoglu2017causalgan}, \textbf{GANITE} \citep{yoon2018ganite}; hybrid method \textbf{DEAR} \citep{shen2022weakly} as well as our framework \textbf{VCI}. A comparative analysis of ours against these related works can be found in Table \ref{table:comparative-analysis}.

\begin{table}[ht]
  \caption{A comparative analysis of related work. ``Variational Objective Distribution'' describes the objective that the ELBOs are derived for in a variational-based method: ``Joint'' indicates joint distributions such as $p(y, x, t)$ or $p(y, z)$ (note that different works could have different notations, here we use $y$, $x$, $t$, $z$ to represent outcome, covariates, treatment, latent respectively), and ``Interventional'' indicates conditional outcome distribution $p(y | x, t)$ or $p(y | t)$. ``Marginal Distribution Alignment'' describes whether the stochastic optimization objective involves mechanisms that match the distribution of outcome constructions with some learnt marginal outcome distribution.}
  \label{table:comparative-analysis}
  \centering
  \resizebox{\textwidth}{!}{
      \begin{tabular}{lccccccc}
            \toprule
            & Type & \begin{tabular}{@{}c@{}} Variational \\ Objective \\ Distribution \end{tabular} & \begin{tabular}{@{}c@{}} Hierarchical \\ Model \\ Structure \end{tabular} & \begin{tabular}{@{}c@{}} Marginal \\ Distribution \\ Alignment \end{tabular} & \begin{tabular}{@{}c@{}} Causal \\ Discovery \end{tabular} & \begin{tabular}{@{}c@{}} End-to-End \\ Counterfactual \\ Supervision \end{tabular} & \begin{tabular}{@{}c@{}} Latent \\ Disentanglement \end{tabular} \\
            \midrule
            CEVAE & Variational & Joint & \tikzxmark & \tikzxmark & \tikzcmark & \tikzxmark & \tikzxmark \\
            [2mm]
            CausalGAN & Adversarial & -- & \tikzxmark & \tikzcmark & \tikzxmark & \tikzxmark & \tikzxmark \\
            [2mm]
            GANITE & Adversarial & -- & \tikzxmark & \tikzcmark & \tikzxmark & \tikzxmark & \tikzxmark \\
            [2mm]
            CPA & Adversarial & -- & \tikzxmark & \tikzxmark & \tikzxmark & \tikzxmark & \tikzcmark \\
            [2mm]
            DEAR & \begin{tabular}{@{}c@{}} Variational + \\ Adversarial \end{tabular} & Joint & \tikzxmark & \tikzxmark & \tikzcmark & \tikzxmark & * \\
            [2mm]
            DeepSCM & Variational & Interventional & \tikzxmark & \tikzxmark & \tikzxmark & \tikzxmark & \tikzxmark \\
            [2mm]
            Diff-SCM & Variational & Interventional & \tikzcmark & \tikzxmark & \tikzxmark & \tikzxmark & \tikzxmark \\
            [2mm]
            CHVAE & Variational & Interventional & \tikzcmark & \tikzxmark & \tikzxmark & \tikzxmark & \tikzxmark \\
            [2mm]
            MED & Variational & Interventional & \tikzcmark & \tikzcmark & \tikzxmark & \tikzxmark & \tikzxmark \\
            [2mm]
            VCI & \begin{tabular}{@{}c@{}} Variational + \\ Adversarial \end{tabular} & Counterfactual & \tikzcmark & \tikzcmark & \tikzxmark & \tikzcmark & \tikzcmark \\
            \bottomrule
        \end{tabular}
    }
    \begin{tabular}{l@{}l}
        * & \footnotesize{DEAR learns restrictive representation for latent $z$ ($z$ is disentangled from treatment $t$ when encoding $t$)} \\
        & \footnotesize{but does not learn consistent representation for $z$ ($t$ is not disentangled from $z$ when encoding $z$).} \\
    \end{tabular}
\end{table}

\textbf{CEVAE} uses variational autoencoders (VAEs) to infer latent confounders and estimate individual treatment effects (ITE). \textbf{DeepSCM} integrates deep learning into SCMs using normalizing flows and variational inference for counterfactual inference. \textbf{Diff-SCM} leverages generative diffusion models, using forward and reverse diffusion processes guided by anti-causal predictors to generate counterfactuals. \textbf{CHVAE} extends VDVAE to a conditional model, performing abduction by encoding images and parent attributes, and includes a penalty for counterfactual conditioning. \textbf{MED} uses hierarchical VAEs with deep causal mechanisms, inferring exogenous noise by conditioning on observed data and parent variables. \textbf{GANITE} uses a counterfactual generator to create proxies for unobserved outcomes and an ITE generator to estimate potential outcomes. \textbf{DEAR} combines a VAE-like generative loss with a supervised loss in a bidirectional generative model, trained with a GAN algorithm requiring knowledge of the causal graph structure and extra labels for latent factors.

As illustrated in Section \ref{proposed-method} and Appendix \ref{sec:comparison-variational}, the primary contribution of our framework \textbf{VCI} is the revolution of VAE formulation which leads to a brand new stochastic optimization scheme that optimizes the ELBO of an actual counterfactual objective. This brings about end-to-end counterfactual supervision and exogenous noise disentanglement as presented in Table \ref{table:comparative-analysis}.

\section{Comparison of Conditional VAE and VCI}
\label{sec:comparison-variational}

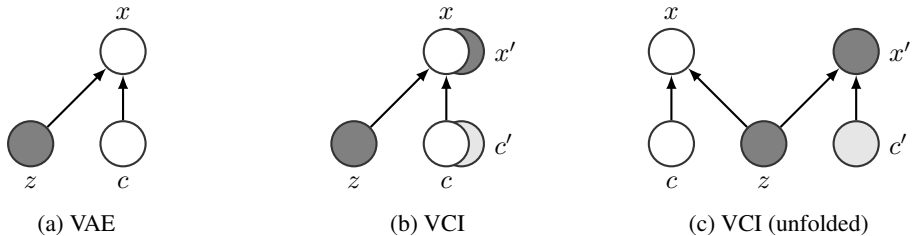
\begin{figure}
    \centering
    \begin{subfigure}[b]{0.25\textwidth}
        \centering
        \begin{tikzpicture}
            \tikzstyle{main}=[circle, minimum size = 6mm, thick, draw =black!80, node distance = 6mm]
            \tikzstyle{connect}=[-latex, thick]
            \tikzstyle{box}=[rectangle, draw=black!100]
              \node[main, fill = black!50] (z) [label=below:$z$] { };
              \node[main, fill = white!100] (c) [right=of z,label=below:$c$] { };
              \node[main, fill = white!100] (x) [above=of c,label=above:$x$] { };
              \path (z) edge [connect] (x)
            		(c) edge [connect] (x);
        \end{tikzpicture}
        \caption{VAE}
        \label{causal_diagram-vae}
    \end{subfigure}
    \hspace{1cm}
    \begin{subfigure}[b]{0.25\textwidth}
        \centering
        \begin{tikzpicture}
            \tikzstyle{main}=[circle, minimum size = 6mm, thick, draw =black!80, node distance = 6mm]
            \tikzstyle{connect}=[-latex, thick]
            \tikzstyle{box}=[rectangle, draw=black!100]
              \node[main, fill = black!50] (z) [label=below:$z$] { };
              \node[main, fill = white!100] (c) [right=of z,label=below:$c$] { };
              \begin{scope}[on background layer]
                \node[main, fill = black!10] (cp) [right=of z,label=right:$c'$, xshift=2mm] { };
              \end{scope}
              \node[main, fill = white!100] (x) [above=of c,label=above:$x$] { };
              \begin{scope}[on background layer]
                \node[main, fill = black!50] (xp) [above=of c,label=right:$x'$, xshift=2mm] { };
              \end{scope}
              \path (z) edge [connect] (x)
            		(c) edge [connect] (x);
        \end{tikzpicture}
        \caption{VCI}
        \label{causal_diagram-folded}
    \end{subfigure}
    \hspace{1cm}
    \begin{subfigure}[b]{0.25\textwidth}
        \centering
        \begin{tikzpicture}
            \tikzstyle{main}=[circle, minimum size = 6mm, thick, draw =black!80, node distance = 6mm]
            \tikzstyle{connect}=[-latex, thick]
            \tikzstyle{box}=[rectangle, draw=black!100]
              \node[main, fill = black!50] (z) [label=below:$z$] { };
              \node[main, fill = black!10] (cp) [right=of z,label=right:$c'$] { };
              \node[main, fill = white!100] (c) [left=of z,label=below:$c$] { };
              \node[main, fill = black!50] (xp) [above=of cp,label=right:$x'$] { };
              \node[main, fill = white!100] (x) [above=of c,label=above:$x$] { };
              \path (z) edge [connect] (x)
            		(z) edge [connect] (xp)
            		(c) edge [connect] (x)
            		(cp) edge [connect] (xp);
        \end{tikzpicture}
        \caption{VCI (unfolded)}
        \label{causal_diagram-unfolded}
    \end{subfigure}
    \caption{The data generating formulation of conditional generative modeling under VAE formulation and counterfactual generative modeling under VCI formulation using standard VAE notations. $p(x | z, c) = p(x' | z, c')$ for $c=c'$. White nodes are observed, light grey nodes are assigned during training and inference, dark grey nodes are unobserved.}
    \label{comparison-causal_diagram}
\end{figure}

\begin{table} 
  \caption{The objective and ELBO for conditional generative modeling under the VAE formulation and counterfactual generative modeling under the VCI formulation, using standard VAE notations. 
  As can be seen, the difference in motivations between the two objectives manifest themselves in the derived ELBOs in some meaningful and intuitive ways: VAE does not have the motivation to maximize counterfactual outcome likelihood, hence no such term $\log [ \hat{p} (x'_{\theta, \phi} | c') ]$ to conduct supervision of counterfactuals; VCI does not have the motivation to denoise or to generate samples from pure noise duing inference time, hence no such term $D_\mathrm{KL} [ q_\phi (z | x, c) \parallel p_\theta (z | c) ]$ to bound latent distribution on some marginal prior.}
  \label{comparison-objective}
  \centering
  \resizebox{\textwidth}{!}{
      \begin{tabular}{lcc}
            \toprule
            & VAE & VCI \\
            \midrule
            Objective & $p(x | c)$ & $p(x' | x, c, c')$ \\
            [2mm]
            ELBO & $\begin{aligned} J_{\theta,\phi}^\mathrm{VAE}(x, c) &= \colorbox{Goldenrod}{$\mathbb E_{q_\phi (z | x, c)} \log \left[ p_\theta (x | z, c) \right]$} \\ &\quad - D_\mathrm{KL} \left[ \colorbox{SkyBlue}{$q_\phi (z | x, c)$} \parallel \colorbox{Gray}{$p_\theta (z | c)$} \right] \end{aligned}$ & $\begin{aligned} J_{\theta,\phi}^\mathrm{VCI}(x, c, c') &= \colorbox{Goldenrod}{$\mathbb E_{q_\phi (z | x, c)} \log \left[ p_\theta (x | z, c) \right]$} + \colorbox{Apricot}{$\log \left[ \hat{p} (x'_{\theta, \phi} | c') \right]$} \\ &\quad - D_\mathrm{KL} \left[ \colorbox{SkyBlue}{$q_\phi (z | x, c)$} \parallel \colorbox{SpringGreen}{$q_\phi (z | x'_{\theta, \phi}, c')$} \right] \end{aligned}$ \\
            \bottomrule
        \end{tabular}
    }
\end{table}

For readers that are more familiar with deep generative modeling than causal inference, we provide a straightforward comparison between conditional VAE and VCI (fundamentally a comparison between conditional generative modeling and counterfactual generative modeling) in this section using VAE's notations. For data $x$ and condition $c$, the differences in formulations and objectives are shown in Figure \ref{comparison-causal_diagram} and Table \ref{comparison-objective}. In causal inference, counterfactual outcome is an individual-level concept – for a given individual that we observed outcome/data $x$ under treatment/condition $c$, what would their outcome have been if they had received treatment/condition $c'$ instead? This is a ``would have'' question, meaning that we seek to find out what the alternative outcome $x'$ would be in a parallel world where everything in the state of the universe remained the same, except that the treatment $c'$ (and its impact) was different. With this in mind, it is not hard to see why the conditional generative modeling formulation is unorthodox for counterfactual generative modeling -- the ELBO serves to optimize the marginal-level likelihood $p(x | c)$, which is interventional (rung 2) and not counterfactual (rung 3) \citep{pearl1995causal}, and the learnt model generates samples towards the marginal-level distribution $p(x |c)$ during inference time, and the question being asked here is ``what will an outcome $x$ be under condition $c$?''. This is a ``will'' question -- it is generating \textit{any} outcome under condition $c$, not given a specific individual, not given a specific state of the universe. Therefore, contrary to prior works in HVAEs and diffusion models \citep{sanchez2022diffusion, monteiro2023measuring, ribeiro2023high} which focus on model design adaptations to make the conditional VAE formulation work in counterfactual generative modeling, we first derive the orthodox formulation and objective for counterfactual generative modeling, then make the necessary model designs afterwards based on the accordingly derived ELBO.

\subsection{Diffusion Models}
\label{sec:discussion-diffusion}

Due to the popularity of the diffusion models \citep{ho2020denoising}, we feel the necessity to include a discussion here specifically about its compatibility with counterfactual generative modeling. It is very important to note that, although diffusion models are the state-of-the-art in generative and conditional generative modeling, it has not been shown that it has better capability in counterfactual generative modeling than ordinary HVAEs with learnable encoder \citep{monteiro2023measuring, ribeiro2023high}, and the reason is straightforward once we truly understand the fundamental incompatibility between the goal of counterfactual inference and the diffusion mechanism. Counterfactual inference entails abducting and preserving the exogenous noise in the original outcome; diffusion models, on the other hand, entail diffusing and denoising the original image -- it is a mechanism that inherently does not respect the consistency assumption (Assumption \ref{assumption-consistency}) and fundamentally contradicts exogenous noise abduction. Even if certain prior work such as \citet{sanchez2022diffusion} uses the learnt noise model to replace random noise in inference time, it is still performing the illogical operation of attempting to construct counterfactual from a noisy version of the original outcome, in which the exogenous noise has already been discarded. And this is on top of the fact that  diffusion models, as models under the VAE formulation, are working towards the wrong interventional objective as discussed above. Prior works that attempted at utilizing diffusion models in counterfactual generative modeling \citep{dash2022evaluating, sanchez2022diffusion, komanduri2024causal} clearly exhibited this flaw: as can be seen from the CelebA results in \citet{dash2022evaluating} and the MNIST results in \citet{komanduri2024causal}, exogenous noise/individuality has been largely discarded in counterfactual generation – because the diffusion model is doing what it is intended to do: diffusing exogenous and generating samples that fit seeminglessly into the conditional/marginal likelihood $p(x | c)$; \citet{sanchez2022diffusion} on the other hand, hardly showed any evidence that the model is capable of preserving individuality -- MNIST was the only benchmark in their experiments and only intervening digit was performed, which does not have measurable counterfactual truth as intervening thickness and intensity in Morpho-MNIST. In general, the ``interventions'' performed in \citet{sanchez2022diffusion} all drastically modified the objects in the original images and makes it quite impossible to tell if exogenous noise are preserved either quantitatively or qualitatively. When one drastically modifies the objects, the problem moves very far away from the consistency assumption, and it becomes really questionable whether the results under such ``intervention'' can even be defined as counterfactuals or not. We want to clarify that we think \citet{dash2022evaluating}, \citet{sanchez2022diffusion} and \citet{komanduri2024causal} are meaningful and well-written works that conducted the important advancement of experimenting diffusion models in counterfactual generative modeling, however, diffusion models have not shown the capability of exogenous noise abduction on the level of state-of-the-art HVAEs \citep{monteiro2023measuring, ribeiro2023high} as it stands.

To give the readers a sense of how much difference it makes to use the VCI framework instead of conditional diffusion models for counterfactual generative modeling, we presented the visual results on CelebA in Figure \ref{fig:celeba-hq} and the results on ``intervening'' digits of MNIST in Figure \ref{fig:morpho-mnist-sample-digit} for readers to compare to aforementioned prior arts.

\section{Comparison of Traditional Causal Formulation and VCI formulation}
\label{sec:comparison-causal}

\begin{figure}
    \centering
    \begin{subfigure}[b]{0.45\textwidth}
        \centering
        \begin{tikzpicture}
            \tikzstyle{main}=[circle, minimum size = 8mm, thick, draw =black!80, node distance = 8mm]
            \tikzstyle{connect}=[-latex, thick]
            \tikzstyle{box}=[rectangle, draw=black!100]
              \node[main, fill = white!100] (X) [label=below:$X$] { };
              \node[main, fill = black!50, minimum size = 4mm] (UX) [left=of X,label=left:{\tiny$U_X$}, xshift=6mm, yshift=4mm] { };
              \node[main, fill = white!100] (T) [right=of X,label=below:$T$, xshift=6mm] { };
              \node[main, fill = black!50, minimum size = 2mm] (UT) [left=of T,label=left:{\tiny $U_T$}, xshift=6mm, yshift=4mm] { };
              \begin{scope}[on background layer]
                \node[main, fill = black!10] (Tp) [right=of X,label=right:$\mathrm{do}(T\equal t')$, xshift=8mm] { };
              \end{scope}
              \node[main, fill = white!100] (Y) [above=of $(X.north)!0.5!(T.north)$,label=left:$Y$, xshift=-1mm] { };
              \begin{scope}[on background layer]
                \node[main, fill = black!50] (Yp) [above=of $(X.north)!0.5!(T.north)$,label=right:$Y_{t'}$, xshift=1mm] { };
              \node[main, fill = black!50, minimum size = 4mm] (UY) [above=of $(X.north)!0.5!(T.north)$,label=left:{\tiny$U_Y$}, yshift=12mm] { };
              \end{scope}
              \path (UX) edge [connect] (X)
                    (UY) edge [connect] (Y)
                    (UY) edge [connect] (Yp)
                    (X) edge [connect] (T)
                    (UT) edge [connect] (T)
                    (X) edge [connect] (Y)
                    (T) edge [connect] (Y);
        \end{tikzpicture}
        \caption{SCM, traditional}
        \label{scm-traditional}
    \end{subfigure}
    \begin{subfigure}[b]{0.54\textwidth}
        \centering
        \begin{tikzpicture}
            \tikzstyle{main}=[circle, minimum size = 8mm, thick, draw =black!80, node distance = 8mm]
            \tikzstyle{connect}=[-latex, thick]
            \tikzstyle{box}=[rectangle, draw=black!100]
              \node[main, fill = white!100] (X) [label=below:$X$] { };
              \node[main, fill = black!50, minimum size = 4mm] (UX) [left=of X,label=left:{\tiny$U_X$}, xshift=6mm, yshift=4mm] { };
              \node[main, fill = white!100] (T) [right=of X,label=below:$T$, xshift=6mm] { };
              \node[main, fill = black!50, minimum size = 2mm] (UT) [left=of T,label=left:{\tiny $U_T$}, xshift=6mm, yshift=4mm] { };
              \begin{scope}[on background layer]
                \node[main, fill = black!10] (Tp) [right=of X,label=right:$\mathrm{do}(T\equal t')$, xshift=8mm] { };
              \end{scope}
              \node[main, fill = black!50] (Z) [above=of X,label=above:$Z$, xshift=10mm] {};
              \node[main, fill = black!50, minimum size = 4mm] (UY) [left=of Z,label=left:{\tiny$U_Y$}, xshift=6mm, yshift=4mm] { };
              \node[main, fill = white!100] (Y) [right=of Z,label=above:$Y$, xshift=6mm] { };
              \begin{scope}[on background layer]
                \node[main, fill = black!50] (Yp) [right=of Z,label=right:$Y_{t'}$, xshift=8mm] { };
              \end{scope}
              \path (UX) edge [connect] (X)
                    (X) edge [connect] (Z)
                    (UY) edge [connect] (Z)
                    (X) edge [connect] (T)
                    (UT) edge [connect] (T)
                    (Z) edge [connect] (Y)
                    (T) edge [connect] (Y);
        \end{tikzpicture}
        \caption{SCM, with $f_Y(X, T, U_Y) = \tilde{f}_Y(f_Z(X, U_Y), T)$}
        \label{scm-extended}
    \end{subfigure}
    
    \begin{subfigure}[b]{0.95\textwidth}
        \centering
        \begin{tikzpicture}
            \tikzstyle{main}=[circle, minimum size = 8mm, thick, draw =black!80, node distance = 8mm]
            \tikzstyle{connect}=[-latex, thick]
            \tikzstyle{box}=[rectangle, draw=black!100]
              \node[main, fill = white!100] (X) [label=below:$X$] { };
              \node[main, fill = black!50, minimum size = 4mm] (UX) [left=of X,label=left:{\tiny$U_X$}, xshift=6mm, yshift=4mm] { };
              \node[main, fill = black!10] (Tp) [right=of X,label=right:$t'$, xshift=6mm] { };
              \node[main, fill = white!100] (T) [left=of X,label=below:$T$, xshift=-6mm] { };
              \node[main, fill = black!50, minimum size = 2mm] (UT) [left=of T,label=left:{\tiny $U_T$}, xshift=6mm, yshift=4mm] { };
              \node[main, fill = black!50] (Z) [above=of X,label=above:$Z$] {};
              \node[main, fill = black!50] (Yp) [above=of Tp,label=right:$Y_{t'}$] { };
              \node[main, fill = white!100] (Y) [above=of T,label=above:$Y$] { };
              \node[main, fill = black!50, minimum size = 4mm] (UY) [left=of Z,label=left:{\tiny$U_Y$}, xshift=6mm, yshift=4mm] { };
              \path (UX) edge [connect] (X)
                    (X) edge [connect] (Z)
                    (UY) edge [connect] (Z)
                    (X) edge [connect] (T)
                    (UT) edge [connect] (T)
                    (Z) edge [connect] (Y)
                    (Z) edge [connect] (Yp)
                    (T) edge [connect] (Y)
                    (Tp) edge [connect] (Yp);
        \end{tikzpicture}
        \caption{SCM, with $f_Y(X, T, U_Y) = \tilde{f}_Y(f_Z(X, U_Y), T)$, unfolded}
        \label{scm-unfolded}
    \end{subfigure}
    \caption{The evolution of our formulation from traditional SCM with observed triplets $(X, T, Y)$. \Cref{scm-extended} is simply extending the production function towards $Y$ with an intermediate variable $Z$ representing two of its inputs (when $f_Z(X, U_Y) = (X, U_Y)$, $\tilde{f}_Y$ is just $f_Y$). Then, \Cref{scm-unfolded} is just the unfolded graphical model of \Cref{scm-extended}, as the $\mathrm{do}()$ operation replaces $T$ in the production function of $Y$ with $t'$, i.e. $Y_{t'} = f_Y(X, t', U_Y)$.}
    \label{comparison-formulation}
\end{figure}
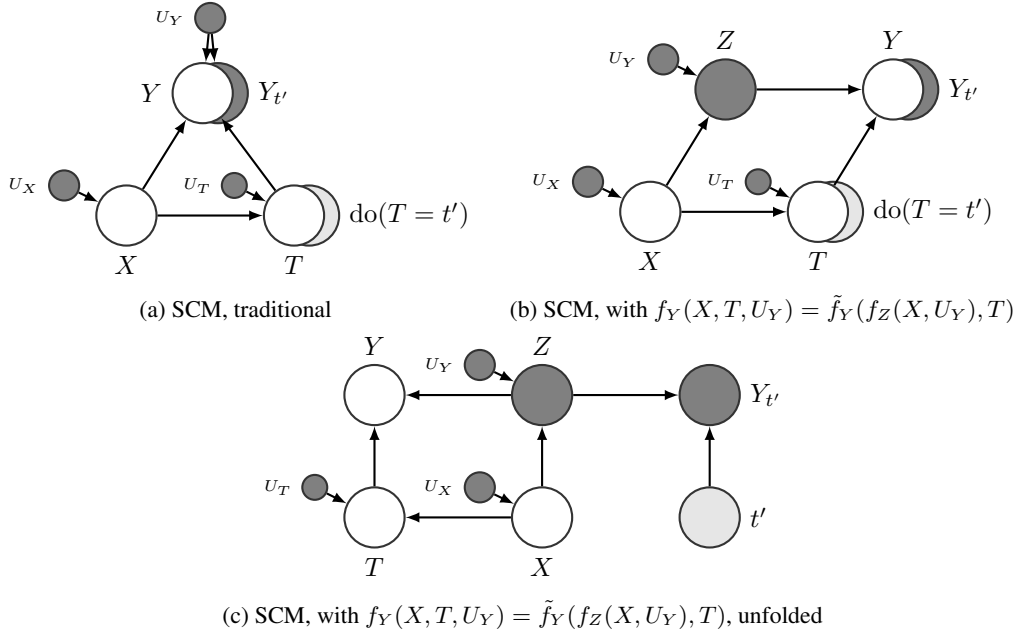

We note that we formulate counterfactuals $Y'$ and $T'$ as separate variables for the cleanliness of graphical model and notation simplicity in variational inference, with the goal to bridge readers with variational inference background and causal inference background. In essence, it is no different from the traditional causal formulation where the $\mathrm{do}()$ operation is directly imposed on factual variable $T$, as $T'$ is just a dummy variable with no observations. Denote $Y_{\mathrm{do}(T=t')}$ as $Y_{t'}$, see Figure \ref{comparison-formulation} for a demonstration of our formulation in traditional notations.

Then under the ignorability assumption (Assumption \ref{assumption-ignorability}, note that there is no $\epsilon_Y$ and $\epsilon'_Y$ in the traditional SCM formulation as there is only one copy of exogenous noise $U_Y$ injected to factual and counterfactual outcome, but in general if there exists $\epsilon_Y$ and $\epsilon'_Y$ independent of $U_Y$, the following result holds as long as $\epsilon_Y, \epsilon'_Y \indep U_T$), we have the following version of Theorem \ref{elbo}:

\begin{corollary}
    \label{elbo-causal}
    Suppose a collection of random variables follows a causal structure defined by the Bayesian network in \Cref{scm-extended}. Then $\log [ p (Y_{t'} | Y, X, T, t') ]$ has the following variational lower bound:
    \begin{align}
        \label{eq-elbo-causal}
        \log \left[ p (Y_{t'} | Y, X, T, t') \right] &\geq \mathbb E_{p (Z | Y, T, X)} \log \left[ p (Y | Z, T) \right] - D \left[ p (Y | X, T) \parallel p (Y_{t'} | X, t') \right] \nonumber \\
        &\quad - D_\mathrm{KL} \left[ p (Z | Y, T, X) \parallel p (Z | Y_{t'}, t', X) \right]
    \end{align}
    where $D [ p \parallel q ] = \log p - \log q$.
\end{corollary}

Now we define dummy variable
\begin{align}
    T' = f_T(X, U'_T),
\end{align}
with $U'_T$ satisfying the ignorability assumption (Remark \ref{remark-ignorability}), and
\begin{align}
    Y'=f_Y(X, T', U_Y) = \tilde{f}_Y(f_Z(X, U_Y), T').
\end{align}
Then, $Y'_{\mathrm{do(T'=t')}} = f_Y(X, t', U_Y) = Y_{t'}$. Hence, we have
\begin{align}
    p (Y_{t'} | Y, X, T, t') = p (Y' | Y, X, T, do(T'=t')) = p (Y' | Y, X, T, T'=t')
\end{align}
for any $t'$. Therefore, \Cref{causal_diagram-bayesian} and Theorem \ref{elbo} is in essence just \Cref{scm-unfolded} and Corollary \ref{elbo-causal} with different notations and variable definitions.

\section{Implicit Counterfactual Supervision}
\label{sec:implicit-cf-supervision}

We demonstrated in Appendix \ref{sec:comparison-variational} that the marginal-level observed likelihood is unorthodox for counterfactual generative modeling and the individual-level likelihood $\log [ p (Y' | Y, X, T, T') ]$ is better motivated for this task. However, for researchers who insist on working towards the traditional likelihood objective $\log [ p (Y | X, T) ]$ composed only of observed variables, there is an approach to conduct counterfactual supervision implicitly -- to view $Y'$ as a latent variable restricted on the prior $p (Y' | X, T')$ and view counterfactual construction task as a latent recognition task:

\begin{proposition}
    \label{elbo-implicit}
    $J(O) = \log [ p (Y | X, T) ]$ has the following variational lower bound:
    \begin{align}
        \label{eq-elbo-implicit}
        J(O) &\geq \mathbb E_{p (Z | Y, T, X)} \log \left[ p (Y | Z, T) \right] - \mathbb E_{p (Z | Y, T, X)} D_\mathrm{KL} \left[ p (Y' | Z, T') \parallel p (Y' | X, T') \right] \nonumber \\
        &\quad - D_\mathrm{KL} \left[ p (Z | Y, T, X) p (Y' | Z, T') \parallel p (Z | Y', T', X) p (Y' | Z, T') \right].
    \end{align}
\end{proposition}

This induces the following ELBO in stochastic optimization:
\begin{align}
    \label{VCI-implicit-objective}
    J_{\theta, \phi}(o) &= \mathbb E_{q_\phi (z | y, t, x)} \log \left[ p_\theta (y | z, t) \right] - \mathbb E_{q_\phi (z | y, t, x)} D_\mathrm{KL} \left[ p_\theta (y' | z, t') \parallel \hat{p} (y' | x, t') \right] \nonumber \\
    &\quad - D_\mathrm{KL} \left[ q_\phi (z | y, t, x) p_\theta (y' | z, t') \parallel q_\phi (z | y', t', x) p_\theta (y' | z, t') \right].
\end{align}
Equation \ref{VCI-implicit-objective} avoids estimating $Y'$ with a single sample as in the objective (Equation \ref{VCI-objective}) derived from the explicit ELBO (Equation \ref{eq-elbo}). However, it could be a lot more sampling inefficient as the latent divergence term requires Monte-Carlo samples from both $Z$ and $Y'$ to be estimated.

\section{Definition of Oracle Consistency, Oracle Restrictiveness, and Disentanglement}
\label{sec:disentanglement}

Let $\mathcal{H} = \{ (\tilde{p}, \tilde{q}) \mid \tilde{p}: \mathcal{S} \to [0, 1]^{\Sigma_\mathcal{V}}, \; \tilde{q}: \mathcal{V} \times \mathcal{X} \to [0, 1]^{\Sigma_\mathcal{S}} \}$ be a hypothesis class of stochastic models $(\tilde{p}, \tilde{q})$. The input space of $\tilde{p}$ is a measurable space $\mathcal{S}$ where a feature vector $s \in \mathcal{S}$ is drawn; the input space of $\tilde{q}$ is the product space of $\mathcal{V}$ and $\mathcal{X}$ where a vector of interest $v \in \mathcal{V}$ and some auxiliary attributes $x \in \mathcal{X}$ are drawn. The outputs of $\tilde{p}$ and $\tilde{q}$ are probability measures on $\mathcal{V}$ and $\mathcal{S}$ respectively. Let the true model $(p^*, q^*)$ be the true conditional distributions $(p_0(V | s), p_0(S | v, x))$ of random variables $V$ and $S$. The following definitions provide criteria under which two parts $S_I$ and $S_{\backslash I}$ of feature variable $S$ separated by an index set $I$ are disentangled by a model.

\begin{definition}[Oracle Consistency]
    \label{def:oracle-consistency}
    Consider $s = (s_I, s_{\backslash I})$ and $s' = (s_I, s_{\backslash I}')$ where
    \begin{align}
        x &\sim p(X) \\
        s_I &\sim p(S_I | x) \\
        s_{\backslash I}, s'_{\backslash I} &\overset{\mathrm{iid}}{\sim} p(S_{\backslash I} | s_I, x)
    \end{align}
    We say that $\tilde{q}$ is consistent with $S_I$ if $p$ is the true distribution $p_0$ and 
    \begin{align}
        \mathbb E_{p_0, p^*} \lVert \tilde{q}_I \circo p^*(s; x) - \tilde{q}_I \circo p^*(s'; x) \rVert = 0,
    \end{align}
    where $\tilde{q}_I(s_I | v, x) = \int_{s_{\backslash I}} \tilde{q}(s_I, s_{\backslash I} | v, x)$. More generally, we say that $(\tilde{p}, \tilde{q})$ is consistent with $S_I$ under $p$ if
    \begin{align}
        \mathbb E_{p, \tilde{p}} \lVert \tilde{q}_I \circo \tilde{p}(s; x) - \tilde{q}_I \circo \tilde{p}(s'; x) \rVert = 0.
    \end{align}
\end{definition}

As discussed in \citet{shu2019weakly}, oracle consistency (or encoder consistency) states that for any fixed choice of $S_I$, resampling $S_{\backslash I}$ should not affect the oracle's measurement of $S_I$. A more general notion of oracle consistency is defined in addition, as it is valuable in real-world optimization to evaluate the estimating model's consistency under its own oracle $\tilde{q} \circo \tilde{p}$.

In contrast, restrictiveness states that for any fixed choice of $S_{\backslash I}$, resampling $S_I$ should not affect the oracle's measurement of $S_{\backslash I}$:
\begin{definition}[Oracle Restrictiveness]
    \label{def:oracle-restrictiveness}
    Consider $s = (s_I, s_{\backslash I})$ and $s' = (s_I', s_{\backslash I})$ where
    \begin{align}
        x &\sim p(X) \\
        s_{\backslash I} &\sim p(S_{\backslash I} | x) \\
        s_I, s'_I &\overset{\mathrm{iid}}{\sim} p(S_I | s_{\backslash I}, x)
    \end{align}
    We say that $\tilde{q}$ is restricted to $S_I$ if $p$ is the true distribution $p_0$ and 
    \begin{align}
        \mathbb E_{p_0, p^*} \lVert \tilde{q}_{\backslash I} \circo p^*(s; x) - \tilde{q}_{\backslash I} \circo p^*(s'; x) \rVert = 0,
    \end{align}
    where $\tilde{q}_{\backslash I}(s_{\backslash I} | v, x) = \int_{s_I} \tilde{q}(s_I, s_{\backslash I} | v, x)$. More generally, we say that $(\tilde{p}, \tilde{q})$ is restricted to $S_I$ under $p$ if
    \begin{align}
        \mathbb E_{p, \tilde{p}} \lVert \tilde{q}_{\backslash I} \circo \tilde{p}(s; x) - \tilde{q}_{\backslash I} \circo \tilde{p}(s'; x) \rVert = 0.
    \end{align}
\end{definition}

Notice that the oracle consistency and restrictiveness of $S_I$ is equivalent to the oracle restrictiveness and consistency of $S_{\backslash I}$, we have the following definition of disentanglement:

\begin{definition}[Disentanglement]
    \label{def:disentanglement}
    We say that $\tilde{q}$ disentangles $S_I$ and $S_{\backslash I}$ if $\tilde{q}$ is consistent with and restricted to $S_I$. More generally, we say that $(\tilde{p}, \tilde{q})$ disentangles $S_I$ and $S_{\backslash I}$ under $p$ if $(\tilde{p}, \tilde{q})$ is consistent with and restricted to $S_I$ under $p$.
\end{definition}

\section{Robust Marginal Effect Estimation}
\label{sec:covar-spec_ATT}

By \citet{van2000asymptotic}, a sequence of estimators $\hat{\Psi}_n$ is asymptotically efficient if $\sqrt{n}(\hat{\Psi}_n - \Psi(p)) = 1 / \sqrt{n} \sum_{k=1}^n \tilde{\psi}_p(W_k) + o_p(1)$ where $\tilde{\psi}_p$ is the efficient influence function of $\Psi(p)$ and $W_k \sim p(W)$. The theorem below gives this efficient influence function for the average treatment effect of the treated (ATT) $\Psi(p) = \mathbb E_p [ Y'_{\mathrm{do}(T'=\alpha)} ]$ and thus provides a construction of an asymptotically efficient regular estimator for $\Psi$:

\begin{theorem}
    \label{variational-ATT}
    Suppose $W: \Omega \rightarrow \mathcal{R}_W$ follows a causal structure defined by the Bayesian network in Figure \ref{causal_diagram}, where the counterfactual conditional distribution $p(Y', T' | Z, X)$ is identical to that of its factual counterpart $p(Y, T | Z, X)$. Then $\Psi(p)$ has the following efficient influence function:
    \begin{align}
        \tilde{\psi}_p(W) &= \frac{I(T = \alpha)}{p(T | X)} (Y - \mathbb E_p\left[ Y | Z, T \right]) + \mathbb E_p\left[ Y' | Z, T'=\alpha \right] - \Psi.
    \end{align}
\end{theorem}

This leads to the construction of $\hat{\Psi}_n (o)$ in Equation \ref{ATT-estimator} which is asymptotically efficient under some regularity conditions \citep{van2006targeted}, with the asymptotic distribution given by
\begin{align}
    \label{ATT-estimator-dist}
     \sqrt{n}(\hat{\Psi}_n -\Psi(p)) \xrightarrow{d} \mathcal{N}(0, E_p\left[\tilde{\psi}_p(W) \tilde{\psi}_p(W)^T\right])
\end{align}
where the variance can be estimated empirically in practice and confidence bounds can be constructed accordingly. The robust estimator within the VCI framework can also be extended to estimating covariate-specific marginal treatment effect $\Xi(p) = \mathbb E_p [ Y'_{\mathrm{do}(X=c, T'=\alpha)} ]$ for a given covariate $c$ of interest:

\begin{corollary}
    \label{variational-ATT_covar-spec}
    Under the same settings as Theorem \ref{variational-ATT}, $\Xi(p)$ has the following efficient influence function:
    \begin{align}
        \tilde{\xi}_p(W) &= \frac{I(X = c)}{p(X)} \tilde{\psi}_p(W).
    \end{align}
\end{corollary}

Note that when $p(X)$ is estimated empirically, the asymptotically efficient estimator for $\Xi(p)$:
\begin{align}
    \label{ATT-estimator_covar-spec}
    \hat{\Xi}_n (o) &= \frac{1}{n} \sum_{i=1}^{n} \frac{I(x_k = c)}{\hat{p}(x_k)} \left\{ \frac{I(t_k = \alpha)}{\hat{e}(t_k | x_k)} \cdot y_k + \left(1-\frac{I(t_k = \alpha)}{\hat{e}(t_k | x_k)}\right) \cdot \mathbb E_{p_\theta} \left[ Y' | z_{k, \phi}, T'=\alpha \right] \right\} \\
    &= \frac{1}{n_c} \sum_{i=1}^{n_c} \left\{ \frac{I(t_{k_i} = \alpha)}{\hat{e}(t_{k_i} | x_{k_i})} \cdot y_{k_i} + \left(1-\frac{I(t_{k_i} = \alpha)}{\hat{e}(t_{k_i} | x_{k_i})}\right) \cdot \mathbb E_{p_\theta} \left[ Y' | z_{{k_i}, \phi}, T'=\alpha \right] \right\}
\end{align}
is simply an application of Equation \ref{ATT-estimator} on the set of observations with covariates $c$, where $\{ k_1, k_2, \dots, k_{n_c} \}$ are indices such that $x_{k_i}=c$.

\section{Formal Definition of Common Causal Assumptions under the VCI Formulation}
\label{sec:causal-assumptions}

\begin{assumption}[Ignorability]
    \label{assumption-ignorability}
    There is no unobserved confounders, i.e. $\tilde{U}_Y \indep U_T$, where $\tilde{U}_Y = (U_Y, \epsilon_Y, \epsilon'_Y)$.
\end{assumption}

Note that this assumption can be extended to the counterfactual treatment variable $T'$ w.o.l.g.:

\begin{remark}[Ignorability]
    \label{remark-ignorability}
    Note that $T'$ in our formulation is a dummy variable with no observations, and we only care about the outcome distribution of $Y'$ under assignment $do(T'=t')$ for any given $t'$. Hence, the ignorability $\tilde{U}_Y \indep U'_T$, and in addition, $U_X \indep U'_T$, can be assumed w.o.l.g. just so that the outcome distributions under $do(T'=t')$ reduces to that under $T'=t'$ conditioned on $X$ or $Z$ for notation simplicity, without violating any actual mechanism on the observed data.
\end{remark}

The ignorability assumption is universally assumed and entailed in our formulation in Figure \ref{causal_diagram}.

\begin{assumption}[Consistency]
    \label{assumption-consistency}
    An individual's observed outcome under a treatment is the same as its potential outcome under that treatment, i.e. $Y = Y'_{\mathrm{do}(T'=T)}$, where $Y'_{\mathrm{do}(T'=T)} = f_Y(Z, T, \epsilon'_Y)$ is yielded by replacing the production equation for $T'$ with $T'=T$.
\end{assumption}

The consistency assumption, commonly stated in causal inference literature, is indeed an assumption and not just a direct consequence of the SCM under our formulation, since $\epsilon_Y$ and $\epsilon'_Y$ are two copies of independent variables. In essence, it states that $\epsilon_Y = 0$ and there is only one copy of exogenous noise $U_Y$ injected to $Y$ and $Y'$:

\begin{remark}[Consistency]
    \label{remark-consistency}
    Under the consistency assumption, $f_Y(z, t, \epsilon_Y)$ must be deterministic for any $z$ and $t$, otherwise we would have $Y \neq Y'_{\mathrm{do}(T'=T)}$ since $f_Y(z, t, \epsilon_Y)$ and $f_Y(z, t, \epsilon'_Y)$ are independent for any $z$ and $t$. Hence there exists deterministic function $\tilde{f}_Y$ such that $\tilde{f}_Y(Z, T, 0) = f_Y(Z, T, \epsilon_Y)$. So w.o.l.g., we let $\epsilon_Y = 0$ under the consistency assumption.
\end{remark}

Note that this assumption does not result in a collapse of counterfactual inference to interventional inference in the traditional sense, as latent $Z$ is unobserved and entails further uncertainty $U_Y$ beyond the observed covariates $X$.

\section{Proof of Theoretical Results}
\label{proofs}

\subsection{Evidence Lower Bound}
\label{elbo-proofs}

\subsubsection{Proof of Theorem \ref{elbo}}
\label{elbo_proof}

\begin{proof}
    By the d-separation \citep{pearl1988probabilistic} of paths on the causal graph defined in \Cref{causal_diagram-bayesian}, we have
    \begin{align}
        \log \left[ p (Y' | Y, X, T, T') \right]
     & = \log \mathbb E_{p (Z | Y, T, X)} \left[ 
        p (Y' | Z, Y, X, T, T') \right] \\
     & \geq \mathbb E_{p (Z | Y, T, X)} \log \left[ 
        p (Y' | Z, Y, X, T, T') \right] \quad \text{(Jensen's inequality)}\\
     & = \mathbb E_{p (Z | Y, T, X)} \log \frac{p (Y', Z | Y, X, T, T')}{p (Z | Y, T, X)} \\
     & = \mathbb E_{p (Z | Y, T, X)} \log \frac{p (Y', Z, Y | X, T, T')}{p (Z | Y, T, X) p (Y | X, T)} \\
     & = \mathbb E_{p (Z | Y, T, X)} \log \frac{p (Y | Z, T) p (Z | Y', T', X) p (Y' | X, T')}{p (Z | Y, T, X) p (Y | X, T)} \\
     & = \mathbb E_{p (Z | Y, T, X)} \log \left[ p (Y | Z, T) \right] - D_\mathrm{KL} \left[ p (Z | Y, T, X) \parallel p (Z | Y', T', X) \right] \nonumber \\
     &\quad - D \left[ p (Y | X, T) \parallel p (Y' | X, T') \right].
    \end{align}
    Note that the above steps still hold with an additional edge from $Z$ or $T$ to $X$ (but not both) if $U_Y$ or $U_T$ is dependent of $U_X$.
\end{proof}

\subsubsection{Proof of Proposition \ref{elbo-implicit}}
\label{elbo-implicit_proof}

\begin{proof}
    By the d-separation \citep{pearl1988probabilistic} of paths on the causal graph defined in \Cref{causal_diagram-bayesian}, we have
    \begin{align}
        \log \left[ p (Y | X, T) \right]
     & = \log \left[ p (Y | X, T, T') \right] = \log \mathbb E_{p (Z, Y' | Y, X, T, T')} \frac{p (Y, Z, Y' | X, T, T')}{p (Z, Y' | Y, X, T, T')}  \\
     & \geq \mathbb E_{p (Z, Y' | Y, X, T, T')} \log \frac{p (Y, Z, Y' | X, T, T')}{p (Z, Y' | Y, X, T, T')} \quad \text{(Jensen's inequality)}\\
     & = \mathbb E_{p (Z, Y' | Y, X, T, T')} \log \frac{p (Y' | X, T') p (Y, Z | Y', X, T, T')}{p (Z, Y' | Y, X, T, T')} \\
     & = \mathbb E_{p (Z, Y' | Y, X, T, T')} \log \frac{p (Y' | X, T') p (Z | Y', T', X) p (Y | Z, T)}{p (Z | Y, T, X) p (Y' | Z, T')} \\
     & = \mathbb E_{p (Z, Y' | Y, X, T, T')} \log \frac{p (Y' | X, T') p (Z | Y', T', X) p (Y' | Z, T') p (Y | Z, T)}{p (Y' | Z, T') p (Z | Y, T, X) p (Y' | Z, T')} \\
     & = - \mathbb E_{p (Z | Y, T, X)} D_\mathrm{KL} \left[ p (Y' | Z, T') \parallel p (Y' | X, T') \right] \nonumber \\
     &\quad - D_\mathrm{KL} \left[ p (Z | Y, T, X) p (Y' | Z, T') \parallel p (Z | Y', T', X) p (Y' | Z, T') \right] \nonumber \\
     &\quad + \mathbb E_{p (Z | Y, T, X)} \log \left[ p (Y | Z, T) \right].
    \end{align}
    Note that the above steps still hold with an additional edge from $Z$ or $T$ to $X$ (but not both) if $U_Y$ or $U_T$ is dependent of $U_X$.
\end{proof}

\subsubsection{Proof of Corollary \ref{elbo-causal}}
\label{elbo-causal_proof}

\begin{proof}
    Note that Theorem \ref{elbo} only relies on the conditional dependency structure in Figure \ref{causal_diagram}, and the only difference between the dependency structure of $(X, Z, T, T', Y, Y')$ in Figure \ref{causal_diagram} and that of $(X, Z, T, t', Y, Y_{t'})$ in Figure \ref{comparison-formulation} is the dependency between $(X, T')$ and $(X, t')$. However, such difference does not change the conditional independence of latent posterior: $p(Z | Y, T, X, t') = p(Z | Y, T, X)$ and $p(Z | Y_{t'}, t', X, T) = p(Z | Y_{t'}, t', X)$, the conditional independence of outcome distribution $p(Y | Z, T, t') = p(Y | Z, T)$ and $p(Y_{t'} | Z, t', T) = p(Y_{t'} | Z, t')$, and the conditional independence of covariate-specific outcome distribution $p(Y | X, T, t') = p(Y | X, T)$ and $p(Y_{t'} | X, t', T) = p(Y_{t'} | X, t')$. Hence, the same proof for Theorem \ref{elbo} applies to Corollary \ref{elbo-causal} by replacing $T'$ with $t'$ and $Y'$ with $Y_{t'}$.
\end{proof}


\subsection{Disentangled Exogenous Noise Abduction}

\subsubsection{Proof of Lemma \ref{lemma:disentanglement}}
\begin{proof}
    Let $S_I = Z$ and $S_{\backslash I} = T$. On one hand, 
    \begin{align}
        &\quad \mathbb E_{p_0, p*} \lVert \tilde{q}_I \circo p^*((s_I, s_{\backslash I}); x) - \tilde{q}_I \circo p^*((s_I, s_{\backslash I}'); x) \rVert \nonumber\\
        &= \mathbb E_{p_0(x), p_0(s_I, s_{\backslash I}, s_{\backslash I}' | x), p_0(v | s_I, s_{\backslash I}), p_0(v' | s_I, s_{\backslash I}')} \lVert \tilde{q}_I(v, x) - \tilde{q}_I(v', x) \rVert \\
        &= \mathbb E_{p_0(x) p_0(z, t, t' | x), p_0(y | z, t), p_0(y' | z, t')} \lVert \tilde{q}_I(y, t, x) - \tilde{q}_I(y', t', x) \rVert \\
        &= \mathbb E_{p_0(x, t, t', y, y')} \lVert q(\cdot | y, t, x) - q(\cdot | y', t', x) \rVert \\
        & \leq \mathbb E_{p_0 (d)} \sqrt{\frac{1}{2} D_\mathrm{KL} \left[ q (\cdot | y, t, x) \parallel q (\cdot | y', t', x) \right]} \quad \text{(Pinsker's inequality)}\\
        &= 0,
    \end{align}
    since Equation \ref{eq-identifiability-1} implies that $D_\mathrm{KL} [ q (\cdot | y, t, x) \parallel q (\cdot | y', t', x) ] = 0$ a.e. with respect to $p_0$. Hence $\tilde{q}$ is consistent with $Z$. On the other hand, $\tilde{q}$ is restricted to $Z$ by design:
    \begin{align}
        &\quad \mathbb E_{p_0, p*} \lVert \tilde{q}_{\backslash I} \circo p^*((s_I, s_{\backslash I}); x) - \tilde{q}_{\backslash I} \circo p^*((s_I', s_{\backslash I}); x) \rVert \nonumber\\
        &= \mathbb E_{p_0(s_I, s_I', s_{\backslash I}, x), p_0(v | s_I, s_{\backslash I}), p_0(v' | s_I', s_{\backslash I})} \lVert \tilde{q}_{\backslash I}(v, x) - \tilde{q}_{\backslash I}(v', x) \rVert \\
        &= \mathbb E_{p_0(z, z', t, x), p_0(y | z, t), p_0(y' | z', t)} \lVert \tilde{q}_{\backslash I}(y, t, x) - \tilde{q}_{\backslash I}(y', t, x) \rVert \\
        &= \mathbb E_{p_0(t)} \lVert \delta_t(\cdot) - \delta_t(\cdot) \rVert \\
        &= 0.
    \end{align}
    Therefore $\tilde{q}$ disentangles $Z$ and $T$.
\end{proof}

\subsubsection{Proof of Proposition \ref{thm:identifiability-1}}
\label{proof:identifiability-1}

\begin{assumption}[Uniqueness]
    \label{assumption-uniqueness}
    Two different individuals do not have the exact same outcome (a.s.) under the same treatment $\alpha$, i.e. for $y \sim p(Y | z, \alpha)$, $y' \sim p(Y | z', \alpha)$, we have $y \neq y'$ a.s. if $z \neq z'$.
\end{assumption}

The uniqueness assumption is weak and automatically satisfied under common assumptions in prior works such as injective decoder, continuous outcome or diffeomorphism \citep{khemakhem2020variational, von2023nonparametric}. Note that this assumption could be unrealistic in traditional causal inference but is very reasonable in high-dimensional outcome settings -- take facial imaging dataset for example, no two individuals with different facial features should look exactly identical pixel-by-pixel under the same treatment.

\begin{proof}
    Denote 
    \begin{align}
        LB(p_., q_.) &:= \mathbb E_{p_0 (d)} \left\{\mathbb E_{q_. (z | y, t, x)} \log \left[ p_. (y | z, t) \right] + \log \left[ \hat{p} (y' | x, t') \right] \right. \nonumber \\
        &\quad - \left. D_\mathrm{KL} \left[ q_. (\cdot | y, t, x) \parallel q_. (\cdot | y', t', x) \right] \right\}
    \end{align}
    and suppose that $(p_*, q_*) \in \operatorname*{arg\,max}_{p_., q_.} LB(p_., q_.)$ but 
    \begin{align}
        \mathbb E_{p_0 (d)} D_\mathrm{KL} \left[ q_* (\cdot | y, t, x) \parallel q_* (\cdot | y', t', x) \right] > 0,
    \end{align}
    which we can factorize as
    \begin{align}
        \mathbb E_{p_0 (x, z, t, t'), p^*} D_\mathrm{KL} \left[ q_* \circo p^* (z, t; x) \parallel q_* \circo p^* (z, t'; x) \right] > 0
    \end{align}
    where $p^*(z, t) = (p_0(Y | z, t), \delta_t(T))$. Then we consider model $(p_{**}, q_{**})$ such that $\forall x, z, t$, we have
    \begin{align}
        q_{**} (Z | f_0 (z, t, 0), t, x) &= \delta_{\gamma(z)}(Z) \label{id1-q**}\\
        p_{**} (Y | \gamma(z), t) &= \delta_{f_0 (z, t, 0)} (Y) \label{id1-p**}
    \end{align}
    where $f_0(z, t, \epsilon_Y) = p_0 (Y | z, t)$ is the true SCM and $\gamma$ is any injective transformation of $z$ (e.g. $\gamma(z) = z$). Equation \ref{id1-q**} is well-defined by Assumption \ref{assumption-consistency} and \ref{assumption-uniqueness}, which guarantees that $f_0 (z, t, 0) \neq f_0 (z', t, 0)$ for $z \neq z'$; Equation \ref{id1-p**} is well-defined due to the injectiveness of $\gamma$. $q_{**} = q_*$, $p_{**} = p_*$ elsewhere. It follows that
    \begin{align}
        &\quad \mathbb E_{p_0 (d)} D_\mathrm{KL} \left[ q_{**} (\cdot | y, t, x) \parallel q_{**} (\cdot | y', t', x) \right] \nonumber\\
        &= \mathbb E_{p_0 (x, z, t, t'), p^*} D_\mathrm{KL} \left[ q_* \circo p^* (z, t; x) \parallel q_* \circo p^* (z, t'; x) \right] \\
        &= \mathbb E_{p_0 (x, z, t, t')} D_\mathrm{KL} \left[ q_{**} (\cdot | f_0 (z, t, 0), t, x) \parallel q_{**} (\cdot | f_0 (z, t', 0), t', x) \right] \\
        &= \mathbb E_{p_0 (x, z, t, t')} D_\mathrm{KL} \left[ \delta_{\gamma(z)}(\cdot) \parallel \delta_{\gamma(z)}(\cdot) \right] \\
        &< \mathbb E_{p_0 (d)} D_\mathrm{KL} \left[ q_* (\cdot | y, t, x) \parallel q_* (\cdot | y', t', x) \right]
    \end{align}
    by Assumption \ref{assumption-consistency}, in the meantime 
    \begin{align}
        &\quad \mathbb E_{p_0 (d)} \mathbb E_{q_{**} (z | y, t, x)} \log \left[ p_{**} (y | z, t) \right] \nonumber\\
        &= \mathbb E_{p_0 (x, z, t), p_0 (y | z, t)} \mathbb E_{q_{**} (z_{**} | y, t, x)} \log \left[ p_{**} (y | z_{**}, t) \right] \\
        &= \mathbb E_{p_0 (x, z, t)} \mathbb E_{q_{**} (z_{**} | f_0 (z, t, 0), x, t)} \log \left[ p_{**} (f_0 (z, t, 0) | z_{**}, t) \right] \\
        &= \mathbb E_{p_0 (x, z, t)} \mathbb \log \left[ p_{**} (f_0 (z, t, 0) | \gamma (z), t) \right] \\
        &= \mathbb E_{p_0 (x, z, t)} \mathbb \log \left[ \delta_{f_0 (z, t, 0)} (f_0 (z, t, 0)) \right] \\
        &= \mathbb E_{p_0 (x, z, t)} \mathbb E_{q_* (z_* | f_0 (z, t, 0), x, t)} \log \left[ \delta_{f_0 (z, t, 0)} (f_0 (z, t, 0)) \right] \\
        &\geq \mathbb E_{p_0 (x, z, t)} \mathbb E_{q_* (z_* | f_0 (z, t, 0), x, t)} \log \left[ p_* (y | z_*, t) \right] \\
        &= \mathbb E_{p_0 (x, z, t), p_0 (y | z, t)} \mathbb E_{q_* (z_* | y, t, x)} \log \left[ p_* (y | z_*, t) \right] \\
        &= \mathbb E_{p_0 (d)} \mathbb E_{q_* (z | y, t, x)} \log \left[ p_* (y | z, t) \right]
    \end{align}
    by Assumption \ref{assumption-consistency}. Hence we have 
    \begin{align}
        LB(p_{**}, q_{**}) > LB(p_*, q_*)
    \end{align}
    which contradicts $(p_*, q_*) \in \operatorname*{arg\,max}_{p_., q_.} LB(p_., q_.)$. Therefore $(p_*, q_*)$ must satisfy 
    \begin{align}
        \mathbb E_{p_0 (d)} D_\mathrm{KL} \left[ q_* (\cdot | y, t, x) \parallel q_* (\cdot | y', t', x) \right] = 0
    \end{align}
    and it follows that $\tilde{q}_*$ disentangles $Z$ and $T$ by Lemma \ref{lemma:disentanglement}.
\end{proof}

\subsubsection{Proof of Proposition \ref{thm:identifiability-2}}
\label{proof:identifiability-2}

\begin{proof}
    The strategy is similar to the proof of Proposition \ref{thm:identifiability-1}. We first prove a statement similar to Lemma \ref{lemma:disentanglement} that states a sufficient condition for empirical disentanglement (\textbf{sufficiency}), and then argue that this condition is attainable by optimizing $J_{p_\mathrm{data}}(p_., q_.)$ (\textbf{attainability}).

    \paragraph{Sufficiency} We prove that if 
    \begin{align}
        \label{eq:emp-disentanglement-suff-cond}
        \mathbb E_{p_\mathrm{data}(x, t, y), p_\mathrm{data}(t' | x), p_\mathrm{data}(t'' | x)} D_\mathrm{KL} \left[ q (\cdot | y'_., t', x) \parallel q (\cdot | y''_., t'', x) \right] = 0
    \end{align}
    where $y'_. = \mathbb E_{p_.(\cdot | \mathbb E_{q(\cdot | y, t, x)} Z, t')} Y$ and $y''_. = \mathbb E_{p_.(\cdot | \mathbb E_{q(\cdot | y, t, x)} Z, t'')} Y$, then $(\tilde{g}_., \tilde{q})$ disentangles $Z$ and $T$ under $p_e = \int_y e \cdot p_\mathrm{data}$.
    
    Let $S_I = Z$ and $S_{\backslash I} = T$. On one hand, notice that for any $x$, $z$ s.t. $p_e(x, z) > 0$, we have $p_e(t | x, z) > 0$ only if $p_e(x, z, t) > 0$ only if $p_e(x, t) > 0$ only if $p_\mathrm{data}(x, t) > 0$ only if $p_\mathrm{data}(t | x) > 0$ for any $t$. It follows that 
    \begin{align}
        &\quad \mathbb E_{p_e, \tilde{g}_.} \lVert \tilde{q}_I \circo \tilde{g}_.(z, t'; x) - \tilde{q}_I \circo \tilde{g}_.(z, t''; x) \rVert \nonumber\\
        &= \mathbb E_{p_\mathrm{data} (x, t, y), e(z | x, t, y), p_e(t', t'' | x, z), \tilde{g}_.} \lVert \tilde{q}_I \circo \tilde{g}_.(z, t'; x) - \tilde{q}_I \circo \tilde{g}_.(z, t''; x) \rVert \\
        &= \mathbb E_{p_\mathrm{data} (x, t, y), p_e(t', t'' | x, \mathbb E_{q(\cdot | y, t, x)} Z)} \lVert \tilde{q}_I(y'_., t', x) - \tilde{q}_I(y''_., t'', x) \rVert \\
        &= \mathbb E_{p_\mathrm{data} (x, t, y), p_e(t', t'' | x, \mathbb E_{q(\cdot | y, t, x)} Z)} \lVert q(\cdot | y'_., t', x) - q(\cdot | y''_., t'', x) \rVert \\
        & \leq \mathbb E_{p_\mathrm{data}, p_e} \sqrt{\frac{1}{2} D_\mathrm{KL} \left[ q (\cdot | y'_., t', x) \parallel q (\cdot | y''_., t'', x) \right]} \quad \text{(Pinsker's inequality)}\\
        &= 0,
    \end{align}
    since Equation \ref{eq:emp-disentanglement-suff-cond} implies that $D_\mathrm{KL} [ q (\cdot | y'_., t', x) \parallel q (\cdot | y''_., t'', x) ] = 0$ a.e. with respect to $p_\mathrm{data}$. Hence $(\tilde{g}_., \tilde{q})$ is consistent with $Z$ under $p_e$. On the other hand, $(\tilde{g}_., \tilde{q})$ is restricted to $Z$ under $p_e$ by design:
    \begin{align}
        &\quad \mathbb E_{p_e, \tilde{g}_.} \lVert \tilde{q}_{\backslash I} \circo \tilde{g}_.(z', t; x) - \tilde{q}_{\backslash I} \circo \tilde{g}_.(z'', t; x) \rVert \nonumber\\
        &= \mathbb E_{p_e(x, t), p_e(z', z'' | x, t), \tilde{g}_.} \lVert \tilde{q}_{\backslash I} \circo \tilde{g}_.(z', t; x) - \tilde{q}_{\backslash I} \circo \tilde{g}_.(z'', t; x) \rVert \\
        &= \mathbb E_{p_e(x, t), p_e(z', z'' | x, t)} \lVert \tilde{q}_{\backslash I}(\mathbb E_{p_.(\cdot | z', t)} Y, t, x) - \tilde{q}_{\backslash I}(\mathbb E_{p_.(\cdot | z'', t)} Y, t, x) \rVert \\
        &= \mathbb E_p \lVert \delta_t(\cdot) - \delta_t(\cdot) \rVert \\
        &= 0.
    \end{align}
    Therefore $(\tilde{g}_., \tilde{q})$ disentangles $Z$ and $T$ under $\int_y e \cdot p_\mathrm{data}$.

    \paragraph{Attainability} Let $A=\{(x_{k_i}, t_{k_i}, y_{k_i})\} |_{i=1}^m$ be the set of all unique entries in $\{(x_k, t_k, y_k)\} |_{k=1}^n$. 
    Suppose that $(p_*, q_*) \in \operatorname*{arg\,max}_{p_., q_.} J_{p_\mathrm{data}}(p_., q_.)$ but 
    \begin{align}
        \mathbb E_{p_\mathrm{data}(x, t, y), p_\mathrm{data}(t' | x), p_\mathrm{data}(t'' | x)} D_\mathrm{KL} \left[ q_* (\cdot | y'_*, t', x) \parallel q_* (\cdot | y''_*, t'', x) \right] > 0,
    \end{align}
    where $y'_* = \mathbb E_{p_*(\cdot | \mathbb E_{q_*(\cdot | y, t, x)} Z, t')} Y$ and $y''_* = \mathbb E_{p_*(\cdot | \mathbb E_{q_*(\cdot | y, t, x)} Z, t'')} Y$, i.e. $\exists l: 1 \leq l \leq n$ and $\exists t^{(1)}, t^{(2)}: p_\mathrm{data}(t^{(1)} | x_l) \cdot p_\mathrm{data}(t^{(2)} | x_l) > 0$ such that 
    \begin{align}
        \label{attain-suppose}
        D_\mathrm{KL} \left[ q (\cdot | y^{(1)}_*, t^{(1)}, x_l) \parallel q (\cdot | y^{(2)}_*, t^{(2)}, x_l) \right] > 0
    \end{align} where $y^{(1)}_* = \mathbb E_{p_*(\cdot | \mathbb E_{q_*(\cdot | y_l, t_l, x_l)} Z, t^{(1)})} Y$ and $y^{(2)}_* = \mathbb E_{p_*(\cdot | \mathbb E_{q_*(\cdot | y_l, t_l, x_l)} Z, t^{(2)})} Y$. Then we consider model $(p_{**}, q_{**})$ such that $\forall (x, t, y) \in A$, we have
    \begin{align}
        q_{**} (Z | \mathbb E_{p_*(\cdot | \mathbb E_{q_*(\cdot | y, t, x)} Z, \alpha)} Y, \alpha, x) = q_* (Z | y, t, x) \label{id2-q**}
    \end{align}
    for any $\alpha$ s.t. $p_\mathrm{data}(T=\alpha | x) > 0$. Equation \ref{id2-q**} is well-defined since 
    \begin{align}
        &\quad \mathbb E_{p_*(\cdot | \mathbb E_{q_*(\cdot | y, t, x)} Z, \alpha)} Y = \mathbb E_{p_*(\cdot | \mathbb E_{q_*(\cdot | y', t', x)} Z, \alpha)} Y \nonumber\\
        &\Leftrightarrow g_*(Y | \mathbb E_{q_*(\cdot | y, t, x)} Z, \alpha) = g_*(Y | \mathbb E_{q_*(\cdot | y', t', x)} Z, \alpha) \Rightarrow \mathbb E_{q_*(\cdot | y, t, x)} Z = \mathbb E_{q_*(\cdot | y', t', x)} Z \\
        &\Leftrightarrow e_*(Z | y, t, x) = e_*(Z | y', t', x) \Rightarrow (y, t) = (y', t') \Rightarrow q_* (Z | y, t, x) = q_* (Z | y', t', x)
    \end{align}
    by uniqueness condition 1) and the construction of $\{ q_. \}$. $q_{**} = q_*$ elsewhere and $p_{**} = p_*$ everywhere. 
    
    Firstly, we show that $(p_{**}, q_{**})$ is still a member of the model class of interest. Since $q_{**}$ is still normally distributed with unit variance by definition and $g_{**}$ still satisfies uniqueness condition 1) by definition, we only need to show that $(g_{**}, e_{**})$ still satisfies uniqueness condition 2). This is because $\forall x, t, y, t': p_\mathrm{data}(x, t, y) \cdot p_\mathrm{data}(t' | x) > 0$, we have $(x, t, y) \in A$ and thus $(x, t, y) = (x, \alpha, \mathbb E_{p_*(\cdot | \mathbb E_{q_*(\cdot | y_{k_i}, t_{k_i}, x_{k_i})} Z, \alpha)} Y)$ for some $(x_{k_i}, t_{k_i}, y_{k_i}) \in A$ only if 
    \begin{align}
        (x, \alpha, \mathbb E_{p_*(\cdot | \mathbb E_{q_*(\cdot | y_{k_i}, t_{k_i}, x_{k_i})} Z, \alpha)} Y) = (x, \alpha, \mathbb E_{g_*(\cdot | z, t') e_* (z | y_{k_i}, t_{k_i}, x_{k_i})} Y) \in A
    \end{align}
    only if $(x, \alpha, \mathbb E_{p_*(\cdot | \mathbb E_{q_*(\cdot | y_{k_i}, t_{k_i}, x_{k_i})} Z, \alpha)} Y) = (x_{k_i}, t_{k_i}, y_{k_i})$ by the fact that $(g_*, e_*)$ satisfies uniqueness condition 2). In this case, we have $(x, t, y) = (x_{k_i}, t_{k_i}, y_{k_i})$ and 
    \begin{align}
        q_{**}(Z | y, t, x) &= q_{**} (Z | \mathbb E_{p_*(\cdot | \mathbb E_{q_*(\cdot | y_{k_i}, t_{k_i}, x_{k_i})} Z, \alpha)} Y, \alpha, x) \\
        &= q_*(Z | y_{k_i}, t_{k_i}, x_{k_i}) = q_*(Z | y, t, x).
    \end{align}
    If $(x, t, y) \neq (x, \alpha, \mathbb E_{p_*(\cdot | \mathbb E_{q_*(\cdot | y_{k_i}, t_{k_i}, x_{k_i})} Z, \alpha)} Y)$ for any $(x_{k_i}, t_{k_i}, y_{k_i}) \in A$, then $q_{**}(Z | y, t, x) = q_*(Z | y, t, x)$ by definition. Therefore we have $q_{**}(Z | y, t, x) = q_*(Z | y, t, x)$. It follows that $y'_{**} = \mathbb E_{g_{**}(\cdot | z, t') e_{**}(z | y, t, x)} Y = \mathbb E_{g_*(\cdot | z, t') e_*(z | y, t, x)} Y = y'_*$ and $(g_{**}, e_{**})$ satisfies uniqueness condition 2) by the fact that $(g_*, e_*)$ satisfies uniqueness condition 2).

    Secondly, we show that $(p_*, q_*)$ is no longer optimal, thus raise a contradiction. On one hand, we must have 
    \begin{align}
        &\quad \mathbb E_{p_\mathrm{data}} D_\mathrm{KL} \left[ q_* (\cdot | y, t, x) \parallel q_* (\cdot | y'_*, t', x) \right] \nonumber\\
        &\geq p_\mathrm{data} (x_l, t_l, y_l) \left\{ p_\mathrm{data} (t^{(1)} | x_l) D_\mathrm{KL} \left[ q_* (\cdot | y_l, t_l, x_l) \parallel q_* (\cdot | y^{(1)}_*, t^{(1)}, x_l) \right] \right. \nonumber\\
        &\quad \left. + p_\mathrm{data} (t^{(2)} | x_l) D_\mathrm{KL} \left[ q_* (\cdot | y_l, t_l, x_l) \parallel q_* (\cdot | y^{(2)}_*, t^{(2)}, x_l) \right] \right\} \\
        & > 0,
    \end{align}
    otherwise we would have $q_* (Z | y_l, t_l, x_l) = q_* (Z | y^{(1)}_*, t^{(1)}, x_l) = q_* (Z | y^{(2)}_*, t^{(2)}, x_l)$ a.e. which contradicts Equation \ref{attain-suppose}. Hence 
    \begin{align}
        &\quad \mathbb E_{p_\mathrm{data}} D_\mathrm{KL} \left[ q_{**} (\cdot | y, t, x) \parallel q_{**} (\cdot | y'_{**}, t', x) \right] \nonumber\\
        &= \sum_{(x, t, y) \in A} p_\mathrm{data} (x, t, y) \sum_{t'} p_\mathrm{data} (t' | x) \cdot D_\mathrm{KL} \left[ q_* (\cdot | y, t, x) \parallel q_{**} (\cdot | y'_{**}, t', x) \right] \\
        &= \sum_{(x, t, y) \in A} p_\mathrm{data} (x, t, y) \sum_{t'} p_\mathrm{data} (t' | x) \cdot D_\mathrm{KL} \left[ q_* (\cdot | y, t, x) \parallel q_* (\cdot | y, t, x) \right] \\
        &= 0 < \mathbb E_{p_\mathrm{data}} D_\mathrm{KL} \left[ q_* (\cdot | y, t, x) \parallel q_* (\cdot | y'_*, t', x) \right]
    \end{align}
    by Equation \ref{id2-q**} and the fact that $q_{**}(Z | y, t, x) = q_*(Z | y, t, x)$ for any $(x, t, y) \in A$ as shown above. On the other hand, we have 
    \begin{align}
        &\quad \mathbb E_{p_\mathrm{data}} \left\{ \mathbb E_{q_{**} (z | y, t, x)} \log \left[ p_{**} (y | z, t) \right] + \log \left[ \hat{p} (y'_{**} | x, t') \right] \right\} \nonumber\\
        &= \mathbb E_{p_\mathrm{data}} \left\{ \mathbb E_{q_* (z | y, t, x)} \log \left[ p_* (y | z, t) \right] + \log \left[ \hat{p} (y'_* | x, t') \right] \right\}
    \end{align}
    simply by the construction of $p_{**}$ and the fact that $q_{**}(Z | y, t, x) = q_*(Z | y, t, x)$, $y'_{**} = y'_*$ for any $(x, t, y) \in A$ as shown above. Hence we have 
    \begin{align}
        J_{p_\mathrm{data}}(p_{**}, q_{**}) > J_{p_\mathrm{data}}(p_*, q_*)
    \end{align}
    which contradicts $(p_*, q_*) \in \operatorname*{arg\,max}_{p_., q_.} J_{p_\mathrm{data}}(p_., q_.)$. Therefore $(p_*, q_*)$ must satisfy 
    \begin{align}
        \mathbb E_{p_\mathrm{data}(x, t, y), p_\mathrm{data}(t' | x), p_\mathrm{data}(t'' | x)} D_\mathrm{KL} \left[ q_* (\cdot | y'_*, t', x) \parallel q_* (\cdot | y''_*, t'', x) \right] = 0
    \end{align}
    and it follows that $(\tilde{g}_*, \tilde{q}_*)$ disentangles $Z$ and $T$ under $\int_y e_* p_\mathrm{data}$ by \textbf{sufficiency}.
\end{proof}

\subsubsection{Proof of Corollary \ref{thm:identifiability-2.2}}

\begin{proof}
    By the proof of Proposition \ref{thm:identifiability-2} (attainability), we have
    \begin{align}
        \mathbb E_{p_\mathrm{data}(x, t, y), p_\mathrm{data}(t' | x), p_\mathrm{data}(t'' | x)} D_\mathrm{KL} \left[ q_* (\cdot | y'_*, t', x) \parallel q_* (\cdot | y''_*, t'', x) \right] = 0
    \end{align}
    where $y'_* = \mathbb E_{p_*(\cdot | \mathbb E_{q_*(\cdot | y, t, x)} Z, t')} Y$ and $y''_* = \mathbb E_{p_*(\cdot | \mathbb E_{q_*(\cdot | y, t, x)} Z, t'')} Y$, hence $\mathbb E_{q_*(\cdot | y'_*, t', x)} Z = \mathbb E_{q_*(\cdot | y''_*, t'', x)} Z$ a.e. on $p_\mathrm{data}$ and it follows that
    \begin{align}
        &\mathbb E_{p_\mathrm{data}(x, t, y), p_\mathrm{data}(t' | x), p_\mathrm{data}(t'' | x)} D_\mathrm{KL} \left[ e_* (\cdot | y'_*, t', x) \parallel e_* (\cdot | y''_*, t'', x) \right] \nonumber \\
        &= \mathbb E_{p_\mathrm{data}(x, t, y), p_\mathrm{data}(t' | x), p_\mathrm{data}(t'' | x)} D_\mathrm{KL} \left[ \delta_{\mathbb E_{q_*(\cdot | y'_*, t', x)} Z} (\cdot) \parallel \delta_{\mathbb E_{q_*(\cdot | y''_*, t'', x)} Z}  (\cdot) \right] = 0.
    \end{align}
    Notice that $y'_* = \mathbb E_{p_*(\cdot | \mathbb E_{e_*(\cdot | y, t, x)} Z, t')} Y$ and $y''_* = \mathbb E_{p_*(\cdot | \mathbb E_{e_*(\cdot | y, t, x)} Z, t'')} Y$, we have $(\tilde{g}_*, \tilde{e}_*)$ disentangles $Z$ and $T$ under $\int_y e_* p_\mathrm{data}$ by the proof of Proposition \ref{thm:identifiability-2} (sufficiency).
\end{proof}


\subsection{Marginal Estimator}

\subsubsection{Proof of Theorem \ref{variational-ATT}}
\label{variational-ATT_proof}

\begin{proof}
    $\Psi(p)$ has the identification $\Psi(p) = \mathbb E_p [\mathbb E_p [Y' | Z, X, T'=\alpha ]] = \mathbb E_p [\mathbb E_p [Y' | Z, T'=\alpha ]]$ under Figure \ref{causal_diagram}. Following \citet{van2000asymptotic}, we define a path $p_\epsilon(\Lambda)=p(\Lambda)(1+\epsilon S(\Lambda))$ on density $p$ of random vector $\Lambda$ as a submodel that passes through $p$ at $\epsilon=0$ in the direction of the score $S(\Lambda)=\frac{d}{d\epsilon} \log [p_\epsilon(\Lambda)] \Big{\rvert}_{\epsilon=0}$. Following the key identity presented in \citet{levy2019tutorial}:
    \begin{align}
        &\quad \frac{d}{d\epsilon} p_\epsilon(\lambda_i | pa(\Lambda_i)=\bar{\lambda}_{i-1}) \Big{\rvert}_{\epsilon=0} \nonumber\\ 
        &= p(\lambda_i | pa(\Lambda_i)=\bar{\lambda}_{i-1}) (\mathbb E[S(\Lambda) | \Lambda_i=\lambda_i, pa(\Lambda_i)=\bar{\lambda}_{i-1}] - \mathbb E[S(\Lambda) | pa(\Lambda_i)=\bar{\lambda}_{i-1}]) \label{EIC_key_identity}
    \end{align}
    where $pa(\Lambda_i)$ denotes the parent nodes of variable $\Lambda_i$, we have
    \begin{align}
        \frac{d}{d\epsilon} \Psi(p_\epsilon) \Big{\rvert}_{\epsilon=0} &= \frac{d}{d\epsilon} \Big{\rvert}_{\epsilon=0} \mathbb E_{p_\epsilon} \left[ \mathbb E_{p_\epsilon} \left[ Y' | Z, T'=\alpha \right] \right] \\
        &= \frac{d}{d\epsilon} \Big{\rvert}_{\epsilon=0} \int_{y', z, x} y' \left[ p_\epsilon(y' | z, T'=\alpha) p_\epsilon(z, x) \right] \\
        &= \int_{y', z, x} y' \frac{d}{d\epsilon} \Big{\rvert}_{\epsilon=0} \left[ p_\epsilon(y' | z, T'=\alpha) p_\epsilon(z, x) \right] \quad \text{(dominated convergence)} \label{EIC_p-eps_start}\\
        &= \int_{y', z, x} y' p(z, x) \frac{d}{d\epsilon} \Big{\rvert}_{\epsilon=0} p_\epsilon(y' | z, T'=\alpha) \\
        &+ \int_{y', z, x} y' p(y' | z, T'=\alpha) \frac{d}{d\epsilon} \Big{\rvert}_{\epsilon=0} p_\epsilon(z, x) \\
        &= \int_w I(t'=\alpha) \frac{p(t' | x)}{p(t' | x)} y' p(y, t | z, x) p(z, x) \frac{d}{d\epsilon} \Big{\rvert}_{\epsilon=0} p_\epsilon(y' | z, t')  \nonumber\\
        &\quad + \int_{y', z, x} y' p(y' | z, T'=\alpha) \frac{d}{d\epsilon} \Big{\rvert}_{\epsilon=0} p_\epsilon(z, x) \label{EIC_p_start}\\
        &= \int_w \frac{I(t'=a)}{p(t' | x)} y' p(y', t' | z, x) p(y, t | z, x) p(z, x) \left\{ S(w) - \mathbb E\left[ S(W) | y, z, x, t, t' \right]\right\} \nonumber\\
        &\quad + \int_{y', z, x} y' p(y' | z, T'=\alpha) p(z, x) \left\{ \mathbb E\left[ S(W) | z, x \right] - \mathbb E\left[ S(W) \right]\right\} \\
        &= \int_w \frac{I(t=a)}{p(t | x)} y p(y, t | z, x) p(y', t' | z, x) p(z, x) \left\{ S(w) - \mathbb E\left[ S(W) | y', z, x, t', t \right]\right\}  \nonumber\\
        &\quad + \int_{y', z, x} y' p(y' | z, T'=\alpha) p(z, x) \left\{ \mathbb E\left[ S(W) | z, x \right] - \mathbb E\left[ S(W) \right]\right\} \\
        &= \int_w S(w) \cdot \frac{I(t=\alpha)}{p(t | x)} y p(w) \nonumber\\
        &\quad - \int_w \mathbb E\left[ S(W) | y', z, x, t, t' \right] p(y', z, x, t, t') \cdot \frac{I(t=\alpha)}{p(t | x)} y p(y | z, t) \nonumber\\
        &\quad + \int_{y', z, x} \mathbb E\left[ S(W) | z, x \right] p(z, x) \cdot y' p(y' | z, T'=\alpha) \nonumber\\
        &\quad - \int_{y', z, x} \mathbb E\left[ S(W) \right] \cdot y' p(y' | z, T'=\alpha) p(z, x) \\
        &= \int_w S(w) \left\{\frac{I(t=\alpha)}{p(t | x)} (y - \mathbb E\left[ Y | z, t \right]) + \mathbb E\left[ Y' | z, T'=\alpha \right] - \Psi \right\} p(w)
    \end{align}
    by the assumption of Theorem \ref{variational-ATT} and factorization according to Figure \ref{causal_diagram}. Hence 
    \begin{align}
        \frac{d}{d\epsilon}\Psi(p_\epsilon) \Big{\rvert}_{\epsilon=0} = \left \langle S(W), \frac{I(T = \alpha)}{p(T | X)} (Y - \mathbb E_p\left[ Y | Z, T \right]) + \mathbb E_p\left[ Y' | Z, T'=\alpha \right] - \Psi \right \rangle_{L^2(\Omega; E)}
    \end{align}
    and we have $\tilde{\psi}_p = I(T = \alpha) / p(T | X) \cdot (Y - \mathbb E_p[ Y | Z, T ]) + \mathbb E_p[ Y' | Z, T'=\alpha ] - \Psi$.
\end{proof}

\subsubsection{Proof of Corollary \ref{variational-ATT_covar-spec}}
\label{variational-ATT_covar-spec_proof}
\begin{proof}
    The proof is a simple extension on the proof of Theorem \ref{variational-ATT}. We have
    \begin{align}
        \frac{d}{d\epsilon} \Xi(p_\epsilon) \Big{\rvert}_{\epsilon=0} &= \frac{d}{d\epsilon} \Big{\rvert}_{\epsilon=0} \mathbb E_{p_\epsilon} \left[ \mathbb E_{p_\epsilon} \left[ Y' | Z, T'=\alpha \right] | X=c \right] \\
        &= \frac{d}{d\epsilon} \Big{\rvert}_{\epsilon=0} \int_{y', z, x} y' \left[ p_\epsilon(y' | z, T'=\alpha) p_\epsilon(z | X=c) \right].
    \end{align}
    Hence, following the same derivation, $p_\epsilon(z, x)$ in Equation \ref{EIC_p-eps_start} and onwards becomes $p_\epsilon(z | X=c)$, while $p(z, x)$ in Equation \ref{EIC_p_start} and onwards becomes $I(x = c) / p(x) \cdot p(z, x)$. Therefore, we have $\tilde{\xi}_p(W) = I(X = c) / p(X) \cdot  \tilde{\psi}_p(W)$.
\end{proof}

\section{Model Architecture for Image Generation Tasks}
\label{sec:architecture}

\begin{figure}[ht!]
    \centering
    \includegraphics[width=0.4\linewidth]{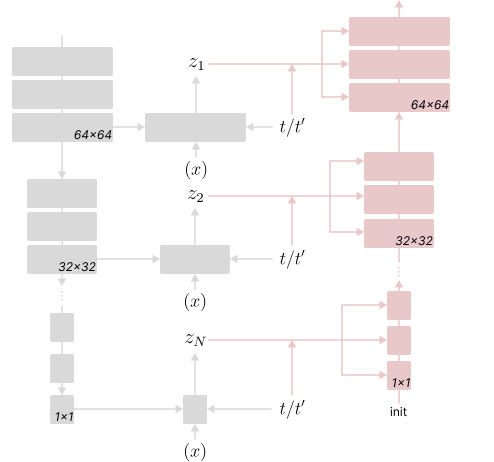}
    \caption{Model architecture of the latent recognition model (grey) and outcome construction model (pink) for image generation tasks.}
    \label{fig:hvci-architecture}
\end{figure}

The specific model architecture used for image generation can be found in Figure \ref{fig:hvci-architecture}. Note that in the image generation tasks in our experiments, we treated every condition as intervenable, thus every condition belongs in $T$ rather than $X$. If this is not the case, the non-intervenable conditions would compose covariates $X$ and would only be inputted to the encoding model but not the decoding model.

To incorporate $t$ ($t'$) and $x$ into the convolutional layers, we first embed them into vector representations (tabular embedding for categorical variables; sinusoidal embedding with non-linear mapping for continuous variable), then repeat and expand each value of the vectors to match the input resolutions of the corresponding blocks and concatenate them to image representations as extra channels, similar to \citet{monteiro2023measuring}. Note that there are other ways to conduct image-vector incorporation such as the scale-shift approach commonly used in diffusion models to incorporate time step variable \citep{sanchez2022diffusion}. Each convolutional block (each rectangle in Figure \ref{fig:hvci-architecture}) contains a 1$\times$1 embedding layer, a 3$\times$3 (or 1$\times$1 if resolution is lower than 3$\times$3) botteleneck layer, and a 3$\times$3 (or 1$\times$1 if resolution is lower than 3$\times$3) output layer.

\section{Detaching Patterns}
\label{detaching-pattern}
In practice, there are a few optional gradient detaching options when evaluating the divergence term to enhance the stability of stochastic optimization. Our default behavior is to use a copy of $q_\phi$ for the evaluation of the divergence term which is updated after every epoch, while preserving the gradient path and gradient computation for $y'_{\theta,\phi}$. This technique is analogous to the handling of surrogate objectives commonly applied in reinforcement learning \citep{lillicrap2015continuous,van2016deep,schulman2017proximal}, and the copy of $q_\phi$ is seen as the target network. In addition, we provide the option of detaching $\phi$ from $y'_{\theta,\phi}$ (focusing on the second purpose described in divergence interpretation: preservation of individuality in decoding) and the option of not detaching the gradient computation for $q_\phi$ (reflecting the first purpose in divergence interpretation: recognition of mutual features in encoding) during the evaluation of the latent divergence term.

\section{Benchmark Adaptations}
\label{benchmark-adaptation}

\subsection{Autoencoder}
\label{ae-adaptation}
The adapted autoencoder reconstructs the outcome during training similar to a generic autoencoder, but takes treatment and covariates as additional inputs. During test time, we simply plug in the counterfactual treatments along with factual outcomes and covariates to generate the counterfactual outcome predictions.

\subsection{GANITE}
\label{ganite-adaptation}
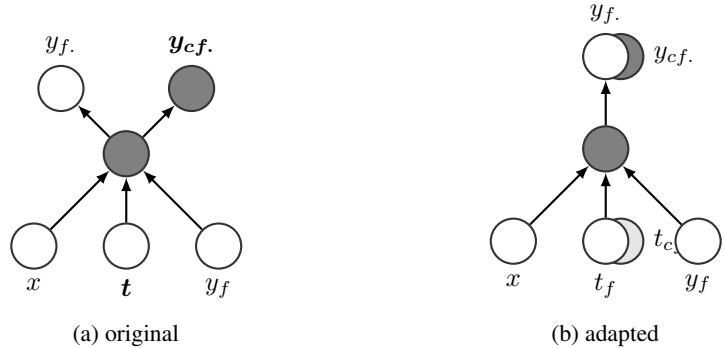
\begin{figure}[hbt!]
    \centering
    \begin{subfigure}[b]{0.45\textwidth}
        \centering
        \begin{tikzpicture}
            \tikzstyle{main}=[circle, minimum size = 6mm, thick, draw =black!80, node distance = 6mm]
            \tikzstyle{connect}=[-latex, thick]
            \tikzstyle{box}=[rectangle, draw=black!100]
              \node[main, fill = white!100] (x) [label=below:$x$] { };
              \node[main, fill = white!100] (t) [right=of x,label=below:$\bm{t}$] { };
              \node[main, fill = white!100] (y) [right=of t,label=below:$y_f$] { };
              \node[main, fill = black!50] (z) [above=of t] { };
              \node[main, fill = white!100] (y_re) [above left=of z,label=above:$y_{f.}$] {};
              \node[main, fill = black!50] (y_cf) [above right=of z,label=above:$\bm{y_{cf.}}$] { };
              \path (x) edge [connect] (z)
                    (t) edge [connect] (z)
            		(y) edge [connect] (z)
            		(z) edge [connect] (y_re)
            		(z) edge [connect] (y_cf);
        \end{tikzpicture}
        \caption{original}
        \label{ganite-original}
    \end{subfigure}
    \begin{subfigure}[b]{0.45\textwidth}
        \centering
        \begin{tikzpicture}
            \tikzstyle{main}=[circle, minimum size = 6mm, thick, draw =black!80, node distance = 6mm]
            \tikzstyle{connect}=[-latex, thick]
            \tikzstyle{box}=[rectangle, draw=black!100]
              \node[main, fill = white!100] (x) [label=below:$x$] { };
              \node[main, fill = white!100] (t_f) [right=of x,label=below:$t_f$] { };
              \begin{scope}[on background layer]
                \node[main, fill = black!10] (t_cf) [right=of x,label=right:$t_{cf}$, xshift=2mm] { };
              \end{scope}
              \node[main, fill = white!100] (y) [right=of t_f,label=below:$y_f$] { };
              \node[main, fill = black!50] (z) [above=of t_f] {};
              \node[main, fill = white!100] (y_re) [above=of z,label=above:$y_{f.}$] {};
              \begin{scope}[on background layer]
                \node[main, fill = black!50] (y_cf) [above=of z,label=right:$y_{cf.}$, xshift=2mm] { };
              \end{scope}
              \path (x) edge [connect] (z)
                    (t_f) edge [connect] (z)
            		(y) edge [connect] (z)
            		(z) edge [connect] (y_re);
        \end{tikzpicture}
        \caption{adapted}
        \label{ganite-adapted}
    \end{subfigure}
    \caption{GANITE's counterfactual generator. $t_{cf}$ is a random sample of $\bm{t}$, passed into the generator as a part of the input $(x, t_{cf}, y_f)$, separately from input $(x, t_f, y_f)$ of the factual generation.}
    \label{ganite}
\end{figure}

GANITE \citep{yoon2018ganite}'s counterfactual generator does not scale with a combination of high-dimensional outcome and multi-level treatment, thus here we only input one randomly sampled counterfactual treatment to the generator and correspondingly construct one counterfactual outcome for each sample. See Figure \ref{ganite} for the original and adapted structure of the model.

The discriminator predicts the logits $l_f$, $l_{cf}$ of $y_f$, $y_{cf.}$ separately. The cross entropy loss of $(l_f, l_{cf})$ against $(1, 0)$ is then calculated.

\section{Experiment Settings}
\label{setting:experiment}

In this section, we describe the settings of our experiments in detail. For complete information on hyperparameter settings, see our codebase at \url{https://github.com/yulun-rayn/variational-causal-inference}.

\subsection{Single-cell Perturbation Datasets}
\label{setting:single-cell-pert}

For both datasets in our experiments, two thousand most variable genes were selected for training and testing. Marson has 2,000 dimensional outcomes, 3 categorical covariates (cell type, donor indicator, stimulation indicator) and 1 categorical treatment (target gene), Sciplex has 2,000 dimensional outcomes, 2 categorical covariates (cell type, replicate indicator) and 1 categorical treatment (used drug). Data with certain treatment-covariate combinations are held out as the out-of-distribution (OOD) set and the rest are split into training and validation set with a four-to-one ratio.

\paragraph{Out-of-Distribution Selections}
We randomly select a covariate category (e.g. a cell type) and hold out all cells in this category that received one of the twenty perturbations whose effects are the hardest to predict. We use these held-out data to compose the out-of-distribution (OOD) set. We computed the Euclidean distance between the pseudobulked gene expression of each perturbation against the rest of the dataset, and selected the twenty most distant ones as the hardest-to-predict perturbations. This is the same procedure carried out by \citet{lotfollahi2021learning}.

\paragraph{Differentially-Expressed Genes}
In order to evaluate the quality of the predictions on the genes that were substantially affected by the perturbations, we select sets of 50 differentially-expressed (DE) genes associated with each perturbation and separately evaluate model performance on these genes. This is the same procedure carried out by \citet{lotfollahi2021learning}.

\paragraph{Hyperparameter Settings} All common hyperparameters of all models are set to the same as the defaults of CPA \citep{lotfollahi2021learning}: an universal number of hidden dimensions $128$; number of layers $6$ (encoder $3$, decoder $3$); an universal learning rate $3^{-4}$, weight decay rate $4^{-7}$. Contrary to CPA, we use step-based learning rate decay instead of epoch-based learning rate decay, and decay step size is set to $400,000$ while decay rate remains the same at $0.1$. Batch size is $64$ for Marson and $128$ for Sciplex.


\paragraph{Other Details} We used the empirical outcome distribution (with Gaussian kernel smoother) stratified by $X$ and $T$ to estimate the covariate-specific outcome model $p(Y | X, T)$. The fully-attached detaching pattern is applied where both $y'_{\theta,\phi}$ and $q_\phi$ are not detached when evaluating the divergence term. Models are trained on Amazon web services' accelerated computing EC2 instance G4dn which contains 2nd Generation Intel Xeon Scalable Processors (Cascade Lake P-8259CL) and up to 8 NVIDIA T4 Tensor Core GPUs.

\subsection{Morpho-MNIST}
\label{setting:morpho-mnist}

We used the original Morpho-MNIST training set for model training, and the original Morpho-MNIST testing set as the observed samples for model testing. The training and testing set have 28$\times$28 dimensional outcomes, no covariates and 2 continuous treatments (thickness, intensity). Three different testing sets are counstructed based on the counterfactual treatment settings: only modifying thickness for $do($th$)$; only modifying intensity for $do($in$)$; randomly modifying thickness or intensity for mix. For each observed sample in the testing set, we randomly sample the amount of modification based on the range of the selected modification.

\paragraph{Data Augmentation} For each image in the training set (28$\times$28), we pad it by a margin of 4 pixels on each side (36$\times$36), then randomly crop a 32$\times$32 region. This is the same procedure carried out by \citet{ribeiro2023high}.

\paragraph{Hyperparameter Settings} The model width, depth and resolutions of our framework are set to the same as \citet{ribeiro2023high}: number of channels for the hierarchical encoding blocks are set to $\{16,32,64,128,256\}$ with resolutions $\{32,16,8,4,1\}$ and number of blocks $\{4,4,4,4,4\}$; number of channels for the hierarchical decoding blocks are set to $\{256,128,64,32,16\}$ with resolutions $\{1,4,8,16,32\}$ and number of blocks $\{4,4,4,4,4\}$. Each block contains a bottleneck layer with one-forth the number of channels as the input/output number of channels. Training is conducted with a batch size of $32$, an universal learning rate $1^{-4}$, and weight decay rate $4^{-5}$. Learning rate decays at epoch 100 and linearly decays to 0 at epoch 200.


\paragraph{Other Details} We used the adversarial training approach to estimate the covariate-specific outcome model $p(Y | X, T)$. The default detaching pattern (see Appendix \ref{detaching-pattern}) is applied where a copy of $q_\phi$ is used as critic to evaluate the divergence term, and updated after every epoch. Models are trained on Amazon web services' accelerated computing EC2 instance G4dn which contains 2nd Generation Intel Xeon Scalable Processors (Cascade Lake P-8259CL) and up to 8 NVIDIA T4 Tensor Core GPUs.

\subsection{CelebA-HQ}
\label{setting:celeba-hq}

The two factors of interest -- smiling and glasses, are regarded as intervenable. Therefore, the training and testing set have 64$\times$64$\times$3 dimensional outcomes, no covariates and 2 categorical treatments. The train-test split of the CelebA-HQ dataset is inherited from the original CelebA dataset.

\paragraph{Data Augmentation} For each image, we crop the 128$\times$128 center region, then randomly crop a 120$\times$120 region within the center region. Afterwards, we resize this region to 64$\times$64, and apply horizontal flip with probability $0.5$. Note that data augmentation is applied to both training and testing set in this experiment, since we are sampling and empirically examining the counterfactual results and not evaluating any metric.

\paragraph{Hyperparameter Settings} The model width and resolutions of our framework are set to the same as \citet{monteiro2023measuring}: number of channels for the hierarchical encoding blocks are set to $\{32,64,128,256,512,1024\}$ with resolutions $\{64,32,16,8,4,1\}$; number of channels for the hierarchical decoding blocks are set to $\{1024,512,256,128,64,32\}$ with resolutions $\{1,4,8,16,32,64\}$. The number of blocks is set to $\{3,12,12,6,3,3\}$ for the encoder and $\{3,3,6,12,12,3\}$ for the decoder. This is chosen to be slightly more shallow than \citet{monteiro2023measuring} ($\{4,12,12,8,4,4\}$ and $\{4,4,8,12,12,4\}$). Each block contains a bottleneck layer with one-forth the number of channels as the input/output number of channels. Training is conducted with a batch size of $32$. An universal learning rate $3^{-4}$, and weight decay rate $4^{-7}$ are inherited from the experiments with single-cell perturbation datasets. Step size for learning rate decay is $400,000$ with decay rate $0.1$.


\paragraph{Other Details} We used the adversarial training approach to estimate the covariate-specific outcome model $p(Y | X, T)$. The default detaching pattern (see Appendix \ref{detaching-pattern}) is applied where a copy of $q_\phi$ is used as critic to evaluate the divergence term, and updated after every epoch. Models are trained on Amazon web services' accelerated computing EC2 instance P3 which contains high frequency Intel Xeon Scalable Processor (Broadwell E5-2686 v4) and up to 8 NVIDIA Tesla V100 GPUs, each pairing 5,120 CUDA Cores and 640 Tensor Cores.

\section{Additional Experiments and Details}

\subsection{Marginal Estimations on Single-cell Perturbation Datasets}
\label{experiment:marginal-est}

In this section, we use the same evaluation metric as Section \ref{experiment:single-cell-pert}, but compute each $R^2$ with the robust marginal estimator and compare the results to that of the empirical mean estimator. Note that on OOD set, the robust estimator reduces to empirical mean since no perturbation-covariates combination exist in validation set. Therefore, we compute the $R^2$ of the marginal estimators using samples from the training set against the true empirical average on the validation set in these experiments. In each run, we train a VCI model for individualized outcome predictions, and calculate the evaluation metric for each marginal estimator periodically during the course of training. Table \ref{result-table2} shows the results on Marson. Note that the goal of robust estimation is to produce less biased estimators with tighter confidence bounds, hence we report the mean and standard error over five independent runs to reflect its effectiveness regarding this goal.

\begin{table}[ht!] 
  \caption{Comparison of marginal estimators on Marson \citep{schmidt2022crispr}}
  \label{result-table2}
  \centering
  \begin{tabular}{lcccc}
    \toprule
    & \multicolumn{2}{c}{All Genes} & \multicolumn{2}{c}{DE Genes}\\
    \cmidrule(r){2-3} \cmidrule(r){4-5}
    Episode & mean & robust & mean & robust \\
    \midrule
    40  & 0.9141 $\pm$ 0.0159 & \textbf{0.9343} $\pm$ \textbf{0.0080} & 0.7108 $\pm$ 0.0735 & \textbf{0.9146} $\pm$ \textbf{0.0305} \\
    80  & 0.9171 $\pm$ 0.0104 & \textbf{0.9349} $\pm$ \textbf{0.0068} & 0.7274 $\pm$ 0.0462 & \textbf{0.9163} $\pm$ \textbf{0.0254} \\
    120 & 0.9204 $\pm$ 0.0097 & \textbf{0.9352} $\pm$ \textbf{0.0063} & 0.7526 $\pm$ 0.0447 & \textbf{0.9182} $\pm$ \textbf{0.0229} \\
    160 & 0.9157 $\pm$ \textbf{0.0043} & \textbf{0.9355} $\pm$ 0.0053 & 0.7383 $\pm$ 0.0402 & \textbf{0.9203} $\pm$ \textbf{0.0199} \\
    \bottomrule
  \end{tabular}
\end{table}

As is shown in the table, the robust estimator provides a crucial adjustment to the empirical mean of model predictions especially on the hard-to-predict elements of high-dimensional vectors. Such estimation could be valuable in many contexts involving high-dimensional predictions where deep learning models might plateau at rather low ceilings.

\subsection{More Complete Ablation Study}
\label{experiment:complete-ablation}

\begin{figure}[ht!]
    \centering
    \includegraphics[width=\linewidth]{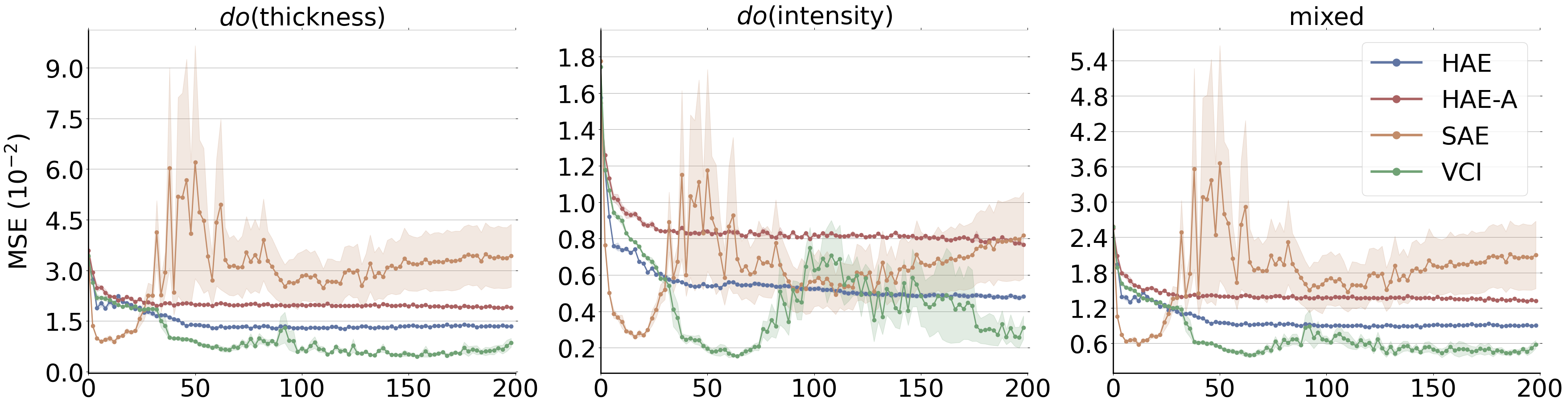}

    \caption{Ablation Study: Added HAE-A representing training VCI with the latent divergence term and without the counterfactual supervision term.}
  \label{fig:complete-morpho-ablation}
\end{figure}

Note that the latent divergence term is essentially a regularization term to counterfactual supervision, as discussed in Section \ref{variational-causal-inference}, hence it would not be meaningful to train VCI with the latent divergence term and without the counterfactual supervision term – it would just further worsen the training-inference disconnection problem described in Section \ref{introduction} without any real benefit. However, we acknowledge that presenting this case would make the ablation study more complete, and therefore have made this inclusion (marked as HAE-A) in Figure \ref{fig:complete-morpho-ablation}. As expected, HAE-A is even worse than HAE.

\subsection{Evaluation on Latent Disentanglement}
\label{experiment:disentanglement-evaluation}

The definition of disentanglement (Definition \ref{def:disentanglement}) is extended from a well-established definition of disentanglement in \citet{shu2019weakly}, based on the oracle consistency (Definition \ref{def:oracle-consistency}) of the latent representation. It is a different notion from latent disentanglement in the non-linear ICA literatures (Appendix \ref{sec:intro-disentanglement}) although sometimes sharing a non-distinguishable terminology. In this context, the best way to evaluate disentanglement is by definition, using oracle consistency, and we present the ablation results in Table \ref{result-oracle-consistency}. Note that the oracle consistency is measured by KL-divergence for computational efficiency due to the lack of analytic form for the total variational distance between Gaussians. As stated by the Pinsker's Inequality, the total variational distance is bounded once KL-divergence is controlled.

\begin{table}[ht!] 
  \caption{Oracle consistency under different strengths (denoted as coefficient $\omega_2$) on the exogenous disentanglement term.}
  \label{result-oracle-consistency}
  \centering
  \begin{tabular}{lc}
    \toprule
    $\omega_2$ & oracle consistency $\downarrow$ \\
    \midrule
    0  & 3979.86 $\pm$ 202.42 \\
    0.01  & 0.84 $\pm$ 0.13 \\
    0.02 & 0.75 $\pm$ 0.11 \\
    0.03 & 0.53 $\pm$ 0.10 \\
    0.04 & 0.51 $\pm$ 0.07 \\
    0.05 & 0.51 $\pm$ 0.08 \\
    \bottomrule
  \end{tabular}
\end{table}

\subsection{Composition, Effectiveness, and Reversibility}
\label{experiment:c-e-r}

We examine our model through the composition, effectiveness, and reversibility metrics \citep{monteiro2023measuring} and use the best benchmark model in Table \ref{result-table-morpho} as a reference. The results are shown in Table \ref{result-c-e-r}. Note that these results only serve model inspection purposes and should not be used as strict performance evaluations in favor of Table \ref{result-table-morpho}. These metrics are merely surrogate measurements of counterfactual models' axiomatic characteristics, and they are inconclusive in deciding the best-performing counterfactual model. For example, a naive model that simply returns the original image as counterfactual outcome at all times is completely useless for counterfactual inference, but will achieve the best score on 2 of these 3 metrics -- a perfect composition score of 0 and a perfect reversibility score of 0. There is a trade-off between composition and effectiveness as well as effectiveness and reversibility, and it is not conclusive which counterfactual model performs the best unless there is a clear domination of one model over the others on all three metrics.

\begin{table}[ht!] 
  \caption{Composition, effectiveness, and reversibility of ours against the best benchmark model in Table \ref{result-table-morpho}. Images are scaled to $[0, 1]$ in evaluations. All losses are on the L-2 scale (note that the factors measured by effectiveness -- thickness and intensity -- are continuous values).}
  \label{result-c-e-r}
  \centering
  \begin{tabular}{lcccc}
    \toprule
    & $\beta$ & composition $\downarrow$ & effectiveness $\downarrow$ & reversibility $\downarrow$ \\
    \midrule
    MED & 1 & 0 $\pm$ 0 & 283.99 $\pm$ 67.76 & 0.01024 $\pm$ 0.00767 \\
    MED & 3 & 0 $\pm$ 0 & 151.13 $\pm$ 86.46 & 0.00816 $\pm$ 0.00306 \\
    \hdashline
    SAE &  & 0.00048 $\pm$ 0.000084 & 189.03 $\pm$ 22.92 & 0.00102 $\pm$ 0.00019 \\
    VCI &  & 0.00026 $\pm$ 0.000055 & 17.64 $\pm$ 6.23 & 0.00076 $\pm$ 0.00015 \\
    \bottomrule
  \end{tabular}
\end{table}

As discussed in Section \ref{sec:exp-celeba-hq}, it is expected that our model does not reach a composition of 0 because the consistency assumption does not necessarily hold strictly and the model does not necessarily perform exact reconstruction in regions directly affected by the intervention, which is not an issue for the purpose of counterfactual generative modeling. Note that the composition metric does not interact with counterfactual interventions at all, while the other two metrics interact with counterfactual interventions at least in some capacity.

\subsection{More Details on Morpho-MNIST Experiments}
\label{experiment:morpho-mnist-std}

\begin{figure}[hbt!]
    \centering
    \includegraphics[width=0.8\linewidth]{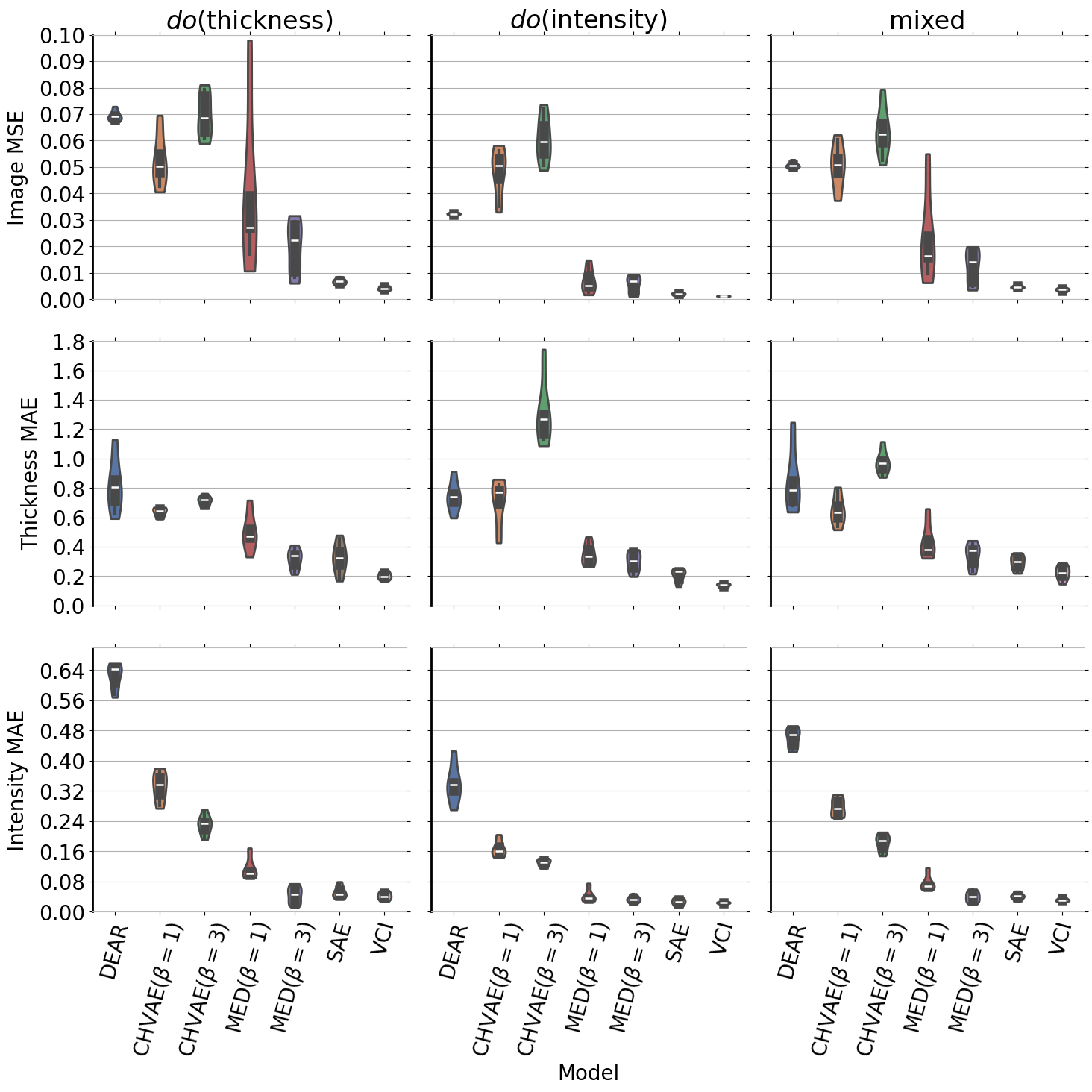}
    \caption{Violin plot of Image MSE, Thickness MAE and Intensity MAE on benchmark models and ours across five independent runs.}
    \label{fig:violin-morpho}
\end{figure}

\begin{figure}[ht!]
    \centering
    \begin{subfigure}{0.44\linewidth}
        \centering
        \begin{subfigure}{0.88\linewidth}
            \centering
            \includegraphics[width=\linewidth]{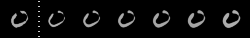}
        \end{subfigure}
    
        \begin{subfigure}{0.88\linewidth}
            \centering
            \includegraphics[width=\linewidth]{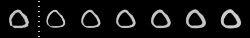}
        \end{subfigure} 
    
        \begin{subfigure}{0.88\linewidth}
            \centering
            \includegraphics[width=\linewidth]{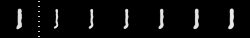}
        \end{subfigure}
    
        \begin{subfigure}{0.88\linewidth}
            \centering
            \includegraphics[width=\linewidth]{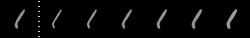}
        \end{subfigure} 
    
        \begin{subfigure}{0.88\linewidth}
            \centering
            \includegraphics[width=\linewidth]{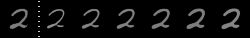}
        \end{subfigure}
    
        \begin{subfigure}{0.88\linewidth}
            \centering
            \includegraphics[width=\linewidth]{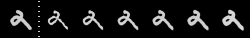}
        \end{subfigure} 
    
        \begin{subfigure}{0.88\linewidth}
            \centering
            \includegraphics[width=\linewidth]{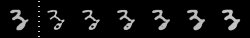}
        \end{subfigure}
    
        \begin{subfigure}{0.88\linewidth}
            \centering
            \includegraphics[width=\linewidth]{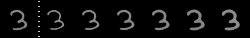}
        \end{subfigure} 
    
        \begin{subfigure}{0.88\linewidth}
            \centering
            \includegraphics[width=\linewidth]{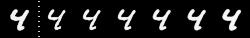}
        \end{subfigure}
    
        \begin{subfigure}{0.88\linewidth}
            \centering
            \includegraphics[width=\linewidth]{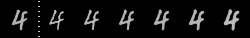}
        \end{subfigure} 
    
        \begin{subfigure}{0.88\linewidth}
            \centering
            \includegraphics[width=\linewidth]{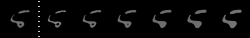}
        \end{subfigure}
    
        \begin{subfigure}{0.88\linewidth}
            \centering
            \includegraphics[width=\linewidth]{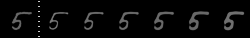}
        \end{subfigure} 
    
        \begin{subfigure}{0.88\linewidth}
            \centering
            \includegraphics[width=\linewidth]{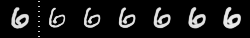}
        \end{subfigure}
    
        \begin{subfigure}{0.88\linewidth}
            \centering
            \includegraphics[width=\linewidth]{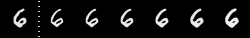}
        \end{subfigure} 
    
        \begin{subfigure}{0.88\linewidth}
            \centering
            \includegraphics[width=\linewidth]{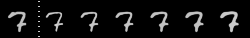}
        \end{subfigure}
    
        \begin{subfigure}{0.88\linewidth}
            \centering
            \includegraphics[width=\linewidth]{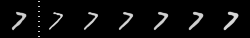}
        \end{subfigure} 
    
        \begin{subfigure}{0.88\linewidth}
            \centering
            \includegraphics[width=\linewidth]{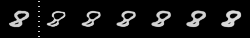}
        \end{subfigure}
    
        \begin{subfigure}{0.88\linewidth}
            \centering
            \includegraphics[width=\linewidth]{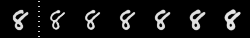}
        \end{subfigure} 
    
        \begin{subfigure}{0.88\linewidth}
            \centering
            \includegraphics[width=\linewidth]{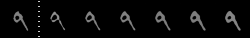}
        \end{subfigure}
    
        \begin{subfigure}{0.88\linewidth}
            \centering
            \includegraphics[width=\linewidth]{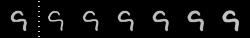}
        \end{subfigure} 
    
    \caption{$do$(thickness).}
    \label{fig:morphoMNIST-sample-thickness}
    \end{subfigure}
    \begin{subfigure}{0.44\linewidth}
        \centering
        \begin{subfigure}{0.88\linewidth}
            \centering
            \includegraphics[width=\linewidth]{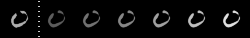}
        \end{subfigure}
    
        \begin{subfigure}{0.88\linewidth}
            \centering
            \includegraphics[width=\linewidth]{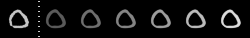}
        \end{subfigure} 
    
        \begin{subfigure}{0.88\linewidth}
            \centering
            \includegraphics[width=\linewidth]{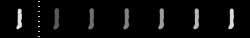}
        \end{subfigure}
    
        \begin{subfigure}{0.88\linewidth}
            \centering
            \includegraphics[width=\linewidth]{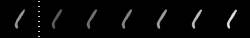}
        \end{subfigure} 
    
        \begin{subfigure}{0.88\linewidth}
            \centering
            \includegraphics[width=\linewidth]{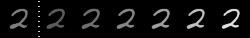}
        \end{subfigure}
    
        \begin{subfigure}{0.88\linewidth}
            \centering
            \includegraphics[width=\linewidth]{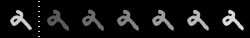}
        \end{subfigure} 
    
        \begin{subfigure}{0.88\linewidth}
            \centering
            \includegraphics[width=\linewidth]{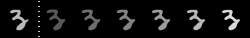}
        \end{subfigure}
    
        \begin{subfigure}{0.88\linewidth}
            \centering
            \includegraphics[width=\linewidth]{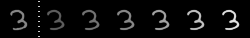}
        \end{subfigure} 
    
        \begin{subfigure}{0.88\linewidth}
            \centering
            \includegraphics[width=\linewidth]{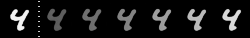}
        \end{subfigure}
    
        \begin{subfigure}{0.88\linewidth}
            \centering
            \includegraphics[width=\linewidth]{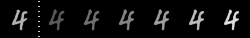}
        \end{subfigure} 
    
        \begin{subfigure}{0.88\linewidth}
            \centering
            \includegraphics[width=\linewidth]{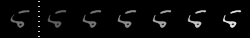}
        \end{subfigure}
    
        \begin{subfigure}{0.88\linewidth}
            \centering
            \includegraphics[width=\linewidth]{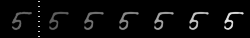}
        \end{subfigure} 
    
        \begin{subfigure}{0.88\linewidth}
            \centering
            \includegraphics[width=\linewidth]{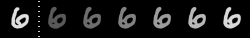}
        \end{subfigure}
    
        \begin{subfigure}{0.88\linewidth}
            \centering
            \includegraphics[width=\linewidth]{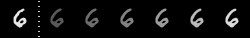}
        \end{subfigure} 
    
        \begin{subfigure}{0.88\linewidth}
            \centering
            \includegraphics[width=\linewidth]{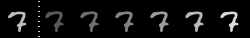}
        \end{subfigure}
    
        \begin{subfigure}{0.88\linewidth}
            \centering
            \includegraphics[width=\linewidth]{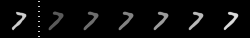}
        \end{subfigure} 
    
        \begin{subfigure}{0.88\linewidth}
            \centering
            \includegraphics[width=\linewidth]{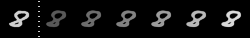}
        \end{subfigure}
    
        \begin{subfigure}{0.88\linewidth}
            \centering
            \includegraphics[width=\linewidth]{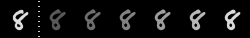}
        \end{subfigure} 
    
        \begin{subfigure}{0.88\linewidth}
            \centering
            \includegraphics[width=\linewidth]{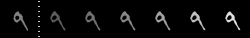}
        \end{subfigure}
    
        \begin{subfigure}{0.88\linewidth}
            \centering
            \includegraphics[width=\linewidth]{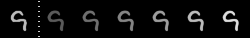}
        \end{subfigure} 
    
    \caption{$do$(intensity).}
    \label{fig:morphoMNIST-sample-intensity}
    \end{subfigure}
    
    \caption{Sampled results on the test set of Morpho-MNIST. The left most image of each set is the original image. These results are non-cherry-picked in terms of model performance (picking was conducted to make sure there is a variety of digits and styles, but not based on performance).}
    \label{fig:morpho-mnist-sample}
\end{figure}

\begin{figure}[ht!]
    \centering
    \begin{subfigure}{0.88\linewidth}
        \centering
        \includegraphics[width=0.44\linewidth]{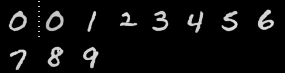}
    \end{subfigure} 

    \begin{subfigure}{0.88\linewidth}
        \centering
        \includegraphics[width=0.44\linewidth]{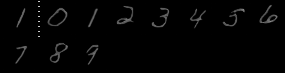}
    \end{subfigure} 

    \begin{subfigure}{0.88\linewidth}
        \centering
        \includegraphics[width=0.44\linewidth]{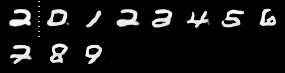}
    \end{subfigure} 

    \begin{subfigure}{0.88\linewidth}
        \centering
        \includegraphics[width=0.44\linewidth]{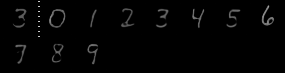}
    \end{subfigure} 

    \begin{subfigure}{0.88\linewidth}
        \centering
        \includegraphics[width=0.44\linewidth]{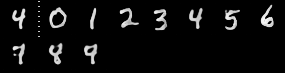}
    \end{subfigure} 

    \begin{subfigure}{0.88\linewidth}
        \centering
        \includegraphics[width=0.44\linewidth]{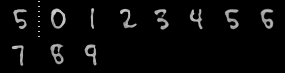}
    \end{subfigure} 

    \begin{subfigure}{0.88\linewidth}
        \centering
        \includegraphics[width=0.44\linewidth]{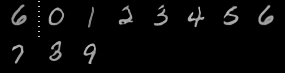}
    \end{subfigure} 

    \begin{subfigure}{0.88\linewidth}
        \centering
        \includegraphics[width=0.44\linewidth]{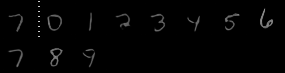}
    \end{subfigure} 

    \begin{subfigure}{0.88\linewidth}
        \centering
        \includegraphics[width=0.44\linewidth]{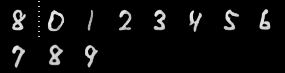}
    \end{subfigure} 

    \begin{subfigure}{0.88\linewidth}
        \centering
        \includegraphics[width=0.44\linewidth]{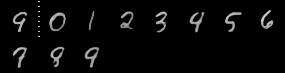}
    \end{subfigure} 

    \caption{Sampled results on the test set of Morpho-MNIST on digit intervention. Note that these results do not have quantifiable counterfactual truths which we can evaluate on the capacity of intervening thickness and intensity in Table \ref{result-table-morpho}.}
    \label{fig:morpho-mnist-sample-digit}
\end{figure}

Table \ref{result-table-morpho-std} shows the standard error of metrics reported in Table \ref{result-table-morpho} across five independent runs. Figure \ref{fig:violin-morpho} shows the violin plot of errors reported in Table \ref{result-table-morpho} and Table \ref{result-table-morpho-std} across the corresponding runs. Figure \ref{fig:morpho-mnist-sample}  and  Figure \ref{fig:morpho-mnist-sample-digit} show some sampled results from VCI.

\begingroup

\setlength{\tabcolsep}{5pt} 

\begin{table}[ht!] 
  \caption{Standard error of metrics.}
  \label{result-table-morpho-std}
  \centering
  \resizebox{\textwidth}{!}{
      \begin{tabular}{lcccccccccc}
        \toprule
        & & \multicolumn{3}{c}{Std. of Image MSE ($\cdot 10^{-2}$)} & \multicolumn{3}{c}{Std. of Thickness (th) MAE} & \multicolumn{3}{c}{Intensity (in) MAE ($\cdot 10^{-1}$)} \\
        \cmidrule(r){3-5} \cmidrule(r){6-8} \cmidrule(r){9-11}
        & $\beta$ & $do($th$)$ & $do($in$)$ & mix & $do($th$)$ & $do($in$)$ & mix & $do($th$)$ & $do($in$)$ & mix \\
        \midrule
        DEAR  & & 0.20 & 0.06 & 0.11 & 0.18 & 0.10 & 0.21 & 0.32 & 0.49 & 0.25 \\
        Diff-SCM  & & 0.02 & 0.02 & 0.02 & 0.01 & 0.01 & 0.01 & 0.01 & 0.02 & 0.01 \\
        CHVAE  & 1 & 0.53 & 0.86 & 0.62 & 0.03 & 0.15 & 0.10 & 0.33 & 0.32 & 0.20 \\
        CHVAE  & 3 & 0.90 & 0.83 & 0.94 & 0.04 & 0.21 & 0.08 & 0.19 & 0.10 & 0.17 \\
        MED  & 1 & 0.92 & 0.26 & 0.59 & 0.11 & 0.08 & 0.12 & 0.31 & 0.19 & 0.23 \\
        MED  & 3 & 1.34 & 0.40 & 0.86 & 0.05 & 0.06 & 0.07 & 0.25 & 0.09 & 0.15 \\
        \hdashline
        SAE  & & 0.12 & 0.05 & 0.05 & 0.10 & 0.04 & 0.05 & 0.15 & 0.10 & 0.08 \\
        VCI  & & 0.07 & 0.01 & 0.07 & 0.03 & 0.01 & 0.05 & 0.11 & 0.05 & 0.06 \\
        \bottomrule
      \end{tabular}
  }
\end{table}

\endgroup

\end{document}